\documentclass[lettersize,journal]{IEEEtran}
\usepackage{amsmath, amsfonts, amssymb, mathtools, cuted}
\usepackage{algorithmic}
\usepackage{algorithm}
\usepackage{array}
\usepackage[caption=false,font=normalsize,labelfont=sf,textfont=sf]{subfig}
\usepackage{textcomp}
\usepackage{stfloats}
\usepackage{url}
\usepackage{verbatim}
\usepackage{graphicx}
\usepackage{xcolor,colortbl}
\usepackage{cite}
\usepackage{makecell}
\usepackage{multirow,multicol}
\usepackage{enumitem}
\usepackage{booktabs}
\usepackage{titlesec}
\usepackage{bbm}
\usepackage{mdframed}
\usepackage{pgfplots,pgfplotstable,tikz}
\pgfplotsset{compat=1.8}

\usepackage[normalem]{ulem} 

\definecolor{lightred}{RGB}{255,200,200}
\definecolor{lightblue}{RGB}{173,216,230}

\newcommand\hlyellow[1]{#1}
\newcommand\hlred[1]{#1}
\newcommand\hlblue[1]{#1}
    \pgfplotstableread{ % data 
Label   2017    2018    2019    2020    2021    2022    2023    2024
AO  0   0   0   0   3   3   9   3
LS  0   3   9   4   14  32  33  3
DM  1   1   3   4   6   7   12  0
FDA 0   0   1   1   3   8   7   0
    }\testdata

\begin{document}

\title{Federated Domain Generalization: A Survey}

\author{Ying~Li,
        Xingwei~Wang,
        Rongfei~Zeng,        
        Praveen~Kumar~Donta,~\IEEEmembership{Senior~Member,~IEEE}, 
        Ilir~Murturi,~\IEEEmembership{Member,~IEEE}, 
        Min~Huang,
        and Schahram~Dustdar,~\IEEEmembership{Fellow,~IEEE}% <-this % stops a space

\IEEEcompsocitemizethanks{\IEEEcompsocthanksitem Ying Li is with the College of Computer Science and Engineering, Northeastern University, Shenyang 110819, China, and Distributed Systems Group, TU Wien, Vienna 1040, Austria. Email: liying1771@163.com. 
\IEEEcompsocthanksitem Xingwei Wang is with the College of Computer Science and Engineering, Northeastern University, Shenyang 110819, China. Email: wangxw@mail.neu.edu.cn. 
\IEEEcompsocthanksitem Rongfei Zeng is with the College of Software, Northeastern University, Shenyang 110819, China. Email: zengrf@swc.neu.edu.cn.
\IEEEcompsocthanksitem Min Huang is with the College of Information Science and Engineering, Northeastern University, Shenyang 110819, China. Email: mhuang@mail.neu.edu.cn.
\IEEEcompsocthanksitem P. K. Donta, I. Murturi, and S. Dustdar are with  Distributed Systems Group, TU Wien, Vienna 1040, Austria. Email: \{pdonta, imurturi, dustdar\}@dsg.tuwien.ac.at. \\
*Corresponding author: Xingwei Wang.
} 
% \thanks{}
}

% The paper headers
\markboth{Proceeding of the IEEE,~Vol.~X, No.~X, June~2024}%
{Li \MakeLowercase{\textit{et al.}}: Federated Domain Generalization: A Survey}

\maketitle

\begin{abstract}
Machine learning typically relies on the assumption that training and testing distributions are identical and that data is centrally stored for training and testing. However, in real-world scenarios, distributions may differ significantly and data is often distributed across different devices, organizations, or edge nodes. 
\hlblue{Consequently, it's to develop models capable of effectively generalizing across unseen distributions in data spanning various domains.} In response to this challenge, there has been a surge of interest in federated domain generalization in recent years. \hlblue{Federated Domain Generalization synergizes Federated Learning and Domain Generalization techniques, facilitating collaborative model development across diverse source domains for effective generalization to unseen domains, all while maintaining data privacy.} \hlyellow{However, generalizing the federated model under domain shifts remains a complex, underexplored issue. This paper provides a comprehensive survey of the latest advancements in this field.} Initially, we discuss the development process from traditional machine learning to domain adaptation and domain generalization, leading to federated domain generalization as well as provide the corresponding formal definition. \hlblue{Subsequently, we classify recent methodologies into four distinct categories: federated domain alignment, data manipulation, learning strategies, and aggregation optimization, detailing appropriate algorithms for each. We then overview commonly utilized datasets, applications, evaluations, and benchmarks. Conclusively, this survey outlines potential future research directions.}
\end{abstract}


\section{Introduction}
\IEEEPARstart{H}{umans} possess an extraordinary aptitude to transfer their knowledge and expertise in novel situations and environments they have not experienced before, while machines struggle to reproduce this proficiency, especially regarding \textit{out-of-distribution} (OOD) data \cite{zhou2022}. 
\hlyellow{This raises pertinent questions regarding the efficacy of \textit{machine learning} (ML) models across different contexts. For instance, can a model train to detect tumors in medical images obtained from one type of scanner maintain its effectiveness when applied to images from another scanner or imaging protocol? Similarly, does a robot trained to execute a task within a simulated environment possess the capability to generalize its skills to the real world?} Additionally, there is a concern about whether a speech recognition system trained on audio recordings from one speaker would perform well when tested on recordings from a different speaker or whether a language model trained on text from one domain (e.g., news articles) would apply to text from a different domain (e.g., social media). \hlyellow{The resolution of these questions hinges on the ML models' ability to navigate the prevalent issue of domain/dataset shift\mbox{\cite{dataset}}. Domain shift refers to the discrepancy in distribution between the source domain employed for training and the target domain utilized for testing\mbox{\cite{domain1, domain2, domain3}}. It is a pervasive issue in ML, significantly impacting the models' generalization and robustness across unseen domains in real-world scenarios.}

Recently, deep learning has made revolutionary advances and emerged as a powerful tool in a variety of fields, including image recognition, natural language processing, speech recognition, and robotics, among others. \hlyellow{However, the prevalent assumption in deep learning algorithms is that all samples in both training and test datasets are \textit{independent and identically distributed} (IID)\mbox{\cite{vsajina2023peer}}—an assumption that often does not hold in OOD scenarios encountered in real-world applications. This discrepancy, known as domain shift, can significantly degrade a model's performance when applied to unseen domains. To tackle the issue of domain shift, the technique of \textit{domain generalization} (DG)\mbox{\cite{blanchard2011generalizing, zhou2022, wang2020generalizing}} has been developed, enhancing model generalization across disparate domains, irrespective of their similarity to the training domain.}

\begin{table*}[!t]
\caption{Comparisons of Relative Surveys.}\label{tab:table1}
\centering
% \fbox{
\begin{tabular}{rccp{9cm}}
    \toprule
    \textbf{ Ref.} & \textbf{Year} & \textbf{Domain} & \textbf{Scope} \\
    \midrule
    \textit{Bonawitz et al.}\cite{bonawitz2019towards} & 2019 & Federated Learning & A scalable production system for FL in the domain of mobile devices.\\
    \hline
    \textit{Yang et al.}\cite{yang2019federated} & 2019 & Federated Learning & A comprehensive secure federated learning framework that includes horizontal, vertical, and transfer learning.\\
    \hline
    \textit{Li et al.}\cite{li2020federated} & 2020 & Federated Learning & Unique characteristics, challenges, current approaches, and future research areas. \\
    \hline
    \textit{Kairouz et al.}\cite{kairouz2021advances} & 2021 & Federated Learning & Recent advancements, the challenges and open problems. \\
    \hline
    \textit{Csurka et al.}\cite{csurka2017domain} & 2017 & Domain Adaptation & An overview of DA and transfer learning in visual applications  \\
    \hline
    \textit{Kuw et al.}\cite{kouw2019review} & 2019 & Domain Adaptation & Sample-based, feature-based, and inference-based methods. \\
    \hline
    \textit{Wilson et al.}\cite{wilson2020survey} & 2020 & \makecell {Domain Adaptation } & A unsupervised domain adaptation methods in deep learning. \\
    \hline
    \textit{Wang et al.}\cite{wang2022generalizing} & 2018 & Domain Generalization & Data manipulation, representation learning, and learning strategy.  \\
    \hline
    \textit{Zhou et al.}\cite{zhou2022} & 2022 & Domain Generalization &  Background, existing methods and theories, and insights and discussions on future research directions \\
    \hline
    \textit{Sheth et al.}\cite{sheth2022domain} & 2022 & Domain Generalization & Invariance via Causal Data Augmentation, Invariance via Causal representation learning, and Invariance via Transferring Causal mechanisms. \\
    \hline
    \textit{Shen et al.}\cite{shen2021towards} & 2021 & Out-of-Distribution Generalization & Definition, methodology, evaluation, and implications of Out-of-Distribution generalization.  \\
    \hline
    \textit{Li et al.}\cite{li2022out} & 2022 & Out-of-Distribution Generalization & Existing methods, theories, and future research directions related to out-of-Distribution generalization.  \\
    \hline
    \textbf{Our work} & 2023 & Federated Domain Generalization & Theories, methodologies, datasets, applications, evaluations, benchmarks, and future research directions. \\
    \bottomrule
\end{tabular}
% }
\end{table*}

There has been significant research in the area of DG, and numerous strategies have been developed to tackle the issue of domain shift. Predominantly, DG methods have been centralized, utilizing a central server that accesses data from all source domains for DG tasks. \hlblue{Specifically, domain alignment methods\mbox{\cite{muandet, li2018domain, li2018deep, shao2019multi, erfani2016robust, motiian2017unified, yoon2019generalizable}} aim to align feature distributions between the source domains and target domains, thereby improving model generalization. This approach directly tackles the issue of domain shift by ensuring that the model perceives the source and target domains as similar, thereby facilitating improved performance on unseen domains. Data augmentation techniques\mbox{\cite{yue2019domain, zhou2021domain, zhou2020learning, volpi2019addressing, shankar2018generalizing, volpi2018generalizing, xu2020robust}} seek to expand the training dataset's size and diversity, enhancing adaptability without new data collection. Data augmentation acts as a complementary strategy to domain alignment by enriching the model's training environment, thereby indirectly reducing the impact of domain discrepancies. Meta-learning strategies\mbox{\cite{li2018learning, balaji2018metareg, li2019feature, liu2020shape, li2019episodic, zhao2021learning, dou2019domain}} focus on optimizing the model’s initialization or optimization process for quick adaptation to new tasks or domains with minimal additional training. This approach addresses the generalization issue by preparing the model to efficiently learn from limited data in new domains, rather than directly aligning domain features or augmenting data. Ensemble learning methods\mbox{\cite{liu2020ms, zhou2021domainel, xu2014exploiting, niu2015multi, ding2017deep, seo2020learning, cha2021domain}} synthesize predictions from models trained on various domains, using ensemble techniques to identify common patterns and enhance generalization to unseen domains. This technique enhances model robustness and generalization by aggregating insights from multiple perspectives, thereby mitigating the risk of overfitting to domain-specific features and elevating performance on unseen domains.} However, these methods may significantly infringe upon data protection laws like the EU/UK \textit{General Data Protection Regulation} (GDPR) \cite{truong2021privacy}, as tremendous privacy-containing data is stored in locally distributed locations due to the rise of ML and the proliferation of \textit{Internet of Things} (IoT) devices \cite{li2023varf}. For instance, public transportation, ride-sharing services, and logistics companies data on the transportation industries; banks, insurance companies, and credit rating agencies data of financial institutions; patient privacy data of healthcare institutions; schools, universities, and online learning platforms data of educational institutions. Therefore, developing a highly generalizable model in real-world scenarios that can ensure both privacy and high model performance is a dilemma \cite{yuan2023collaborative}. On one hand, it requires careful consideration of privacy concerns across multiple domains. On the other hand, without simultaneous access to source domains, it would be difficult to accurately identify and learn domain-invariant features for improving model generalization.
\begin{figure*}[htbp]
	\begin{center}
		\includegraphics[width=\linewidth]{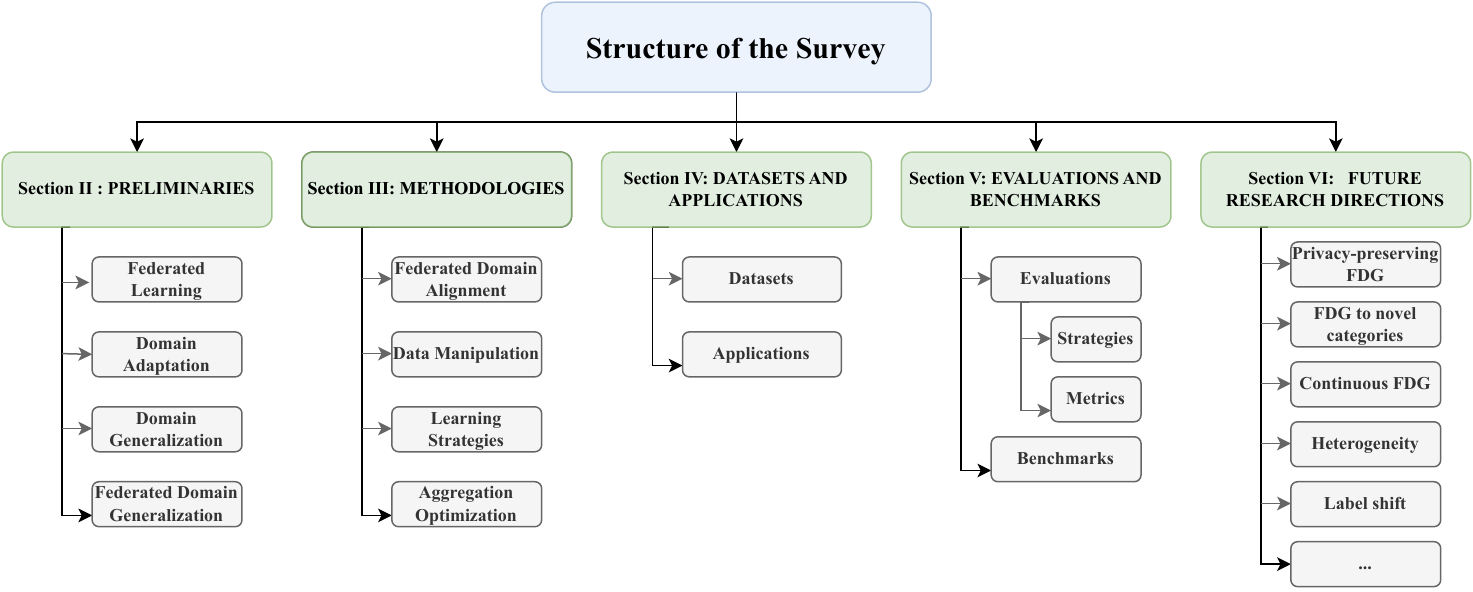}
		\caption{The Structure of our proposed survey.}
		\label{structure}
	\end{center} 
\end{figure*}

\hlred{Resolving the above dilemma necessitates a thorough comprehension of \textit{federated learning} (FL)\mbox{\cite{mcmahan2016federated}}. As a promising paradigm within distributed machine learning, FL facilitates the training of algorithms across a wide array of decentralized edge devices or servers. Each of these clients retains its local data samples, thereby preventing the need for direct data exchange. This approach contrasts with traditional centralized learning methods, which necessitates aggregating data on a single server. FL primarily aims to utilize rich and diverse data sources, ensuring user privacy and reducing the need for extensive data transmission. Building upon this, \textit{federated domain generalization} (FDG)\mbox{\cite{liu2021feddg, zhang2023federated, wu2021collaborative, yuan2023collaborative}} leverages FL alongside DG techniques to enable multiple source domains to collaboratively learn a model that adeptly generalize well to unseen domains while keeping their data private. However, the technical complexity of achieving model generalization across domain shifts in FL settings remains a challenge with limited research focus thus far. Given the evolving landscape of DG research, the introduction of the FL, characterized by distributed data domains, presents new challenges that necessitate innovative approaches.} As stated above, most conventional solutions achieve DG by accessing multi-source domains in a centralized manner, while each client can only access its local data in FDG. In addition, local optimization of the FL model may result in a bias toward its data distribution, which reduces its generalizability to new target domains. \hlred{As a result, conventional DG methods are unsuitable for the unique requirements of FDG scenarios.}
\begin{table}[!b]
\caption{Frequently used Acronyms}\label{tab:Acronyms}
\centering
% \fbox{
\begin{tabular}{c|l}
    \toprule
     \textbf{Acronym} & \textbf{Meaning}  \\
    \midrule
     ML & Machine Learning   \\ 
     FL & Federated Learning  \\
     DA & Domain Adaptation \\
     FDA & Federated Domain Adaptation\\
     DG & Domain Generalization \\
     FDG & Federated Domain Generalization  \\
     DIR & Domain-Invariant Representation  \\
     MMD & Maximum Mean Discrepancy \\
     DANN & Domain Adversarial Neural Networks\\ 
     DFKD & Data-free Knowledge Distillation \\ 
     UDA & Unsupervised Domain Adaptation \\
     WD  & Wasserstein Distance\\
     SGD & Stochastic Gradient Descent \\
     NLP & Natural language processing\\
     MTSSL & Multi-Task Self-Supervised Learning  \\
     IID & Independent and Identically Distributed \\
     IoT & Internet of Things \\
     OOD & Out of Distribution \\
     FID & Fréchet inception distance \\ 
     LFRL & Lifelong Federated Reinforcement Learning \\ 
     FZSL & Federated Zero-Shot Learning\\
     GAN & Generative Adversarial Networks \\ 
     GDPR & General Data Protection Regulation\\
     IRM & Invariant Risk Minimization  \\
     KL & Kullback-Leibler Divergence\\
     KD & Knowledge Distillation \\
     MAML & Model-Agnostic Meta-Learning \\
     mIoU & mean Intersection over Union\\
    \bottomrule
\end{tabular}
% }
\end{table}

This paper presents the first comprehensive survey on the topic of FDG, aiming to introduce recent advances in this field with an emphasis on its theories, methodologies, datasets, applications, evaluations, benchmarks, and future research directions. To our knowledge, there have been few recent attempts \cite{bonawitz2019towards, yang2019federated, li2020federated, kairouz2021advances, csurka2017domain, kouw2019review, wilson2020survey, wang2022generalizing, zhou2022, sheth2022domain, shen2021towards, li2022out} to explore this area, but their focus differs significantly from the scope of our survey. Table~\ref{tab:table1} provides a summary of comparisons between existing but related surveys and our work. Specifically, Bonawitz \textit{et al.} \cite{bonawitz2019towards} introduced a system design for federated learning (FL) on mobile devices utilizing TensorFlow, addressing the challenges and open questions in mobile phone domains. Yang \textit{et al.} \cite{yang2019federated} developed a secure FL framework as a solution to the obstacles faced by AI in industries where data exists in isolated islands and data privacy and security are strengthened. The framework includes horizontal FL, vertical FL, and federated transfer learning (FTL), and can be applied to various businesses successfully. Li \textit{et al.} \cite{li2020federated} \hlyellow{explored the principles of FL, which entails training statistical models on remote devices or in siloed data centers while maintaining data localization. The authors elaborated on the intrinsic challenges of FL, summarizing existing strategies and articulating avenues for future investigation. Kairouz \textit{et al.} \mbox{\cite{kairouz2021advances}} surveyed recent advancements, pinpointing the challenges and open problems in FL research.} 

\hlblue{Kouw \textit{et al.}\mbox{\cite{kouw2019review}} conducted a comprehensive review of recent advancements in DA, examining the fundamental question: how can a classifier effectively learn from a source domain and generalize to a target domain? The authors presented different approaches to solving the problem of DA in ML and they are categorized into sample-based, feature-based, and inference-based methods.} Csurka \textit{et al.} \cite{csurka2017domain} provided an overview of DA and transfer learning in visual applications. They discuss the state-of-the-art methods for different scenarios, including shallow and deep methods, and relate DA to other ML solutions. \hlred{Wilson \textit{et al.} \mbox{\cite{wilson2020survey}} conducted an extensive survey on the methodologies used in single-source unsupervised deep DA, which integrates deep learning with DA techniques to reduce reliance on costly target domain annotations. They performed a detailed comparative analysis of various approaches within this realm, focusing on the evaluation of alternative methods, discerning unique and shared components, analyzing outcomes and theoretical frameworks, and investigating potential applications and future research directions.}

\hlblue{DG extends the ideas behind DA by preparing a model to generalize across unseen target domains without requiring access to their data during the training phase.} Wang \textit{et al.} \cite{wang2022generalizing} reviewed recent advances in DG, which deals with the call into question of developing models  capable of generalizing to unseen test domains. The authors also discuss the theories, algorithms, datasets, applications, and potential research topics related to DG. Zhou \textit{et al.} \cite{zhou2022} provided a comprehensive literature review of DG in ML, in which the review covers the background, existing methods and theories, and potential future research directions. Sheth \textit{et al.} \cite{sheth2022domain} discussed the problem of distribution shifts in ML models and how it affects their generalization capabilities. They survey various DG methods that leverage causality to identify stable features or mechanisms that remain invariant across different distributions. \hlblue{OOD Generalization broadens the concept of DG to address the challenge of ensuring model robustness when encountering data that significantly deviates from the training distribution, including entirely new scenarios that were not represented in the training data. Shen \textit{et al.}\mbox{\cite{shen2021towards}} discussed the OOD generalization problem in ML and provided a comprehensive review of existing methods, evaluation metrics, and future research directions in OOD generalization. The primary objective is to develop and rigorously assess models and approaches capable of adapting to distributional changes, demonstrating robust generalization, and resolving the prevalent issue of inconsistencies in data distributions encountered in real-world settings.} Li \textit{et al.} \cite{li2022out} reviewed recent advances in OOD generalization on graphs, which goes beyond the in-distribution hypothesis and presents a detailed categorization of methodologies into three categories based on their integration points within the graph ML pipeline. 
\begin{figure*}[t]
	\begin{center}
		\includegraphics[height=11.5cm, width=14cm]{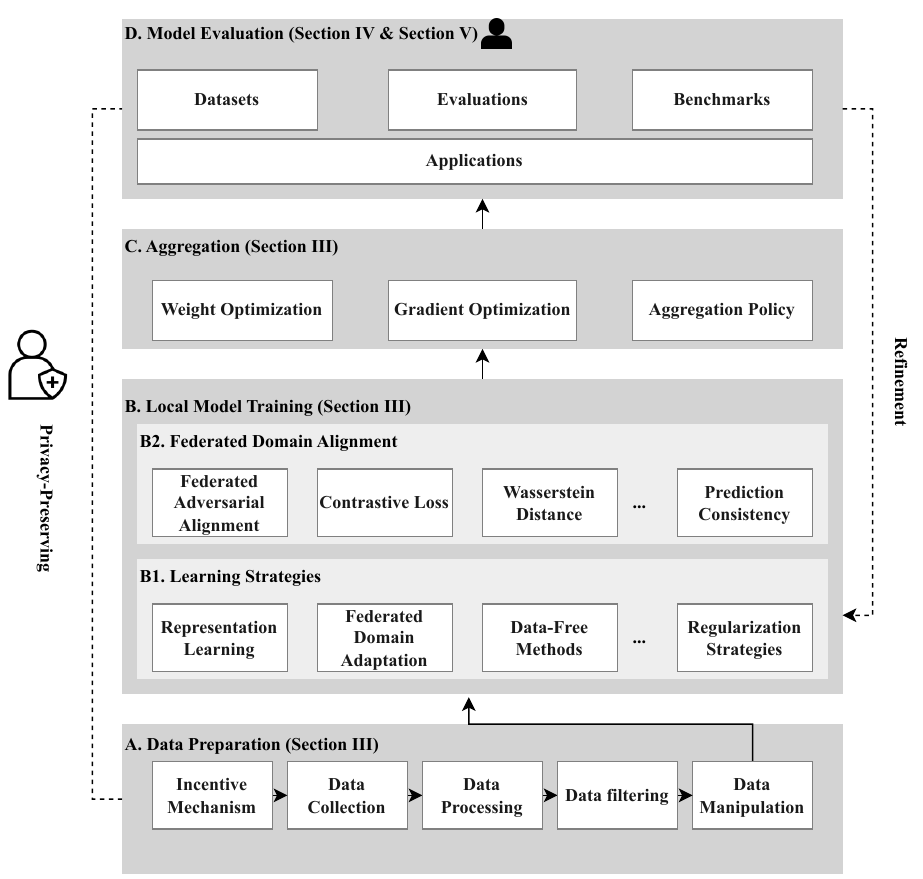}
		\caption{The architecture and workflow of FDG.}
		\label{workflow}
	\end{center} 
\end{figure*}

\hlblue{FDG can be seen as an extension of OOD generalization to the FL paradigm, where the call into question is not only to handle unseen distributions but also to do so in a manner that is compatible with the decentralized and privacy-preserving principle of FL.} Despite the notable advancements in FL and DG across various fields recently, a comprehensive survey that cohesively covers theories, methodologies, datasets, applications, evaluations, benchmarks, and future research directions in FDG has yet to be undertaken. Our objective is to furnish researchers with an in-depth review of the current landscape and to catalyze further investigation in FDG and related fields. To achieve this, we have conducted an extensive review of the existing literature and synthesized the main ideas and contributions of the most relevant works. We aim for this survey will serve as a valuable resource for researchers aiming to expand their knowledge of FDG and to foster progress in this critical area.

The organization of this paper is delineated as follows, illustrated in Fig.~\ref{structure}. Frequently used acronyms are summarized using table~\ref{tab:Acronyms}. We begin by providing an introduction to the field of FL, traditional ML, DA, DG, and FDG in Section~\ref{PRELIMINARIES}. Then, we introduce the existing methodologies for FDG, including federated domain alignment, data manipulation, learning strategies, and aggregation optimization algorithms in Section~\ref{methods}. Next, we present the datasets and applications in Section~\ref{dataset} and evaluations and benchmarks in Section~\ref{evaluations}. Furthermore, in Section~\ref{future}, we provide a concise summary of the insights derived from existing research and engage in a discussion concerning the future research directions in the field. Finally, a comprehensive conclusion of this paper is presented in Section~\ref{conclusion}, summarizing the key findings and insights derived throughout the study. On the analysis of the sections, we derive a generalized architecture and workflow of FDG, as depicted in Fig.~\ref{workflow}. 

\section{PRELIMINARIES} \label{PRELIMINARIES}
In this section, we begin by providing an overview of FL, DA, DG, and FDG. Subsequently, we discuss the integration of FL and DG. For the convenience of readers, we have summarized frequently used notations in Table~\ref{tab:Notations}.

\begin{table}[!b]
\caption{Frequently used Notation.}\label{tab:Notations}
\centering
% \fbox{
% \resizebox{\linewidth}{!}{
\begin{tabular}{c|p{6cm}}
    \toprule
     \textbf{Notation} & \textbf{Meaning} \\
    \midrule
     $\mathcal{N}$ & The set of clients\\
     $\mathbb{N}$ & The number of clients in FL.\\
     $\mathcal{M}$ & The set of end users in FL.\\
     $\mathbb{M}$ & The number of end users in FL.\\
     $\mathbb{S}_i$ & The dataset of client $i$\\
     $|\mathbb{S}_i|$ & The size of local dataset of client $i$ \\
     $\mathcal{H}$ & Hypothesis space.    \\
     $h$ & Labeling function from a $\mathcal{H}$.    \\
     $\mathcal{X}$ & Non-empty input space.    \\
     $\mathcal{Y}$ & Output space.    \\
     $P_{XY}$ & Probability distribution of the joint input-output space.   \\
     $\mathcal{S}$ & Domain. \\
     $M$ & The number of Domains.  \\
     $\mathcal{S}_{source}$ & The set of the source domains. \\
     $\mathcal{S}_{target}$ & The set of the target domains. \\
     $M$ & The number of source domains.  \\
     $n$ & The size of the domain.    \\
     $\mathcal{S}^i$ & Domain $i$. \\
     $n_i$ & The size of the domain $i$.    \\
    $\varkappa_\imath$ & The selected subset from the local dataset $\mathcal{S}_i$ of client $i$ for local model training. \\
     $|\varkappa_\imath|$ & The size of the chosen subset $\varkappa_i$ from client $i$.\\
     $\omega_t$ & Aggregated global model at round $t$. \\
     $\omega_{t+1}^\imath$ & Local model from client $i$ at round $t+1$. \\
     $\omega^*$ & The optimal weights of the global model.\\
     $\eta$ & The learning rate of FL training.\\
     $\theta(h,\hat{h})$ & The difference in the source domain between two labeling function $h$ and $\hat{h}.$ \\
     $\theta^s(h)$ & The risk of labeling function $h$ on the source domains. \\
     $\theta^t(h)$ & The risk of labeling function $h$ on the target domains. \\
     $\lambda_\mathcal{H}$ & Complexity of $\mathcal{H}$ for prediction tasks. \\
     $d$ & \textit{Vapnik-Chervonenkis} dimension of $\mathcal{H}$. \\
     $\mathbb{E}$ & Expectation operator. \\
     $\mathcal{U}^s$ & Unlabeled samples of size $k$ from source domain.\\
     $\mathcal{U}^t$ & Unlabeled samples of size $k$ from target domain. \\
     $\hat \psi(h)$ & The estimation of the average risk over all possible target domains.\\
     $\pi_i$ & $\pi$ A normalized mixing weights.\\
     $F^t((x^t_i)$ & The feature representation of the target domain data.\\
     $MMD^2(P,Q)$ & The square of maximum mean discrepancy.\\
    \bottomrule
\end{tabular}
% }
\end{table}

\subsection{Federated Learning}
Artificial intelligence has swiftly emerged as a fundamental element of our daily lives, fueled by notable advancements and breakthroughs in deep learning techniques across various domains. Moreover, the widespread adoption of IoT devices has resulted in enormous amounts of data being stored in locally distributed locations. In addition, with the proliferation of sensitive and private data, concerns over data privacy have become increasingly prominent. In response to these concerns, FL has been introduced as a solution to \hlyellow{uphold} data privacy concerns while facilitating collaborative ML.

\begin{figure*}[htbp]
	\begin{center}
		\includegraphics[height=8.5cm, width=14cm]{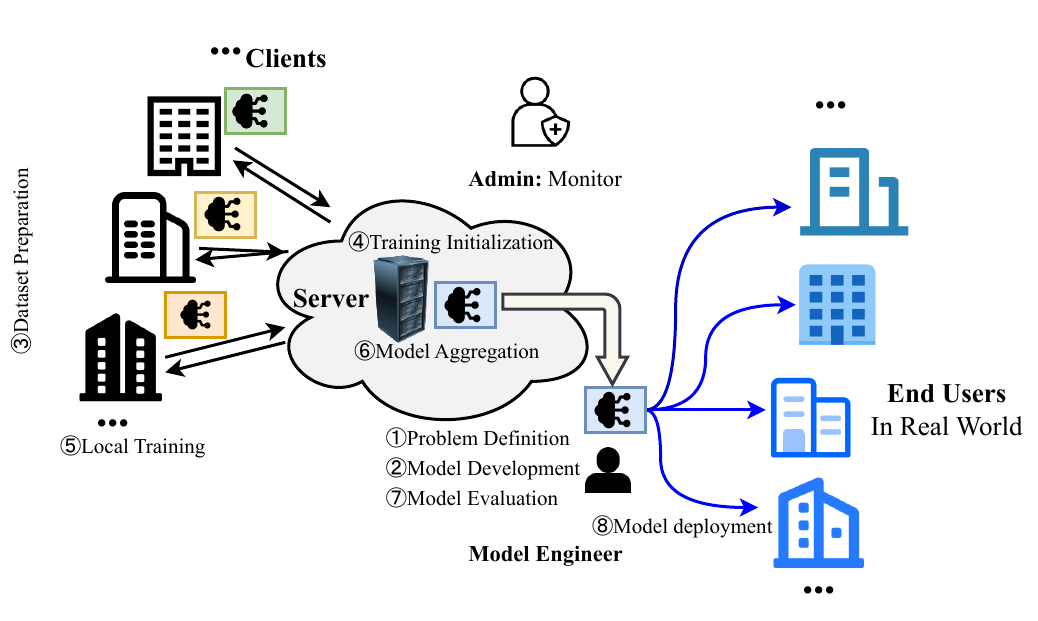}
		\caption{The lifecycle of a trained model in FL System.}
		\label{FL}
	\end{center} 
\end{figure*}

As illustrated in Fig.~\ref{FL}, we depict a scenario consisting of a model engineer, an admin, a central server, $\mathbb{N}$ clients, and $\mathbb{M}$ end users. The set of clients is denoted by $\mathcal{N}=\left\{1, 2, \cdots, \imath, \cdots, \mathbb{N}\right\}$, while the set of end users is denoted by $\mathcal{M}=\left\{1, 2, \cdots, \jmath, \cdots, \mathbb{M}\right\}$. Each client $\imath \in\mathcal{N}$ possesses a local dataset $\mathbb{S}^\imath$, which may change over time. The dataset's size, denoted by $\mathbb{S}^\imath$, allows clients to engage iteratively in FL processes, reflecting the dynamic nature of their datasets. For local model training, a client may select a subset $\varkappa^\imath \subseteq \mathbb{S}^\imath$, with its size represented by $|\varkappa^\imath|$. Next, we will introduce the lifecycle, key actors, and typical processes within an FL system, respectively.

\subsubsection{The Lifecycle of a Trained Model in FL System.}
\ 
\newline
\indent The FL process is typically driven by a model engineer who develops a model for a particular application. Here's a high-level overview of a typical FL workflow:
\begin{description}
\item[\textbf{Step 1:} ]~\textbf{Problem Definition.} The first step in the FL workflow is to define the problem that the model will solve. This involves identifying the data sources and the target task, as well as specifying the performance metrics that will be used to evaluate the model.
\item[\textbf{Step 2:} ]~\textbf{Model Development.} Once the problem is defined, the model engineer develops a model architecture that is suitable for the FL paradigm. This involves selecting the appropriate optimization algorithm, defining the model parameters, and deciding how to partition the data among the participating clients.

\item[\textbf{Step 3:} ]~\textbf{Dataset Preparation.} The dataset is prepared by the data owners, who collect or generate the data that will be used for training the model. The data is partitioned among the clients according to the protocol defined in Step 2.

\item[\textbf{Step 4:} ]~\textbf{Training Initialization.} The central server initializes the training process by sending the initial model weights to the participating clients. This model can be randomly initialized or pre-trained on a large dataset.

\item[\textbf{Step 5:} ]~\textbf{Local Training.} The participating clients perform local training using their own data and the initial model weights. This training is typically done using stochastic gradient descent (SGD) or a variant of it.

\item[\textbf{Step 6:} ]~\textbf{Model Aggregation.} After local training, the clients return their updated model weights to the central server, which aggregates them using an algorithm such as Federated Averaging (FedAvg) or other Optimized aggregation algorithms.

\item[\textbf{Step 7:} ]~\textbf{Model Evaluation.} The engineer evaluates the aggregated model on a validation dataset to determine its accuracy and generalization performance. If the performance is unsatisfactory, the training process can be repeated with a different set of initial weights or hyperparameters.

\item[\textbf{Step 8:} ]~\textbf{Model Deployment.} Once the model is trained, it can be deployed to predict new and unseen domains. In some cases, the model may need to be fine-tuned on new data to adapt to changing conditions or to improve its performance.
\end{description}

Overall, the workflow involves multiple iterations of local training, model aggregation, and evaluation, and it requires close collaboration between the model engineer and the participating clients. The success of an FL project depends on careful planning, effective communication, and the ability to adapt to changing conditions and requirements.

\subsubsection{The Key Actors in FL System.}

~
\newline
\indent FL encompasses a variety of stakeholders, each contributing distinct roles in the model's training and deployment phases. The following outlines the key actors within an FL framework:

\begin{description}
\item[\textbf{Model Engineer:}] The Model Engineer plays a pivotal role in the FL system, tasked with the development of the ML model and the establishment of the training protocol. This role involves selecting the appropriate model architecture, optimization algorithm, and partitioning strategy for the distributed data. The model engineer also evaluates the performance of the trained model and fine-tunes it as needed.

\item[\textbf{Admin:}] In FL, an admin is a role that involves managing the overall operation of the FL system. An admin may be an individual or a team responsible for configuring and monitoring the system, as well as ensuring its security, privacy, and compliance with legal and ethical requirements.

\item[\textbf{Central Server:}] The central server coordinates the FL training process by collecting and aggregating the model parameters from the participating clients. The server performs the model aggregation using algorithms.

\item[\textbf{Participating Clients:}] Participating clients are the devices or organizations that contribute their local data and computational resources to the FL training process. Participating clients perform local training on their data and share their model parameters with the central server for aggregation.

\item[\textbf{End Users:}] End-users, comprising individuals or organizations, utilize the trained model for predictions or decision-making in novel or previously unseen domains. End-users may include consumers, businesses, or government agencies that benefit from the insights generated by the model. Depending on the application requirements, the trained model may be deployed on a mobile device, a cloud server, or an edge device.

\item[\textbf{Communication Channel:}] The communication channel is the medium used for transmitting the model weights between the central server and the participating clients. It can be a wireless or wired network connection and must be designed to ensure the privacy and security of the data transmitted between the parties involved.
\end{description}
Overall, an FL system involves multiple actors collaborating to train and deploy a machine-learning model. The success of an FL project depends on the effective coordination and communication between these actors, as well as the careful consideration of privacy, security, and performance concerns.

\subsubsection{The Typical Process of FL Training.}
\ 
\newline
\indent \hlblue{FL endeavors to construct a robust global model by harnessing insights from distributed data without compromising privacy. The objective of FL is to achieve a global model $\omega^*$ that optimally minimizes the global loss function $L(\omega)$ across all participating clients. Achieved through iterative local training and aggregation, this process enables learning from varied data distributions without direct data access. The typical FL training process unfolds through several essential steps:}

\begin{description}
\item[\textbf{Step 1: Client Selection.}] \hlyellow{In FL, client selection is a pivotal step that determines participant involvement in the training process. The objective is to identify a subset of clients, denoted as $\mathcal{N}^s$ from all clients $\mathcal{N}$, where $\mathcal{N}^s \subseteq \mathcal{N}$, ensuring a representative set that enriches the global model without compromising data privacy or overextending computational resources. Selection strategies may vary from random to more sophisticated methods tailored to identify suitable participants. This is imperative as inappropriate clients — such as poor data quality (e.g., noise, bias, or irrelevance), limited diversity in data, heightened privacy risks, substantial resource constraints, or data that poorly align with project goals—can detract from the process's effectiveness, security, and efficiency. Consequently, meticulous client selection is essential to preserve the integrity of the global model and secure successful training outcomes.}

\item[\textbf{Step 2: Download.}] In this step, the clients first need to download the global model aggregated by the central server in the previous round $\mathrm{t}$ (the global model is randomly initialized in the first round).
\begin{equation}\label{eq1}
	{\omega_\mathrm{t}^\imath} = \omega_\mathrm{t}.
\end{equation}
where $\omega_\mathrm{t}$ denotes the global model in the round $\mathrm{t}$.
\item[\textbf{Step 3: Local Training.}] The participating clients perform local training on their respective domains using the downloaded model. 
\begin{equation}\label{eq2}
	{\omega_{\mathrm{t}+1}^\imath} = \omega_\mathrm{t}^\imath - \eta\triangledown L^\imath(\omega^\imath_\mathrm{t}; \varkappa^\imath).
\end{equation}
where $\eta$ denotes the learning rate, $L^\imath(\omega^\imath_\mathrm{t}; \varkappa^\imath)$ represents the local loss function for client $\imath$ at round t, ${\omega_{\mathrm{t}+1}^\imath}$ represents the updated local model parameters for client $\imath$ at round $\mathrm{t}+1$. 
\item[\textbf{Step 4: Upload.}] The locally trained models are then transmitted to the central server for aggregation.

\item[\textbf{Step 5: Model Aggregation.}] The central server aggregates the model parameters from the participating clients using a suitable algorithm. The aggregation process aims to combine the local model updates into a single global model that reflects the collective knowledge of all clients.

\begin{equation}\label{eq3}
	{\omega_{\mathrm{t}+1}} = \frac{\sum_{\imath \in \mathcal{N}^s} |\varkappa^\imath| \omega_{\mathrm{t}+1}^\imath}{\sum_{\imath \in \mathcal{N}^s} |\varkappa^\imath|}
\end{equation}
where $\omega_{\mathrm{t}+1}$ denotes the aggregated global model in the round $\mathrm{t}+1$.

\item[\hlblue{\textbf{Step 6: Iteration.}}] \hlblue{Return to Step 2 until the convergence criteria are met\mbox{\cite{zeng2020fmore}}, then terminate the training. The iterative nature of this process ensures continuous refinement of the model, leading to the optimal global model $\omega^*$ that minimizes the global loss function $L(\omega)$:}
\begin{equation}\label{eq4}
	L(\omega) = \frac{\sum_{\imath \in \mathcal{N}^s} |\varkappa^\imath| L^\imath(\omega;\varkappa_\imath)}{\sum_{{\imath \in \mathcal{N}^s}} |\varkappa^\imath|}.
\end{equation}
where $L^\imath(\omega;\varkappa_\imath)$ is the loss for subset $\varkappa_\imath$ given $\omega$.
\begin{equation}\label{eq5}
	{\omega}^* = \mathop{\arg\min}_{\omega} {L(\omega)}.
\end{equation}
\end{description}

\hlblue{This process not only safeguards data privacy but also leverages diverse data distributions to enhance the model's accuracy and generalizability. The culmination of this rigorously designed process is a global model that encapsulates the collective knowledge of all clients, finely tuned to achieve the common goal of minimizing the global loss function.}

\subsection{Domain Adaptation}
\begin{figure*}[t]
	\begin{center}
		\includegraphics[height=7cm, width=14cm]{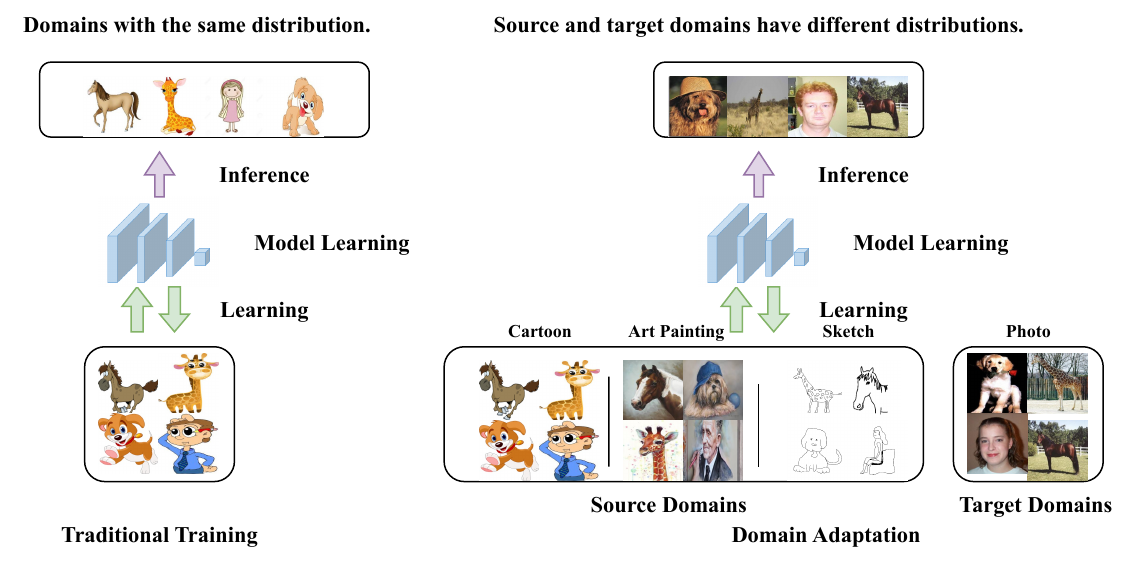}
		\caption{Comparison of traditional ML tasks (left) and domain adaptation tasks (right).}
		\label{compare1}
	\end{center} 
\end{figure*}

Recent studies \cite{recht2019imagenet, hendrycks2019benchmarking, yang2021generalized} have indicated that deep learning models are vulnerable to significant performance degradation when evaluated on out-of-distribution datasets, even with minor variations in the data generation process. Specifically, it demonstrated that deep learning models exhibit limited generalization capacity, which can significantly impact their performance when deployed in real-world scenarios. DA \cite{saenko2010adapting, lu2020stochastic, saito2018maximum, ganin2015unsupervised, long2015learning, liu2020open, li2021learning} emerged as a strategy to circumvent the OOD data challenge by collecting a subset of data from the target domain, thereby enabling the adaptation of a model trained on a source domain with disparate distributions to the target domain. Fig.~\ref{compare1} compares traditional ML tasks and DA tasks. The basic idea behind DA is that, despite variations between source and target domains, shared underlying structures or patterns can be exploited to enhance model performance in the target domain, as delineated in \textbf{Definition 2}.

\noindent\textbf{Definition 1} (Domain). A domain refers to a collection of data that is sampled from a joint distribution. Consider a nonempty input space denoted as $\mathcal{X}$ and an output space denoted as $\mathcal{Y}$. The domain, denoted as $\mathcal{S}$, comprises $n$ samples $\mathcal{S} = \{{(x_j, y_j)}_{j=1}^n \sim P_{XY}\}$, where $x_j \in \mathcal{X} \subset \mathbb{R}^\mathfrak{d}$ represents a $\mathfrak{d}$-dimensional input sample, $y_j \in \mathcal{Y} \subset \mathbb{R}$ represents the corresponding output label, and $P_{XY}$ denotes the probability distribution over the joint input-output space, where $X$ and $Y$ represent the respective random variables.

\noindent\textbf{Definition 2} (Domain Adaptation). Let $h$ represent any labeling function from the hypothesis space $\mathcal{H}$. Given $M$ source domains denoted as $\mathcal{S}_{source} = \{\mathcal{S}^i | i = 1, 2, \cdot\cdot\cdot, M\}$ where $\mathcal{S}^i = \{(x_j^i, y_j^i)\}_{j=1}^{n_i}$ represents the $i$-th domain with true labeling functions $h^{*s}$, it is observed that the joint distributions between each pair of domains differ ($P_{XY}^i \neq P_{XY}^{\hat{\imath}}$, for $1 \leq i \neq \hat{\imath} \leq M$). The goal of DA is to reduce the disparity between the source domains and target domains, enabling effective generalization on the target domain $\mathcal{S}_{target}$, where $\mathcal{S}_{target}$ is accessible during training but has a distinct distribution ($P_{XY}^{target} \neq P_{XY}^i$ for $i \in {1,2, \ldots, M}$), with corresponding true labeling functions denoted as $h^{*t}$.

Therefore, the difference between two labeling functions, $h$ and $\hat{h}$, within the source domain can be quantified as follows:

\begin{equation}\label{eq6}
        \theta^s(h, \hat{h}) = \mathbb{E}_{x \sim P_X^s}[h(x) \ne \hat{h}(x)] = \mathbb{E}_{x \sim P_X^s}[|h(x) - \hat{h}(x)|].
\end{equation}

Similarly, we can get $\theta^t$ when $x \sim P_X^t$. Let $\theta^s(h) := \theta^s(h, h^{*s})$ and $\theta^t(h) := \theta^t(h, h^{*t})$ denote the risk associated with labeling function $h$ in the source domain and target domain, respectively. 
 
DA seeks to reduce the target domain risk, $\theta^t(h)$. However, directly assessing this risk is often impractical due to the lack of explicit knowledge about the target domain's true distribution.  To overcome the issue, Ben \textit{et al.} \cite{ben2010theory} proposed a theoretical framework to approximate the target domain risk, $\theta^t(h)$, using the more accessible source domain risk, $\theta^s(h)$:

\begin{equation}\label{eq7}
        \theta^t(h) \leq \theta^s(h) + d_{\mathcal{H}\Delta\mathcal{H}}(P_X^s, P_X^t) + \lambda_\mathcal{H}.
\end{equation}
where the discrepancy between cross-domain distributions is quantified by the $\mathcal{H}\Delta\mathcal{H}$-divergence $d_{\mathcal{H}\Delta\mathcal{H}}(P_X^s, P_X^t)$, while the complexity of $\mathcal{H}$ for prediction tasks on the source domain and target domain is assessed by the ideal joint risk $\lambda_\mathcal{H}$. Notably, the $\mathcal{H}\Delta\mathcal{H}$-divergence offers superior finite sample guarantees compared to other divergence measures and allows for non-asymptotic bounds. The non-asymptotic bound ensures that the bound holds a high probability, regardless of the sample size, which is particularly useful in applications where the sample size is small or limited. This means that with a finite sample size, the $\mathcal{H}\Delta\mathcal{H}$-divergence can provide a more accurate estimate of the difference between two probability distributions than other measures. 

\noindent\textbf{Theorem 1} (DA error bound (non-asymptotic)\cite{ben2010theory, wang2022generalizing}). Let $\mathcal{U}^s$ and $\mathcal{U}^t$ represent unlabeled samples of size $k$ from the two domains. The \textit{Vapnik-Chervonenkis} (VC) dimension \cite{vapnik1994measuring} of $\mathcal{H}$, denoted as $d$. For any $h \in \mathcal{H}$ and $\epsilon \in (0, 1)$, the following inequality holds with a probability of at least $1 - \epsilon$:

\begin{equation}\label{eq8}
        \theta^t(h) \leq \theta^s(h) + \hat d_{\mathcal{H}\Delta\mathcal{H}}(\mathcal{U}^s, \mathcal{U}^t) + \lambda_\mathcal{H} + 4\sqrt{\cfrac{2d\log(2k) + log(\frac{2}{\epsilon})}{k}}
\end{equation}
where $\hat d_{\mathcal{H}\Delta\mathcal{H}}(\mathcal{U}^s, \mathcal{U}^t)$ denotes the estimated difference of $d_{\mathcal{H}\Delta\mathcal{H}}(P_X^s, P_X^t)$ across two distinct domains based on limited data samples. DA error bound (non-asymptotic) delineates a concept wherein learning algorithms are tailored to generalize towards a target domain utilizing a finite set of samples derived from both the source domain and target domain during the training phase. This error bound quantitatively measures the variance in risk between the target domain and source domain, encapsulating the difference in average loss. Specifically, it quantifies the distance between the average loss in the target and source domains. The bound is non-asymptotic, which means that it is valid for any size of the sample set, and thus enables effective control of the generalization performance of the algorithm in practical applications.

\hlred{In DA, the discrepancy between the source domain distribution $P_X^s$ and the target domain distribution $P_X^t$ poses a significant challenge. The distribution difference in the above error bound, denoted as $d(P_X^s, P_X^t)$, can impede the generalization capability of models trained on the source domain when applied to the target domain. Nonetheless, although direct control over the distribution difference can be demanding, a potential approach is to train a representation function $g: \mathcal{X} \rightarrow \mathcal{Z}$ that maps the input data $x$ to a representation space $\mathcal{Z}$, aiming to minimize the discrepancy between the distribution of representations from the two domains. Consequently, this approach aims to align the representation distributions of both domains more closely, thereby enhancing model generalization across domains. This technique, known as \textit{DA based on domain-invariant representation} (DA-DIR), focuses on identifying a representation space $\mathcal{Z}$ where the distributions $P_{Z}^s$ and $P_{Z}^t$ from the source domain and target domain, respectively, are minimized. Achieving this involves employing methods such as adversarial training, discrepancy minimization, domain confusion loss, and transfer learning.}

DA emerges as a formidable technique for enhancing ML model performance in instances where the source domain and target domain exhibit divergent distributions, a prevalent challenge across numerous real-world contexts.

\subsection{Domain Generalization}
\begin{figure*}[htbp]
	\begin{center}
		\includegraphics[width=0.86\textwidth]{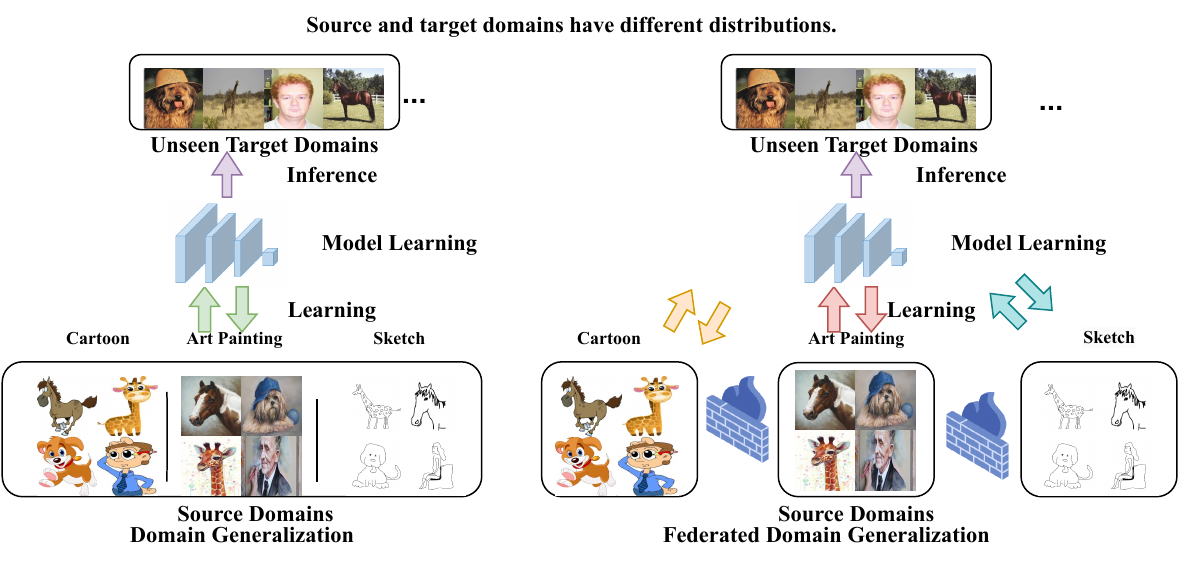}
		\caption{Comparison of domain generalization task (left) and FDG task (right).}
		\label{compare2}
	\end{center} 
\end{figure*}

\hlred{DA operates under the premise that either labeled or unlabeled data from the target domain is available for model adaptation, a condition not always met in real-world scenarios. Acquiring labeled data from the target domain can be prohibitively time-consuming, costly, or unfeasible due to ethical or legal restrictions. Moreover, the target domain's data distribution might remain elusive, complicating the application of conventional DA methods that depend on understanding this distribution. To tackle the issue, DG has been identified as a potent strategy to enhance the generalization capacity of ML models across novel and unseen domains. This is achieved by training models on a heterogeneous collection of datasets encompassing various domains. The DG methodology is depicted on the left side of Figure~\ref{compare2} and is concisely defined in \textbf{Definition 3}.}

\noindent\textbf{Definition 3} (Domain Generalization). Given $M$ source domains $\mathcal{S}_{source} = \{\mathcal{S}^i | i = 1, 2, \cdot\cdot\cdot, M\}$ where $\mathcal{S}^i = \{(x_j^i, y_j^i)\}_{j=1}^{n_i}$ represents the $i$-th domain, it is observed that the joint distributions between each pair of domains differ ($P_{XY}^i \neq P_{XY}^{\hat{\imath}}$, for $1 \leq i \neq \hat{\imath} \leq M$). The objective of DG is to develop a robust and generalizable labeling function $h: \mathcal{X} \rightarrow \mathcal{Y}$ solely utilizing the $M$ source domain data, which minimizes the prediction error on an unseen target domain $\mathcal{S}_{target}$, where \hlblue{$P_{XY}^{target} \neq P_{XY}^i$} for $i \in \{1,2, \ldots, M\}$:

\begin{equation}\label{eq9}
        \min_{h} \mathbb{E}_{(x,y) \in \mathcal{S}_{target}} \left[\ell(h(x), y)\right].
\end{equation}
where $\mathbb{E}$ denotes the expectation operator and $\ell(\cdot)$ represents the loss function used to disparity the difference between the predicted output $h(x)$ and the true output $y$. The ultimate aim is to identify the optimal function $h(\cdot)$ that reduces the expected loss on $\mathcal{S}_{target}$, thereby demonstrating the model’s capacity to effectively generalize to novel and unseen domains. 

In DG, the concept of an underlying hyper-distribution is pivotal for understanding the relationship between the source domain and the target domain. The probability distribution represents the joint distribution over all possible $(x, y)$ for both the source domain and target domain. Specifically, assume that all potential target distributions adhere to a hyper-distribution $\mathcal{P}$: $P_{XY}^t \sim \mathcal{P}$, and the same is true for the source distributions: $P_{XY}^1, P_{XY}^2,\cdot\cdot\cdot, P_{XY}^M \sim \mathcal{P}$. \hlred{This framework facilitates generalization to any target domain, regardless of its familiarity, by training a classifier that integrates domain-specific information $P_X$ into its input mechanism. Consequently, for a domain characterized by the distribution $P_{XY}$, predictions are structured as $y = h(P_X; x)$. For such a function $h(\cdot)$, its average risk across all conceivable target domains is determined by the subsequent formula:}

 \begin{equation}\label{eq10}
        \psi(h) = \min_{h} \mathbb{E}_{P_{XY} \sim \mathcal{P}}  \mathbb{E}_{(x,y) \sim P_{XY}} \left[\ell(h(P_X, x), y)\right].
\end{equation}

However, it is often infeasible to evaluate expectations exactly. Instead, these expectations can be approximated using finite domains/distributions adhering to a specified probability distribution $\mathcal{P}$, along with finite samples of $(x, y)$ drawn from each distribution. Given that $P_{XY}^1, P_{XY}^2, \cdot\cdot\cdot, P_{XY}^M \sim \mathcal{P}$, we can use the source domains and their corresponding supervised data to estimate the expectation. This is because the source domains and their associated data are assumed to be drawn from the same underlying distribution $\mathcal{P}$. As a result, the estimation of the average risk over all possible target domains based on the labeling function $h(\cdot)$ is as follows:
\begin{equation}\label{eq11}
        \hat \psi(h) = \cfrac{1}{M} \sum_{i=1}^{M}\cfrac{1}{n^i} \sum_{j=1}^{n^i} \ell\left(h(\mathcal{U}^i, x_j^i), y_j^i\right).
\end{equation}
where $M$ is the total number of domains, $n^i$ is the number of samples in domain $i$, the supervised dataset $\mathcal{U}^i = \{x_j^i | (x_j^i, y_j^i) \in \mathcal{S}^i\}$ represents the empirical estimation for the marginal distribution $P^i_X$ of the features in domain $i$.

\noindent\textbf{Theorem 2} (Domain generalization error bound \cite{albuquerque2019adversarial, wang2022generalizing}). Let $\gamma := \min_{\pi \in \Delta_M} d_\mathcal{H}\left(P_X^t, \sum_{i=1}^M \pi_i P_X^i\right)$ with minimizer $\pi^*$ represents the minimum distance between the target domain distribution $P_X^t$ and the convex hull $\Lambda$ formed by the source domain distributions $P_X^i$. $\Delta_M$ represents the $(M$-$1)$-dimensional simplex that approximates the target domain distribution $P_X^t$ with a convex hull $\Lambda$ of source domain distributions $P_X^i$. The minimizer $\pi^*$ corresponds to the weights that give the best approximation $P_X^* = \sum_{i=1}^M \pi_i^*P_X^i$ of $P_X^t$ within $\Lambda$, where $\pi$ denotes a normalized mixing weights. The parameter $\rho = sup_{P_X^{\prime}, P_X^{\prime\prime} \in \Lambda} d_\mathcal{H}(P_X^{\prime}, P_X^{\prime\prime})$ represents the diameter of the convex hull $\Lambda$. Then, the domain generalization error bound can be expressed as:
\begin{equation}\label{eq12}
        \theta^t(h) \leq \sum_{i=1}^M \pi_i^* \theta^i(h) + \frac{\gamma + \rho}{2} + \lambda_\mathcal{H},(P_X^t, P_X^*).
\end{equation}
where $\lambda_{\mathcal{H},(P_X^t, P_X^)}$ represents the ideal joint risk across the target domain $P_X^t$ and the domain with the best approximator distribution $P_X^*$. $(M$-$1)$-dimensional simplex is often used to represent the mixing weights of source domains for approximating the target domain distribution. Specifically, for $M$ source domain distributions, they can be viewed as $M$ points in Euclidean space, and these points can be connected to form an $(M$-$1)$-dimensional simplex. In this simplex, each point corresponds to a source domain distribution, and each edge corresponds to a mixing weight of a source domain distribution (i.e., a real number between 0 and 1). And the sum of mixing weights for each point is 1, representing their convex combination. Therefore, these mixing weights can be used to approximate the target domain distributions, thus achieving the purpose of DG. In practical implementation, some optimization algorithms are usually needed to determine the values of these mixing weights so that the approximated target domain distributions can be as close as possible to the actual one.

\hlred{Similar to DA, the theoretical framework for DG emphasizes the importance of leveraging domain-invariant representation techniques. These strategies aim to achieve two primary objectives: firstly, to minimize the risk across all source domains, as indicated by the initial term of the theoretical bound; and secondly, to diminish the discrepancies in the representation distribution between the source and target domains. The latter objective is quantitatively assessed using metrics $\gamma$ and $\rho$, which serve to measure the divergence between the representations of source and target domains. By effectively minimizing this error bound, the model is trained to excel in generalizing across diverse domains, thereby ensuring robust performance on unseen domains.}

To sum up, DG is a promising approach for developing models that adeptly navigate the challenges posed by distribution differences across various source domains. It achieves this by narrowing the distribution gap between the source domain and the target domain and fostering domain-invariant representations. This approach ensures that the models are well-equipped to excel in real-world scenarios, particularly when deployed on previously unseen target domains.

\subsection{Fedrated Domain Generalization}
Recently, the proliferation of IoT devices has created enormous amounts of diverse data from multiple domains, providing ideal conditions for DG research. The fundamental principle of DG is to access the multi-source distributions in the learning process. However, the sensitive data generated by IoT devices, such as healthcare and transportation data, are often subject to strict privacy regulations, making it difficult to implement traditional DG methods that rely on centralized data processing. To tackle the issue of training models on data from multiple domains and obtaining a robust and generalizable predictive function while preserving data privacy, FDG has emerged as a promising solution \cite{liu2021feddg}. FDG has many applications in areas such as healthcare, finance, and autonomous vehicles, where the data is often distributed across multiple devices or servers, and the privacy of the data is a critical concern. The process of the FDG task is illustrated on the right of Fig.~\ref{compare2} and summarized in \textbf{Definition 4}.

\noindent\textbf{Definition 4} (FDG). Given $M$ source domains  $\mathcal{S}_{source} = \{\mathcal{S}^i | i = 1, 2, \cdot\cdot\cdot, M\}$ involved in FL where $\mathcal{S}^i = \{(x_j^i, y_j^i)\}_{j=1}^{n_i}$ denotes the $i$-$th$ domain. The joint distributions between each pair of domains are different: $P_{XY}^i \ne P_{XY}^{\hat{\imath}}, 1 \leq i \ne \hat{\imath} \leq M$. The objective of FDG is to develop a robust and generalizable labeling function $h: \mathcal{X} \rightarrow \mathcal{Y}$ using only $M$ source domains data that can minimize the prediction error on unseen target domains $\mathcal{S}_{target}$ (where $P_{XY}^{target} \ne P_{XY}^i$ for $i \in \{1,2, \cdot\cdot\cdot, M\}$)) while each domain $i$ can only access its local data $\mathcal{S}^i$:
\begin{equation}\label{eq13}
        \min_{h} \mathbb{E}_{(x,y) \in \mathcal{S}_{target}} \left[l(h(x), y)\right].
\end{equation}
where $\mathbb{E}$ denotes the expectation operator and $l(\cdot)$ represents the loss function used to quantify the discrepancy between the predicted output $h(x)$ and the true output $y$. The primary goal is to identify an optimal function $h(\cdot)$ that minimizes the expected loss on $\mathcal{S}_{target}$, thereby indicating the model's proficiency in generalizing effectively to novel and unseen domains.

\begin{figure}[b]
	\begin{center}
		\includegraphics[height=4.5cm, width=6cm]{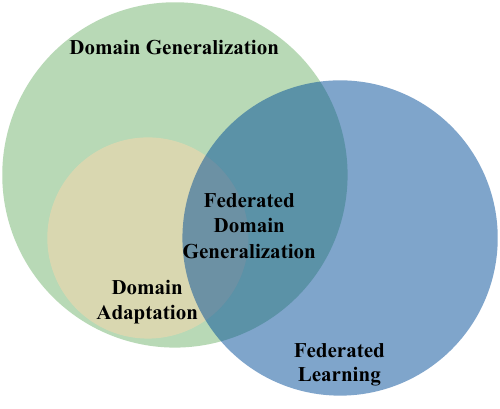}
		\caption{The scope of the proposed survey on federated domain generalization.}
		\label{scope}
	\end{center} 
\end{figure}

It is crucial to acknowledge that the distinctions among these concepts may overlap, with their scopes illustrated in Figure~\ref{scope}. The specific techniques and approaches used in each area can vary, but they all revolve around adapting or generalizing models to different domains or decentralized data settings. 

% \subsection{Challenges}
% FDG is a promising research area in ML that aims to develop models capable of generalizing well across multiple unseen domains while achieving privacy-preserving. However, the implementation of FDG encounters several challenges that need to be \hlyellow{solved} to achieve this goal. One of the primary difficulties faced in FDG is domain shift, which occurs when a model trained on data from one domain performs poorly on data from a different domain due to significant differences in statistical properties. Additionally, data heterogeneity, limited data availability, and privacy concerns pose significant dilemmas for FL. The distribution of data across multiple devices or organizations in FDG can result in significant heterogeneity, limited data for training, and privacy concerns due to the distribution of data. Moreover, high communication overhead can hinder efficient communication between domains during FL, which can further exacerbate these obstacles. 

% Collectively, by \hlyellow{considering} these challenges, FDG can advance the development of privacy-preserving, robust, and generalized ML models. This will not only benefit academic research but also have practical implications in real-world scenarios where the ability to generalize across multiple unseen domains is crucial.

\section{Methodologies: A Survey} \label{methods}
\begin{figure*}[htbp]
	\begin{center}
		\includegraphics[width=\textwidth]{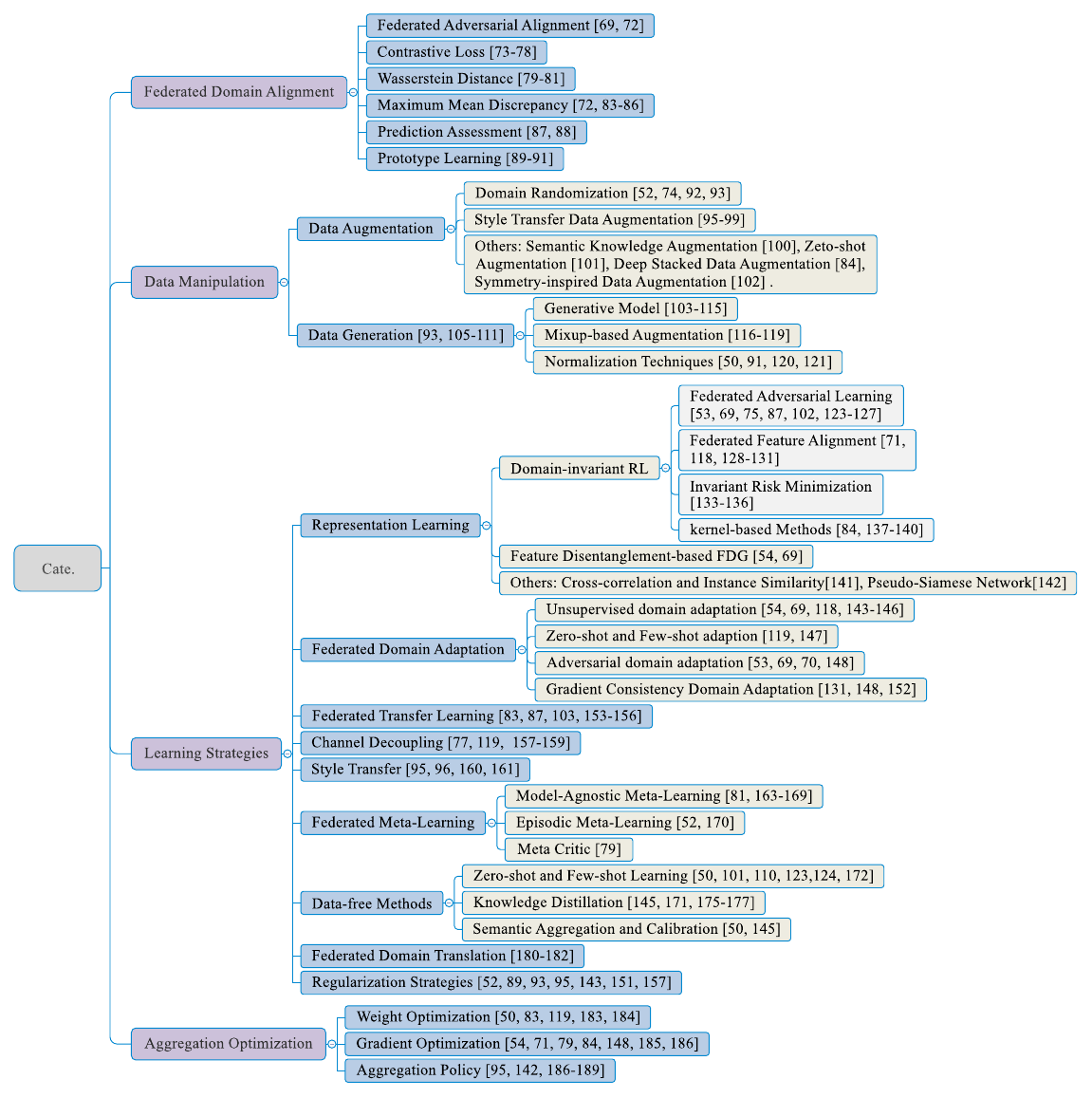}
		\caption{Categorization of FDG Methods.}
		\label{fig_Categorization}
	\end{center} 
\end{figure*}

\hlyellow{FDG methods have surfaced as a compelling approach to tackle domain shift and heterogeneity issues in FL, simultaneously ensuring data privacy.} Although its potential to enhance model generalization and robustness in federated settings has attracted attention in recent years, research in this area remains relatively sparse. \hlyellow{In this section, we aim to fill this gap by providing an extensive review of the current literature on FDG, categorizing it into distinct groups according to their methodologies and motivations, as depicted in Fig.~\ref{fig_Categorization}.}

\subsection{Federated Domain Alignment}
Federated domain alignment \cite{peng2019federated, zhao2023federated, sun2023feature} is of utmost importance in FDG as it enhances model generalization across multiple domains, promotes data efficiency, preserves data privacy, and enables the applicability of FL in real-world scenarios. The primary motivation of federated domain alignment is to bolster the generalization potential of models across various domains. By aligning the domains and harmonizing their data representations, federated domain alignment strives to mitigate domain shifts and enhance the performance of models on unseen domains. This aspect is crucial in practical applications, especially where data privacy and security are of utmost importance. Federated domain alignment facilitates efficient knowledge transfer while maintaining the decentralized and secure nature of sensitive data. The objective of alignment-based FDG methods, focusing on joint distribution, is articulated as follows:
\begin{equation}\label{Joint}
    \begin{aligned}
    P\left(X, Y\right) & = \sum_{i=1}^n P\left(D_i\right)P\left(X|D_i\right)P\left(Y|X, D_i\right)  \\
                       & = \sum_{i=1}^n P\left(D_i\right)P\left(Y|D_i\right)P\left(X|Y, D_i\right)
    \end{aligned}
\end{equation}
where $P\left(X, Y\right)$ represents the joint distribution of input features $X$ and corresponding labels $Y$ across all domains. $P\left(D_i\right)$ represents the distribution of domain $i$, indicating the probability of selecting samples from domain $i$. $P\left(X|D_i\right)$ represents the conditional distribution of input features $X$ given domain $i$, capturing the domain-specific characteristics of the data. $P\left(Y|X, D_i\right)$ represents the conditional distribution of labels $Y$ given input features X and domain $i$, describing the relationship between the features and labels in each domain.

$P\left(X, Y\right)$ sums over all domains $i$ to consider the contribution of each domain to the overall joint distribution $P(X, Y)$. The two equations represent two different decompositions of the joint distribution, highlighting different relationships between the variables. In FDG, our main focus is on aligning the marginal distributions of the source domains. Specifically, we aim to align $P(D_i)P(X|D_i)$ to reduce the distribution shift between domains. This is because we assume that the posterior probability $P(Y|X, D_i)$ remains relatively stable and has a smaller impact on the model. Therefore, domain alignment methods primarily concentrate on adjusting the distribution of input features to mitigate differences across domains and improve the model's generalization performance on unseen domains.

\paragraph{Adversarial Feature Alignment}
\hlblue{Adversarial feature alignment involves techniques that leverage adversarial learning to align feature distributions across different domains, thereby reducing domain discrepancy and enhancing model generalizability. In FDG, adversarial feature alignment plays a critical role in generating models that are not only accurate but also robust to variations in data distribution, ultimately leading to improved performance on unseen data.}

\hlblue{Peng \textit{et al.} \mbox{\cite{peng2019federated}} proposed federated adversarial alignment to align the feature distributions} between different domains in FL by dividing optimization into two independent steps: training domain-specific local feature extractors and training a global discriminator. The following steps are as follows: 

1) Local Feature Extractors: A feature extractor $G_s$ is trained for the source domain $\mathcal{S}^i$, and a feature extractor $G_t$ is trained for the target domain $\mathcal{S}^t$. 

2) Adversarial Domain Alignment: For each source-target domain pair $\left(\mathcal{S}_{s}, \mathcal{S}_{t}\right)$, an adversarial domain identifier $DI$ is trained to align the feature distributions of the two domains adversarially. Initially, the DI is fine-tuned to precisely identify the originating domain of the features. Following this, the generator models $\left(G_s, G_t\right)$ are trained to confuse $DI$ by producing features that make it tough for $DI$ to differentiate between the source and target domains. The objective of $DI_i$ is defined as follows:
\begin{equation}\label{eq14}
    \begin{aligned}
     L_{{adv}_{DI_i}} \left(\mathcal{S}_{s}, \mathcal{S}_{t}, G_s, G_t\right) = & -\mathbb{E}_{{s_s} \sim {\mathcal{S}_s}} \left[\log DI_i(G_s(\mathcal{S}_s)) \right] \\
    & - \mathbb{E}_{{s_t} \sim {\mathcal{S}_t}} \left[\log (1 - DI_i(G_t(\mathcal{S}_t))) \right]
    \end{aligned}
\end{equation}

In the second step, the objective for updating the generators $G_s$ and $G_t$ (denoted as $L_{{adv}_{G}}$), while \hlblue{keeping $L_{{adv}_{DI_i}}$} unchanged, is defined as follows:
\begin{equation}\label{eq15}
    \begin{aligned}
     L_{{adv}_G} \left(\mathcal{S}_s, \mathcal{S}_t, DI_i\right) = & -\mathbb{E}_{{s_s} \sim {\mathcal{S}_s}} \left[\log DI_i(G_s(\mathcal{S}_s)) \right] \\
    & - \mathbb{E}_{{s_t} \sim {\mathcal{S}_t}} \left[\log DI_i(G_t(\mathcal{S}_t)) \right]
    \end{aligned}
\end{equation}

\hlblue{Wang \textit{et al.}\mbox{\cite{wang2022federated}} developed a multi-client feature alignment framework employing adversarial learning, wherein a \textit{Generative Adversarial Network} (GAN) leverages class-wise information to generate adaptive reference distributions. This methodology, aimed at extracting DG features across diverse clients, effectively reduces distribution discrepancies. The training process is described as follows.} 

\hlblue{1) Feature Extraction and Generation: A feature extractor $F(\cdot)$ extracts real features by generating a representation $f^{s, i} = F(x^{s, i})$ for each source client $i$, where $x^{s, i}$ is the input data. Following this, a generator $G(\cdot)$ employs random noise $z$ and a one-hot vector $y$ to produce fake features $f_{fake}^{s, i} = G(z^{s, i}|y)$, maintaining consistency across clients by adhering to a shared, predefined distribution $p(f_{fake})$.} 

\hlblue{2) Discriminator Update Process: The discriminator $D(\cdot)$ is refined through the integration of both real and fake features, where real features serve as negative samples and fake ones as positive, employing a least-squared estimation approach within its loss function $L_{adv\_d}^i$ to bolster convergence and discriminative proficiency between feature types.}
% \begin{mdframed}[backgroundcolor=lightblue] 
\begin{equation}\label{advd}
L_{adv\_d}^i = \mathbb{E}_{z \backsim p(f_{fake}^{s,i})}[{D(f_{fake}^{s,i}|y)}^2] - \mathbb{E}_{x^{s,i} \backsim p(f^{s,i})}[{(1 -D(f^{s,i}|y))}^2]
\end{equation}
% \end{mdframed}

\hlblue{3) Generator and Feature Extractor Update Process: With the discriminator's parameters fixed post-training, the feature extractor and generator are updated through an adversarial process. The feature extractor aims to produce features that the discriminator perceives as positive (real), whereas the generator strives to create fake features that are indistinguishable from real ones to the discriminator. This adversarial interplay is captured in the loss functions $L_{adv\_f}^i$ and $L_{adv\_f}^i$ for the feature extractor and generator, respectively.}
% \begin{mdframed}[backgroundcolor=lightblue] 
\begin{equation}\label{advf}
L_{adv\_f}^i = \mathbb{E}_{x^{s,i} \backsim p(f^{s,i})}[{(1 -D(f^{s,i}|y))}^2]
\end{equation}
% \end{mdframed}
% \begin{mdframed}[backgroundcolor=lightblue] 
\begin{equation}\label{advg}
L_{adv\_g}^i = \mathbb{E}_{z \backsim p(f_{fake}^{s,i})}[{D(f_{fake}^{s,i}|y)}^2]
\end{equation}
% \end{mdframed}

\hlblue{4) Iterative Optimization: The training process iteratively optimizes the discriminator and generator until reaching a Nash equilibrium, indicative of the discriminator's diminished ability to distinguish between real and synthetic features.}

\hlblue{While both the above methodologies share the overarching goal of leveraging adversarial learning for domain generalization in FL, they differ in their specific techniques and the scope of their application. The former concentrated on domain-specific alignment, whereas the latter pursued a more generalized strategy for feature alignment across multiple clients.}

\paragraph{Minimizing Contrastive Loss}
The contrastive loss \cite{motiian2017unified, yoon2019generalizable} encourages similar representations for samples from the same domain and dissimilar representations for samples from different domains by maximizing agreement within positive pairs and minimizing agreement within negative pairs. The essence of minimizing contrastive loss is to sculpt a representation space where akin samples are drawn nearer, while disparate samples are distanced. The mathematical formulation of contrastive loss is delineated as follows:
\begin{equation}
    L_{Contrastive} = \frac{1}{2N} \sum_{i=1}^N y \dot{d}^2 + (1-y) max(margin-\dot{d}, 0)^2
\end{equation}
where $N$ is the number of sample pairs, \hlblue{$y$ is a binary label indicating whether a pair of samples are similar (positive pair, $y=1$) or dissimilar (negative pair, $y$=0)}, $\dot{d}$ is the distance between the representations of the paired samples, and $margin$ is a hyperparameter that defines the minimum desired distance between representations of dissimilar samples.

Li \textit{et al.} \cite{li2021model} proposed the MOON (model-contrastive federated learning) framework that leverages model-level contrastive learning to bridge the gap between global and local features and improve the performance of FL, surpassing other state-of-the-art algorithms. As an extension of \cite{li2021model}, an improved contrastive loss \cite{guo2022feddebias} is utilized by considering the projected features of pseudo-data as positive pairs and incorporating the projected local feature of both pseudo-data and local data as negative pairs. To enhance model invariance across domains, an alignment loss is deployed between the original sample and its hallucinated counterpart using a negative-free contrastive loss \cite{xu2023federated} at the logit level. \hlyellow{To tackle the underexplored issues of data heterogeneity and class imbalance in FL, a novel privacy-preserving framework, FedIIC\mbox{\cite{wu2022federated}}, has been proposed. This framework adeptly calibrates deep models to tackle imbalanced training through the use of intra-client and inter-client contrastive learning techniques, resulting in superior performance across both realistic and established scenarios.} Tan \textit{et al.} \cite{tan2022federated} proposed a lightweight framework called FedPCL (Federated Prototype-wise Contrastive Learning), enabling clients to jointly learn by fusing representations from multiple pre-trained models, improving each client's ability to leverage class-relevant information in a personalized manner while maintaining compact shared knowledge for efficient communication. \hlyellow{Liu \textit{et al.}\mbox{\cite{liu2023fedcl}} introduced a Federated Contrastive Learning (FedCL) approach that integrated contrastive learning with FL to enhance model generalization across heterogeneous medical data, demonstrating superior performance in medical image classification.}

To sum up, this approach helps to align the representations of different domains, reducing the domain shift and enhancing the model's ability to generalize across diverse domains.

\paragraph{Wasserstein Distance} 
The Wasserstein distance, also known as the Earth Mover's distance, is a measure of the dissimilarity between probability distributions. In FDG, the Wasserstein distance serves as a quantitative measure to evaluate the disparity between distributions across various domains. This approach facilitates the development of models with enhanced generalizability to previously unseen domains.

Chen \textit{et al.} \cite{chen2022learning} introduced a Wasserstein-based feature critic with meta-optimization into FL, which effectively improved the robustness and generalization capability of the proposed method. Nguyen \textit{et al.} \cite{nguyen2022generalization} proposed a Wasserstein distributionally robust optimization scheme called WAFL for FL, which is more general than related approaches and is robust to all adversarial distributions inside the Wasserstein ball. Lin \textit{et al.} \cite{lin2020collaborative} introduced a robust federated meta-learning framework that utilizes Wasserstein distance and Lagrangian relaxation to efficiently optimize the distance metric, enabling gradient-based methods to solve the problem effectively. 

In summary, the integration of the Wasserstein distance within the objective function of FDG methods facilitates the learning of domain-invariant representations, enabling improved generalization across federated domains. 
\paragraph{Minimizing Maximum Mean Discrepancy}
\textit{Maximum Mean Discrepancy} (MMD) \cite{gretton2012kernel} serves as a statistical metric for quantifying the differences in probability distributions across various domains. In FDG, MMD is employed to reduce the distributional shifts between domains, thereby facilitating domain alignment. The objective is to minimize MMD to foster domain-invariant representations, enabling the model to capture and leverage commonalities across domains while retaining unique domain features. The calculation of MMD is as follows:
\begin{equation}
    MMD^2\left(P, Q\right) = \left \| \frac{1}{n'} \sum_{j=1}^{n'} \upsilon\left(x_j\right) - \frac{1}{n''} \sum_{\hat{\jmath}=1}^{n''} \upsilon\left(y_{\hat{\jmath}}\right)\right \| ^2
\end{equation}
where $P$ and $Q$ are probability distributions of the feature representations of two domains, $n'$ and $n''$ are the number of samples in each domain, and $\upsilon(\cdot)$ is a feature mapping function that maps the samples to a high-dimensional space.

Chen \textit{et al.} \cite{chen2022federated} implemented D-WFA using the MMD distance between the source and target clients, which helps to reduce domain differences and improve the generalization capability. To sum up, we can effectively \hlyellow{solve} the domain shift problem and enhance the generalization capabilities of FL models by incorporating MMD in FDG. The proposed method \cite{tian2021privacy} enables the aggregated gradient to incorporate information from multiple domains by connecting gradient alignment with MMD, aiming for better generalization on unseen domains. Zhang \textit{et al.} \cite{zhang2022data2} employed MMD to measure and optimize the distance between different domains by calculating the squared distance between the kernel embeddings of the data representation distributions within the reproducing kernel Hilbert space, with unbiased empirical estimation used to reduce computation load. \hlyellow{Zhang \textit{et al.}\mbox{\cite{zhang2023blockchain}} introduced a blockchain-based decentralized federated transfer learning approach for collaborative machinery fault diagnosis. This method overcomes the issues of data quality and quantity by guaranteeing data security and privacy, thereby achieving high testing accuracies and presenting a viable solution for real-world industrial applications. Wang \textit{et al.}\mbox{\cite{wang2022federated}} developed a federated adversarial domain generalization network for machinery fault diagnosis, facilitating collaborative training between a central server and multiple clients to ensure data privacy. This approach resulted in significant performance improvements and showed potential for tasks related to data privacy-preserving generalization in diagnostics.}

Overall, minimizing the MMD facilitates FDG by encouraging feature representations from various domains to exhibit similar means in high-dimensional space, thereby reducing distributional discrepancies. This promotes the alignment of domains and improves the generalization capability of models across different domains.

\paragraph{Prediction Assessment}
\hlyellow{Prediction consistency and prediction disagreement represent two interconnected aspects of model evaluation in FDG, each playing a distinct role in enhancing the generalization capabilities of federated learning systems across decentralized domains.}

\hlyellow{\textbf{Prediction consistency} refers to the similarity in predictions made by different models trained across various clients or by a single model under different conditions for the same input. In FDG, consistency is indicative of a model's ability to learn domain-invariant features that are robust across unseen domains. High prediction consistency suggests that models are capturing the underlying patterns in the data that are common across domains, thereby improving the model's generalization to new and unseen domains.} Zhang \textit{et al.} \cite{zhang2021federated1} leveraged prediction consistency to enhance the generalization capability of the model. The inference results for the target domains are obtained by majority voting on the predictions made by multiple source domains. Considering the fact that these source domains have different parameters, prediction consistency generally improves the model's generalization capability while reducing the risk of overfitting. Thus, it is desirable to minimize variations in the inference results among the source domains. To achieve this, the loss function $L_p$ is defined as follows:
\begin{equation}\label{}
\begin{aligned}
    &\min L_p = \frac{1}{n^t M} \sum^{n^t}_{j=1} \sum^{M}_{i=1} 	\lVert I^{s,i}\left(F^t(x^t_j)\right) - \bar I\left(F^t(x^t_j)\right)\rVert, \\
    &\bar I\left(F^t(x^t_j)\right) = \frac{1}{M} \sum^{M}_{i=1} I^{s,i}  \left(F^t(x^t_j)\right)
\end{aligned}
\end{equation}
where $n^t$ represents the number of target client data samples, $M$ represents the number of source inference models, $F^t(x^t_j)$ represents the feature representation of the target domain data $x^t_j$ in the model, $I^{s, i}$ represents the inference module at the $i$-th source domain, $\bar I\left(F^t(x^t_j)\right)$ denotes the mean value of predictions obtained from the source inference models. 

\hlyellow{\textbf{Prediction disagreement} underscores the variations in outcomes generated by distinct models or a single model under varying conditions for the same input, revealing critical insights into the diversity and domain-specific characteristics of decentralized datasets. This divergence is instrumental in fostering the development of models with enhanced generalizability and robustness within FDG, thereby ensuring the aggregated model's efficacy across diverse unseen domains.} \hlyellow{Guo \textit{et al.}\mbox{\cite{guo2023out}} introduced FEDIIR, a novel approach that implicitly learns invariant relationships for OOD generalization in federated learning, leveraging prediction disagreement and inter-client gradient alignment to enhance model performance while adhering to privacy and communication efficiency principles.}

\hlyellow{While prediction consistency aims to ensure that the model learns generalizable features, prediction disagreement offers insights into model diversity and domain-specific biases. Together, they provide a comprehensive framework for evaluating and improving the generalization performance of federated models. By judiciously analyzing and leveraging both consistency and disagreement, FDG can effectively address the challenges posed by domain drift and data decentralization, leading to models that are both robust and highly generalizable across diverse domains.}

\paragraph{Prototype Learning}
\hlyellow{Prototype learning, a pivotal concept in ML, emphasizes identifying `prototypes'—representative samples within a dataset that encapsulates the diversity present across the data. In FDG, prototype learning entails each client pinpointing prototypes specific to their domain, while a global model synthesizes this data to discern shared prototypes, thereby uncovering commonalities across domains and concurrently safeguarding data privacy.}

\hlyellow{Huang \textit{et al.}\mbox{\cite{huang2023rethinking}} introduced Federated Prototypes Learning for addressing domain shift in FL by constructing cluster and unbiased prototypes to provide domain knowledge and a fair convergence target, demonstrating its effectiveness and efficiency on tasks. Liu \textit{et al.}\mbox{\cite{liu2023cross}} introduced PGCT, a novel prototype-guided cross-training mechanism for FL, designed to address data heterogeneity and knowledge forgetting by leveraging client-specific data prototypes for consistent representation learning and feature augmentation, demonstrating superior performance over existing methods. Yu \textit{et al.}\mbox{\cite{yu2023contrastive}} introduced a prototype-based contrastive loss to enhance class-wise alignment within the embedding space, thereby improving the model's generalization capabilities.}

\hlyellow{Leveraging prototype learning within FDG adeptly mitigates the issues posed by domain shift and privacy concerns. This approach enables the crafting of models that are not only adept at preserving privacy and efficiency but also excel in generalizing to novel, unseen domains.}

\subsection{Data Manipulation}
The performance of ML models is significantly influenced by the quality and quantity of the training data, especially in DG where the models are expected to generalize well to unseen domains. However, acquiring a large and diverse set of training data that covers all possible domains is often infeasible or expensive. In addition, the privacy concerns associated with sensitive data make it difficult to collect and store data in a centralized manner. In FDG, data manipulation techniques \hlyellow{have been devised to surmount the aforementioned difficulties, thereby generating more diverse and representative training datasets.} One such technique is data augmentation, which can be used to create additional training data by modifying the existing data through techniques such as rotation, scaling, and adding noise. Furthermore, adversarial data augmentation can be used to generate difficult examples that can help improve the model's robustness against domain shifts and adversarial attacks. 

The objective of data manipulation can be expressed mathematically as:
\begin{equation}\label{eq18}
    \min_{\omega} \sum_{i=1}^{M} \nu_i \mathcal{L}_i\left(\mathcal{\hat S}^i, \omega\right) + \xi R\left(\omega\right)
\end{equation}
where $M$ is the total number of domains, $\omega$ represents the model parameters to be learned, $\nu_i$ \hlblue{represents} the importance of domain $i$ in FL and is determined based on the quantity and quality of the domain $i$, $\mathcal{L}_i(\mathcal{\hat S}^i, \omega)$ measures the performance of the model on the domain $i$ and includes both real and synthetic data generated by data generation techniques. \hlblue{
The hyperparameter $\xi$ regulates the balance between fitting the manipulated data and model complexity, where a higher $\xi$ favors simplicity and may increase training error, while a lower $\xi$ permits greater complexity, enhancing fit but risking overfitting.} The regularization term $R(\omega)$ encourages the model to learn domain-invariant features that are less sensitive to domain shifts and can improve generalization performance. Depending on the specific techniques used and the formula for manipulating the data $\hat{\mathcal{S}^i}$, existing work on data manipulation can be broadly categorized into two main categories: data augmentation and data generation.

\subsubsection{Data augmentation-based FDG}
Data augmentation-based FDG involves applying various transformations to the existing training data, such as rotating, scaling, or adding noise, to create new training data that are similar to the original ones. This can help increase the diversity and robustness of the training data and improve the model's generalization performance to unseen domains. Here, the $\mathcal{\hat S}^i = \mathcal{S}^i \cup A^i$, where $\mathcal{S}^i$ is the local real dataset of domain $i$ and $A^i$ is the set of random perturbations used for data augmentation.

\paragraph{Domain randomization}
In FDG, domain randomization can be used to simulate the variability of the data across different clients by randomizing certain aspects of the data during training. This can help the model learn to be robust to domain shifts and better generalize to unseen clients.

Duan \textit{et al.} \cite{duan2019astraea} \hlblue{used} local domain randomization augmentation to create a globally balanced distribution. Guo \textit{et al.} \cite{guo2022feddebias} proposed FedDebias, a unified algorithm that leverages mean-based random noise augmentation that can reduce the learning bias on local features and augment model efficacy without introducing real data. Liu \textit{et al.} \cite{liu2021feddg} employed the data augmentation of random rotation, scaling, and flipping in FL tasks. Atwany \textit{et al.} \cite{atwany2022drgen} applied data augmentations including resizing, horizontal and vertical flipping, random grayscale, ColorJitter, random rotation, translation, and Gaussian blur to fundus images to improve model performance and intra-domain generalizability.

\paragraph{Style Transfer Data Augmentation}
Style transfer data augmentation \cite{yang2020fda} is a technique that leverages the principles of neural style transfer to generate new training examples by combining the content of one image with the style of another, enhancing the diversity and richness of the training dataset.

Shenaj \textit{et al.} \cite{shenaj2023learning} took full advantage of the source dataset during the pre-training stage with style transfer data augmentation, randomly loading the target styles in the source images to mimic the target distributions. Chen \textit{et al.} \cite{chen2023federated} proposed cross-client style transfer in FL involves each client computing and sharing their local styles with the central server, which can be either single-image styles or a local image style bank, to facilitate the data augmentation process and enhance diversity in the training data. \hlyellow{Georgiadis \textit{et al.}\mbox{\cite{georgiadis2022st}} introduced style transfer FL aimed at COVID-19 image segmentation, tackling data variability and privacy constraints. This method utilizes a denoising CycleGAN at each client node to enhance robustness and attain performance on par with centrally-trained models. Lewy \textit{et al.}\mbox{\cite{lewy2023statmix}} developed an augmentation process that partitions images into batches within each node, computes statistics from a randomly chosen image, and then applies these statistics to augment each image in the batch, effectively emulating a node-specific style transfer augmentation. Yan \textit{et al.}\mbox{\cite{yan2023simple}} presented FedRDN, an innovative and potent data augmentation approach designed to alleviate feature shifts in skewed feature distribution scenarios within FL. This approach showcases robust scalability and generalizability, providing a plug-and-play solution that seamlessly bolsters the efficacy of diverse FL models.}

% \paragraph{Symmetry-inspired Data Augmentation}
% \hlyellow{Utilizing the symmetrical nature of biological structures, Symmetry-inspired data augmentation enriches training data variety via horizontal and vertical image flips, thereby imbuing models with superior generalization capabilities for invariant features. It also increases data volume while preserving memory efficiency—essential for FDG under resource constraints.}

Apart from the data augmentation-based FDG mentioned above, there are other popular techniques for data augmentation. \textit{Zero-shot augmentation} generates diverse and realistic augmented samples by leveraging external knowledge or pre-trained models without requiring any additional labeled data, enhancing model generalization and performance. Hao \textit{et al.} \cite{hao2021towards} introduced Fed-ZDA, a FL system that leverages zero-shot data augmentation to generate pseudo-exemplars of unseen classes while ensuring privacy, aiming to enhance fairness and accuracy performance uniformity across clients in federated networks. \textit{Semantic knowledge augmentation} is a technique used in ML to improve model performance and generalization by leveraging external semantic knowledge to generate more accurate, diverse, and semantically consistent training data. Sun \textit{et al.} \cite{sun2022federated} enrich the semantic space by combining text embedding, Gaussian noise, and attribute labels during the training process, making the model
better adapt to the instance-level visual space and enhancing performance and generalization capabilities. \textit{Deep stacked Ddata augmentation}-based FDG leverages the power of deep learning, data augmentation, and FL techniques to enhance the generalization capability of models across all domains. By augmenting the data and aligning the gradients, the models become more robust and capable of handling unseen domains, thereby improving the performance and applicability of FL in real-world scenarios. Tian \textit{et al.} \cite{tian2021privacy} proposed a deep stacked transformation data augmentation approach (called BigAug) for generalizing models to unseen domains, achieving good generalization on several unseen datasets. \hlyellow{Yu \textit{et al.}\mbox{\cite{yu2023san}} proposed a novel data augmentation method, termed Symmetry-Inspired Data Augmentation, which notably expands the sample size while minimizing memory requirements, thus significantly advancing FDG within the realm of medical imaging. These methods have proven effective in creating diverse, domain-invariant training samples, thereby enabling efficient knowledge transfer and adaptation across disparate domains, all the while maintaining data privacy under the FL paradigm.}

\subsubsection{Data generation-based FDG}
Data generation-based FDG involves synthesizing new training samples from scratch using generative models like Variational Autoencoders or Generative Adversarial Networks. \hlblue{In essence, the generative models used in data generation-based FDG are designed to produce new samples that strike a balance between similarity and diversity. The similarity in statistical properties ensures that the new samples are representative of the source domain, facilitating effective learning. Meanwhile, the introduction of variations in content and appearance broadens the model's exposure to potential variations it might encounter in unseen target domains, thereby improving its generalization capabilities.} Data generation-based FDG can help overcome the limitations of limited and biased training data and improve the model's ability to generalize to new and unseen domains.
Here, the $\mathcal{\hat S}^i = \mathcal{S}^i \cup C^i$, where $\mathcal{S}^i$ is the local real dataset of domain $i$ and $C^i$ is the virtual domain dataset used for data generation. 
\paragraph{Generative model}
The generative model can be used to generate synthetic data that captures the common characteristics of the different domains, enabling better DG. The choice of generative model for FDG depends on factors such as the characteristics of the data from different domains, the availability of labeled data in each domain, and the desired level of DG. 

Zhang \textit{et al.} \cite{zhang2022data} proposed FedDA, a data augmentation method based on GAN for generating high-quality overlap samples by learning features from both overlap and non-overlap data, effectively expanding the available training data and enhancing the effect of data augmentation. \hlyellow{Guo \textit{et al.}\mbox{\cite{guo2023federated}} demonstrated the efficacy of ACGAN in managing data imbalance and enhancement tasks. Leveraging its mechanism, ACGAN presented a viable data conduit for preserving data privacy in federated transfer learning, marking a novel application for ACGAN.} To enable membership inference on other participants, Chen \textit{et al.} \cite{chen2020beyond} proposed a novel approach that utilizes locally deployed GANs to generate samples with all labels, thereby obtaining the data distribution necessary for the inference process. Zhang \cite{zhang2020gan} leveraged GANs to generate synthetic data with the same underlying distribution as the original dataset, guided by the target FL model's discriminator, thereby enriching the training dataset and improving the imitative capability of the generated images. \hlblue{Zhang \textit{et al.}\mbox{\cite{zhang2021feddpgan}} proposed DPGAN framework allows different hospitals to employ a privacy-preserving data augmentation method by utilizing distributed DPGAN models to generate high-quality training samples, addressing the issue of insufficient training data and subsequently achieving accurate detection using the ResNet model in FL.} 

Yan \textit{et al.} \cite{yan2020variation} synthesized samples by a privacy-preserving generative adversarial network, which solves the cross-client variation problem and protects privacy with wide applicability. Li \textit{et al.} \cite{li2022ifl} proposed improved FL-GAN to learn globally shared GAN models by aggregating locally trained generator updates with the maximum mean difference. Sariyildiz \textit{et al.} \cite{sariyildiz2019gradient}  built a generative model to generate training samples for unseen classes upon the WGAN \cite{xian2018feature, arjovsky2017wasserstein} that takes a combination of noise vectors and class embeddings as input, enabling the production of class-specific samples based on the information provided by the class embedding. Tang \textit{et al.} \cite{tang2022virtual} introduced virtual homogeneity learning, a method for handling data heterogeneity in FL, which generates a separable and homogeneous virtual dataset by utilizing StyleGAN \cite{karras2019style} from shared noise across clients to improve convergence speed and generalization performance while preserving privacy. \hlyellow{Zhou \textit{et al.}\mbox{\cite{zhou2023fedfa}}
developed FEDFA, an innovative FL algorithm that mitigates feature shift among clients by probabilistically augmenting local feature statistics based on global information from the entire federation, offering a robust solution to improve model performance in the presence of Non-IID data distributions.}

\paragraph{Mixup-based augmentation}
The models mentioned above can serve as starting points, and further customization or hybrid approaches may be necessary based on the specific requirements of the FL setting. Mixup-based augmentation is forced to learn features that are common across the clients by mixing samples from different clients during training, thereby improving its ability to generalize to new, unseen clients.

\hlyellow{Shin \textit{et al.}\mbox{\cite{shin2020xor}} introduced XorMixup, a privacy-preserving data augmentation technique based on XOR operations, designed to tackle the Non-IID data hurdle in FL. This approach involves collecting encoded data samples from various devices and decoding them exclusively with each device's data, ensuring privacy and enhancing data diversity.} Yoon \textit{et al.} \cite{yoon2021fedmix} proposed FedMix, a privacy-protected data augmentation technique, averaging local batches and subsequently applying Mixup in local iterations to generate augmented data. The proposed approach aims to protect the privacy of local data in FL scenarios while improving performance in Non-IID settings. Yao \textit{et al.} \cite{yao2022federated} generated a certain proxy to approximate the target distribution for aligning features from different domains via mixup-based augmentation. \cite{yangclient} allows the generation of diverse domains by mixing local and global feature statistics (MixIG) while keeping data private, where MixIG is constructed by randomly interpolating instance and global statistics.

\subsubsection{Normalization Techniques}
\hlyellow{Normalization techniques unify the distribution of data across features or datasets, enhancing optimization efficiency and mitigating issues like gradient disappearance or amplification. Within FDG, these techniques are crucial for reconciling data heterogeneity, thus aiding in the development of consistent and generalizable attributes within robust global models.}

\hlyellow{Yu \textit{et al.}\mbox{\cite{yu2023contrastive}} developed an enhanced instance normalization module designed to concentrate on task-relevant features while minimizing domain-specific information, thereby boosting the discriminative and generalization capabilities of the local model. Li \textit{et al.}\mbox{\cite{li2021fedbn}} introduced FedBN, leveraging local batch normalization to address feature shift, outperforms traditional FedAvg and advanced FedProx for non-i.i.d. data in extensive testing. Zhu \textit{et al.}\mbox{\cite{zhu2023mla}} developed a Batch-Instance Style Normalization (BIN) block tailored for FL to address the domain gap attributed to stylistic variances. This BIN block is integrated with the segmentation backbone network, forming BIN-Net, which proficiently learns intra-domain features and concurrently neutralizes inter-domain style disparities, all without necessitating data access from other centers. Yuan \textit{et al.}\mbox{\cite{yuan2023collaborative}} employed hybrid batch-instance normalization and collaboration of frozen classifiers to improve the generalization of feature extractors. Yu \textit{et al.}\mbox{\cite{yu2023san}} proposed a masked adaptive instance normalization technique, addressing the challenge of generalizing models to unseen sites by minimizing inter-site discrepancies and standardizing input images from different sites into a site-unrelated style.} 

By incorporating data generation techniques in FDG, models can benefit from increased diversity and representativeness of the training data. This helps in capturing the shared features and reducing the domain-specific differences, ultimately leading to improved generalization performance across domains.

\subsection{Learning Strategies}
In addition to data manipulation, the field of FDG has garnered considerable attention within the broader ML paradigm. FDG research can be classified into several distinct approaches, including representation learning-based FDG, federated DA-based FDG, federated transfer learning-based FDG, federated adversarial learning-based FDG, style transfer-based FDG, federated meta-learning-based FDG, and other strategies. These approaches are devised to tackle the issue of training models that can effectively generalize to unseen domains by employing various techniques and principles.

\subsubsection{Representation Learning}
Representation learning \cite{bengio2013representation} has long been a focal point in ML, playing a pivotal role in the success of DG. Representation learning in FDG follows the same principles as representation learning in general but is tailored to the FL setting. The basic principle involves training a neural network to learn a mapping function that transforms the input data from each domain into a common representation space. The objective function used in representation learning for FDG typically includes components that encourage domain-invariant representations while preserving the utility of the learned representation for downstream tasks. The goal is to acquire domain-invariant representations that encapsulate common features across various federated domains, facilitating models to generalize effectively to unseen domains. In FDG, we decompose the prediction function $w$ as $w = u \circ z$, where $z$ represents the representation learning function and $u$ represents the classifier function. Mathematically, it can be formulated as:
\begin{equation}\label{eq19}
\min_{u, z} \sum_{s \in \mathcal{S}} \left[\mathbb{E}_{\left(x,y\right) \sim P\left(s\right)} \mathbb{L}_t(u(z(x)), y) + \chi \mathbb{L}_d(z(x), D)\right]
\end{equation}
where $\mathcal{S}$ represents the set of source domains, $\mathbb{E}_{\left(x,y\right) \sim P\left(s\right)}$ represents the expectation over the data samples $\left(x, y\right)$ from a specific source domain $s \in \mathcal{S}$. $\mathbb{L}_t\left(\cdot\right)$ denotes the task-specific loss function, which measures the discrepancy between the predicted output of the representation function $u\left(z\left(x\right)\right)$ and the true label $y$ for a given input $x$. $\mathbb{L}_d\left(\cdot\right)$ represents the domain discrepancy loss function, which quantifies the difference between the representations $z\left(x\right)$ and the domain labels $D$. The domain discrepancy loss function can be defined using various approaches, such as maximum mean discrepancy, adversarial DA, or adversarial neural networks, depending on the specific method used for FDG. This loss encourages the learned representations to be domain-invariant, capturing shared knowledge across different source domains. The hyperparameter $\chi$ controls the trade-off between the task loss and the domain discrepancy loss, determining the relative importance of task performance and domain invariance in the learned representations.

\paragraph{Domain-invariant representation-based FDG}
Domain-invariant representation-based FDG refers to the problem of learning models that can generalize well to unseen domains in the FL setting, while also capturing domain-invariant representations.  \\
\indent \textbf{\textit{Federated adversarial learning:}} Federated domain adversarial learning combines the concepts of FL and adversarial training to train a global model on multiple source domains. By incorporating a domain classifier, the model learns domain-invariant representations that generalize well across domains.

Xu \textit{et al.} \cite{xu2023federated} outlined a novel federated adversarial domain hallucination (FADH) learning framework for FDG, making the final model more robust to unseen domain shifts. Zhang \textit{et al.} \cite{zhang2023federated} 
introduced FedADG, employing federated adversarial learning to harmonize distributions across diverse source domains. This method yields a universal feature representation with strong generalization capabilities across unfamiliar target domains while safeguarding local data privacy. Peng \textit{et al.} \cite{peng2019federated} \hlyellow{extended} adversarial adaptation techniques to the constraints of the FL setting and \hlyellow{leveraged} feature disentanglement to bolster knowledge transfer. Zhang \textit{et al.} \cite{zhang2021federated1} \hlyellow{introduced} deep adversarial networks as a solution to effectively bridge the gap between domain distributions while maintaining data privacy. Micaelli \textit{et al.} \cite{micaelli2019zero} developed a training approach in which a student network learns to align its predictions with those of a teacher network, solely relying on an adversarial generator to discover images where the student exhibits poor alignment with the teacher and utilizing them for student training, without relying on any data or metadata. Zhang \textit{et al.} \cite{zhang2022fedzkt} developed a knowledge-agnostic approach for on-device knowledge transfer by adversarially training a generative model with the global model to obtain higher accuracy and better generalization performance. Dalmaz \textit{et al.} \cite{dalmaz2022specificity} introduced a novel specificity-preserving FL technique for MRI contrast translation. This approach leveraged an adversarial model to dynamically normalize feature maps across the generator based on site-specific latent variables. \hlyellow{Kang \textit{et al.}\mbox{\cite{kang2022privacy}} introduced a fine-grained adversarial domain adaptation approach aimed at minimizing feature dimensionality, improving model interpretability, and enabling the acquisition of domain-invariant features. Zhang \textit{et al.}\mbox{\cite{zhang2024multi}} developed a Multi-hop Graph Pooling Adversarial Network tailored for cross-domain Remaining Useful Life prediction, addressing the domain drift among varied clients by accommodating the diverse data distribution. Yu \textit{et al.}\mbox{\cite{yu2023san}} utilized a gradient reversal layer within the U-net encoder to foster domain-invariant representations, bolstering model generalization.}

In summary, federated adversarial learning contributes to improved generalization performance while ensuring privacy preservation and data security in the FL setting.

\indent \textbf{\textit{Federated feature alignment:}} \hlyellow{
Federated feature alignment, employed in FDG, tackles the issue of domain shift across varied domains within an FL environment.} It aims to align the feature representations of data samples from different domains to improve the generalization capability of the federated model.

Nguyen \textit{et al.} \cite{nguyen2022fedsr} presented FedSR, a novel representation learning framework for FL, which enables domain generalization while preserving privacy and decentralization, and incorporates CMI and L2R regularization techniques to improve generalization performance. To align features \cite{yao2022federated} from two domains in FL without direct access to target data, a proxy \hlyellow{was} constructed using a reweighted mixup of server data, which effectively improves model generalization, robustness, and stability, and further sampling \hlyellow{was} performed based on client data density using Gaussian Mixture Models (GMM) to create a proxy set. Zhang \textit{et al.} \cite{Zhang_2021_ICCV} proposed the FedUFO method, an adversary module is introduced to minimize feature divergence among different clients, while consensus losses were employed to mitigate inconsistencies in optimization objectives, enabling unified feature learning and alignment in Non-IID FL. Sun \textit{et al.} \cite{sun2023feature} proposed FedKA, a novel FDG method that utilizes feature distribution matching and a federated voting mechanism to align domain-specific features and generate pseudo-labels for global model fine-tuning, enabling domain-invariant learning in the presence of unknown client data. To learn feature representations from raw input data, Kang \textit{et al.} \cite{kang2020fedmvt} utilized two neural networks for each party, where one network captures weakly shared representations and the other learns domain-specific representations and three loss terms are proposed to enforce the desired learning objectives. \hlyellow{Wang \textit{et al.}\mbox{\cite{wang2023trfeddis}} presented TrFedDis, a trailblazing network that advanced FL through feature disentangling and uncertainty-aware decision fusion, thereby boosting both efficacy and dependability in Non-IID domain features.}

\indent \textbf{\textit{Invariant risk minimization (IRM):}} IRM \cite{arjovsky2019invariant} is an innovative learning paradigm that aims to estimate nonlinear, invariant, and causal predictors from multiple training environments to achieve robust generalization to OOD data. In deep learning, IRM consists of a feature encoder ($\mathfrak{g}$) and a classifier ($\mathfrak{h}$). Here, domain $i$ $\mathcal{S}^i = \{(x_j^i, y_j^i)\}_{j=1}^{n_i}$ is collected under multiple source domains $i \in I$. The formulation of IRM is to minimize the loss function:
\begin{equation}
\begin{aligned}
     & \min \limits_{\mathfrak{h},\mathfrak{g}} \sum_i \mathcal{R}^i\left(\mathfrak{h} \circ \mathfrak{g}\right), \\
     & \text{subject to:}\quad \mathfrak{h} \in \mathop{\arg\min}\limits_{\mathfrak{h}'} \mathcal{R}^i\left(\mathfrak{h}' \circ \mathfrak{g}\right), \text{for all}\ i \in I.
\end{aligned}
\end{equation}
where $\mathcal{R}^i(f) := \mathbb{E}_{x^i, y^i}{[l(f(x^i), y^i)]}$

The authors in \cite{arjovsky2019invariant} treat the classifier $\mathfrak{h}$ as a fixed scalar, resulting in the optimization problem formulation as follows.

Francis \textit{et al.} \cite{francis2021towards} utilized IRM to learn invariant predictors across client domains, achieving optimal empirical risk on all participating clients. The central objective of the FedGen \cite{venkateswaran2023fedgen} framework was to learn invariant predictors by aggregating local models and masks from participating clients, utilizing a penalty term similar to IRM as a regularizer to discourage overfitting to specific data distributions and to encourage generalization across distributions. Zhang \textit{et al.} \cite{zhang2021adaptive} introduced distributionally robust neural networks, domain adversarial neural networks, and MMD feature learning, as well as referencing previous work on invariance methods such as correlation alignment and IRM, providing a comprehensive evaluation of these approaches. Luca \textit{et al.} \cite{de2022mitigating} proposed a federated version of IRM that follows a similar procedure to FedAvg, but differs from other federated IRM approaches (such as CausalFed and CausalFedGSD) that involves data communication either through intermediate representations or a shared subset of data, making our Fed-IRM distinct in terms of data sharing. 
% \hlyellow{Zhao \textit{et al.}\mbox{\cite{zhao2023federatedfault}} presented the development of an integrated edge-cloud FL framework designed to mitigate data privacy concerns within the Industrial IoT. This introduced a domain-agnostic fault diagnosis model, developed through a two-stage training mechanism, which exhibits notable generalization capabilities and ensures data privacy.}

Overall, IRM could be used in FDG to solve the problem of domain shift and ensure model robustness across different domains, which aims to learn representations that are invariant to domain-specific variations while still being informative for the task at hand.

\indent \textbf{\textit{Kernel-based methods:}} 
Kernel-based methods provide a powerful framework to address domain shifts and improve generalization performance for FDG. These methods can capture complex relationships and align distributions between different domains by leveraging kernel functions and operating in high-dimensional feature spaces, leading to more robust and transferable models in FL settings.

Tian \textit{et al.} \cite{tian2021privacy} developed a novel DG method based on gradient aggregation, leveraging the gradient as a kernel mean embedding to align distributions across domains and improve model generalization capability under privacy constraints, achieving better performance and generalization capability under the privacy-preserving condition. \hlyellow{Huang \textit{et al.}\mbox{\cite{huang2021fl}} introduced the FL neural tangent kernel, a novel analytical framework for overparameterized ReLU neural networks trained in FL, which theoretically converged at a linear rate to a global optimum and demonstrated promising generalization capabilities.} Different distributions of observed data from multiple sources are modeled separately, and a kernel-based method was used to adaptively learn their similarities \cite{vo2022adaptive}, resulting in an adaptive factor that measures the similarity between the distributions. In the collaborative learning setting among distributed clients facilitated by a central server, Salgia \textit{et al.} \cite{salgia2023collaborative} introduced a kernel-based bandit algorithm using surrogate Gaussian process models, achieving near-optimal regret performance, and demonstrated the effectiveness of using sparse approximations to reduce communication overhead among clients. Hong \textit{et al.} \cite{hong2021communication} proposed eM-KOFL and pM-KOFL algorithms for Online FL achieve near-optimal performance while minimizing communication costs, with pM-KOFL demonstrating comparable performance to vM-KOFL (or eM-KOFL) across different online learning tasks.  

\paragraph{Feature disentanglement-based FDG }
Feature disentanglement is a widely adopted strategy for mitigating challenges related to domain shift and negative transfer when handling different domains, aiming to separate the domain-invariant features and domain-specific features from training samples\cite{kouw2019review, wilson2020survey}. In FDG, feature disentanglement-based methods extend the principles of feature disentanglement to the FL setting and aim to learn disentangled representations by leveraging the distributed data from multiple source domains while ensuring privacy preservation and data locality.

\hlyellow{Peng \textit{et al.}\mbox{\cite{peng2019federated}} significantly advanced the field by adapting adversarial adaptation techniques to the specific constraints of FL. This adaptation, which utilized feature disentanglement, successfully tackled the dilemmas associated with domain shift while preserving privacy.} Wu \textit{et al.} \cite{wu2021collaborative} optimized domain-invariant feature extractors for central aggregation and domain-specific classifiers for central ensembling, allowing for selective knowledge disentanglement between domain-invariant feature representations and domain-specific classification information. 

By disentangling the features and capturing the domain-invariant factors, feature disentanglement-based FDG methods enable the trained model to generalize well to unseen domains by leveraging the shared knowledge across domains. \hlyellow{Such strategies are instrumental in overcoming domain shift problems and enhancing model transferability within FL contexts.}

% \paragraph{Cross-correlation and Instance Similarity}
% \hlyellow{Huang \textit{et al.}\mbox{\cite{huang2023generalizable}} introduced FCCL+, a novel approach to FL that addresses model heterogeneity and catastrophic forgetting by leveraging cross-correlation and instance similarity with non-target distillation, demonstrating enhanced discriminability and generalization across diverse settings.}

% \paragraph{Gradient Reversal Layer}
% \hlyellow{The Gradient Reversal Layer serves as a domain invariance mechanism within deep learning architectures. In the realm of FDG, the Gradient Reversal Layer proves instrumental in fostering feature representations that are universally applicable across diverse clients, mitigating bias induced by data distribution disparities.}

\hlyellow{Beyond the representation learning-based FDG approaches previously discussed, several other techniques have garnered attention. Huang \textit{et al.}\mbox{\cite{huang2023generalizable}} introduced FCCL+, an innovative FL strategy that mitigates model heterogeneity and catastrophic forgetting through the use of cross-correlation and instance similarity alongside non-target distillation, thereby enhancing discriminability and generalization across varied environments. Song \textit{et al.}\mbox{\cite{song2024federated}} implemented a Pseudo-Siamese Network within local clients to assess discrepancies between client-specific and global models, improving the delineation of the feature space for fault diagnosis models. This method ensures that the feature representations learned do not excessively conform to the idiosyncrasies of the local domains, thus preserving efficacy in novel domains.}

\subsubsection{Federated Domain Adaptation}
Federated domain adaptation (FDA) is an approach that synergistically combines FL and DA techniques to mitigate the problem of domain shift in a decentralized setting, thereby improving model performance when source and target domains exhibit dissimilar distributions or come from distinct domains.
\paragraph{Unsupervised domain adaptation}
Unsupervised domain adaptation (UDA) is an ML technique \hlyellow{designed to adapt models from a source domain, where labeled data is available, to a target domain that lacks labeled data.} In FDG, UDA techniques aim to learn domain-invariant representations that capture shared knowledge across domains while reducing domain-specific differences. The goal is to bridge the domain gap and enable models to generalize well to unseen domains without access to labeled data in those domains. 

Peng \textit{et al.} \cite{peng2019federated} proposed an unsupervised FDA method that aligns learned representations across different domains with the data distribution of target domains to solve domain shift. Wu \textit{et al.} \cite{wu2021collaborative} proposed a new approach called Collaborative Optimization and Aggregation (COPA) by employing hybrid batch-instance normalization and collaboration of frozen classifiers to optimize a generalized target model for decentralized DG and UDA. Zhuang \textit{et al.} \cite{zhuang2022federated} introduced FedFR, a method for UDA in face recognition, which combines clustering-based DA and FL to improve performance on a target domain with different data distributions from the source domain. Zhou \textit{et al.} \cite{zhou2022source} introduced STU-KD, a scheme for privacy-preserving adaptation of a compact model to edge devices, which combines DA techniques with knowledge distillation to efficiently transfer knowledge from a large target model to the compact model. The proposed DualAdapt method \cite{yao2022federated} is an FDA technique that enables the model to adapt to the target domains while preserving knowledge learned from the source domains, achieving high accuracy while minimizing communication costs and computational resources required on client devices. Niu \textit{et al.} \cite{niu2023mckd} introduced a novel federated UDA strategy termed knowledge filter to adapt the central model to the target data when unlabeled target data is available. \hlyellow{Qiu \textit{et al.}\mbox{\cite{qiu2023federated}} proposed \textit{Federated Semi-Supervised Learning} (FSSL) method, employing federated pseudo-labeling for distributed medical image domains, enhances FDG by effectively utilizing unlabeled data.}

\hlyellow{In summary, UDA in FDG is pivotal for mitigating domain shift issues and boosting model transferability across federated domains. This enables models to achieve strong performance in unfamiliar domains, even with limited labeled data.}

\paragraph{Zero-shot and few-shot adaption}
\hlyellow{Zero-shot and few-shot adaptation techniques tackle data scarcity and bolster model generalization. These methods empower models to function effectively in tasks or domains with minimal or no explicit training data, thereby enhancing adaptability to novel scenarios.} In FDG, zero-shot and few-shot adaptation techniques aim to enhance model generalization across distributed domains, 

Yang \textit{et al.} \cite{yangclient} introduced zero-shot adaptation with estimated statistics to solve the difficulties of FDG, where the zero-shot adapter facilitates the learned global model in directly bridging a significant domain gap between seen and unseen clients during inference. \hlyellow{Huang \textit{et al.}\mbox{\cite{huang2022few}} tackled the issues of few-shot, model-agnostic FL by unveiling a framework that utilizes public datasets to enhance performance in the face of limited private data. This approach effectively navigates issues like inconsistent labels and domain gaps by employing model-agnostic FL techniques and latent embedding adaptation.}

\hlyellow{Zero-shot and few-shot adaptation techniques in FDG tackle domain shifts and privacy concerns, thereby enhancing the performance of FL in real-world scenarios characterized by limited data access and distribution constraints.}

\paragraph{Adversarial domain adaptation}
Adversarial domain adaptation (ADA) can also be applied in FDG, where the goal is to learn models that can generalize well across multiple domains without explicitly sharing data. In this scenario, each domain has its data distribution and potentially different labeling conventions or data biases.

\hlyellow{Zeng \textit{et al.}\mbox{\cite{zeng2022gradient}} introduced a novel strategy, one-common-source ADA, aimed at addressing domain shifts in each target domain (private data) by leveraging a common source domain (public data). This strategy includes the use of ADA with gradient matching loss for pre-training encoders. Zhao \textit{et al.}\mbox{\cite{zhao2023federated}} developed a federated multi-source DA method that integrates transfer learning with FL for machinery fault diagnosis, prioritizing data privacy. This method employed federated feature alignment to reduce feature distribution discrepancies and introduced a joint voting mechanism for refining the global model using pseudo-labeled samples. Consequently, this method achieves precise identification of target data while ensuring the protection of data privacy. Peng \textit{et al.}\mbox{\cite{peng2019federated}} pioneered an approach to FDA by synchronizing the learned representations among distributed nodes with the target node's data distribution. This method leverages adversarial adaptation techniques, a dynamic attention mechanism, and feature disentanglement, thereby facilitating knowledge transfer and confronting the domain shift task inherent in FL. Zhang \textit{et al.}\mbox{\cite{zhang2023federated}} proposed FedADG, a federated adversarial domain generalization strategy that harmonizes distributions across various domains through a reference distribution. This method guarantees a universal feature representation that generalizes well, simultaneously safeguarding local data privacy.} 

% Zhao \textit{et al.}\mbox{\cite{zhao2023federatedfault}} presented PrADA, an innovative federated adversarial domain adaptation method, addressing the insufficient sample and feature challenges in cross-silo federated domain adaptation. The approach advances model interpretability and transferability by extracting semantically meaningful, high-order features from categorized feature groups within a secure framework.
By incorporating ADA into FDG, models can learn domain-invariant representations that capture common features across multiple domains, enabling the final global model to generalize well to unseen domains, even in the absence of explicit data sharing.

\paragraph{Gradient Consistency Domain Adaptation}
\hlblue{Gradient consistency\mbox{\cite{hiasa2018cross, Ma_2023_CVPR}} pertains to ensuring uniformity or alignment in the gradients—direction and magnitude of model learning—across diverse data domains during model parameter optimization. The aim of gradient consistency in FDG is to ensure that, despite the distributed and heterogeneous data environment, models can learn a unified representation that generalizes across all domains by aligning gradients.}

\hlblue{Zeng \textit{et al.}\mbox{\cite{zeng2022gradient}} proposed a method called gradient matching FDA (GM-FDA), which aims to minimize domain discrepancy by utilizing a public image dataset and training resilient local federated models specifically for target domains. In\mbox{\cite{zeng2022gradient}}, the gradient matching loss  $L_{GM}$ is expressed as the expected cosine distance between $\dot g_s(\vartheta)$ and $\dot g_t(\vartheta)$, considering all possible values of $\vartheta$:}
% \begin{mdframed}[backgroundcolor=lightblue] 
\begin{equation}\label{eq20}
L_{GM} = \underset{\clap{\scriptsize $\vartheta$}}{\mathbb{E}} \bigg [1 - \frac{\dot g_s\left(\vartheta\right)^T \dot g_t\left(\vartheta\right)}{\lVert \dot g_s\left(\vartheta\right)\rVert_2 \times \lVert \dot g_t\left(\vartheta\right)\rVert_2}\bigg ]
\end{equation}
% \end{mdframed}
\hlblue{where $\dot g_s(\vartheta)$ represents the expected gradient vector from the source domain, $\dot g_t(\vartheta)$ represents the expected gradient vector from the target domain, and the gradients are computed concerning the compatibility model parameters $\vartheta$. $\lVert \cdot \rVert$ represents the Euclidean norm and $\underset{\clap{\scriptsize $\vartheta$}}{\mathbb{E}}$ denotes the expectation over the compatibility model parameters $\vartheta$. Minimizing the gradient matching loss $L_{GM}$ aims to align the gradient vectors between the source and target domains, facilitating domain-invariant representation learning and improving the generalization capability across federated domains.}

\hlblue{Zhu \textit{et al.}\mbox{\cite{zhu2022fedilc}} developed an approach to mitigate the issue of domain shift in FL by utilizing gradient covariance and the geometric mean of Hessians. This strategy effectively captures the consistency across and within silos. Compared to alternative domain adaptation methods, this approach demonstrates superior performance. The gradient covariance between two random variables $X$ and $Y$ is defined as follows:}
% \begin{mdframed}[backgroundcolor=lightblue] 
    \begin{equation}
    Cov(X, Y) = E[(X-E[X])(Y-E[Y])]
    \end{equation}    
% \end{mdframed}
\hlblue{where $E[X]$ is the expected value (mean) of $X$, and $E[Y]$ is the expected value of $Y$. The expected value operator $E[\cdot]$ calculates the mean of its argument. By leveraging gradient covariance, it's possible to facilitate the integration of learning from multiple tasks or domains by aligning gradient directions, thereby improving model generalization and performance on a wider range of tasks.}

% \hlblue{Jeong \textit{et al.}\mbox{\cite{jeong2023federated}} proposed a novel algorithmic framework called \textit{Federated Gradient Matching Pursuit} (FedGradMP) to solve the sparsity-constrained minimization problem in the FL setting.} 
    
% \hlblue{Step 1 (Random Mini-Batch Selection): On each client, a mini-batch dataset is randomly selected. \\
% \indent Step 2 (Stochastic Gradient Computation): For the selected mini-batch, compute its stochastic gradient. \\
% \indent Step 3 (Subspace Merging): Merge the subspace associated with the previously estimated local model with the closest subspace of dimension to the stochastic gradient.  \\
% \indent Step 4 (Minimization Problem Solving): Solve the minimization problem for the local objective function over the merged subspace.  \\
% \indent Step 5 (Identifying the Closest Subspace): Determine the closest subspace of dimension to the solution from Step 4. Then project onto this subspace for the next local update.\\
% \indent Step 6 (Aggregation and Global Update): After all clients have completed their local iterations, the server aggregates and updates the global model.}

\hlblue{Wei \textit{et al.}\mbox{\cite{wei2024multi}} introduced a multi-source collaborative gradient discrepancy minimization method for FDG. This approach facilitates intra-domain and inter-domain gradient matching across various source domains, with the objective of developing a model that remains unbiased towards specific domains and exhibits superior performance on unseen target domains. Specifically, the method employs intra-domain gradient matching to synchronize the gradients of classifiers for original and augmented images from isolated source domains. This strategy aims to steer local models towards capturing the essential semantic content present in both types of images.}

% Wei \textit{et al.}\mbox{\cite{wei2024multi}} proposed a multi-source collaborative gradient discrepancy minimization method for FDG, which conducts intra-domain and inter-domain gradient matching among different source domains, aiming to learn a model that is not biased by specific domains and performs well on unseen target domains, while also addressing the challenges of data isolation due to privacy concerns. Specifically, intra-domain gradient matching was proposed to align gradients of classifiers on original and augmented images from isolated source domains, thereby guiding local models towards essential semantic content inherent in both image types.

% \begin{mdframed}[backgroundcolor=lightblue] 
    \begin{equation}
    L_{gm}^{intra} = 1 - sim(g_i, g'_i) 
    \end{equation}    
% \end{mdframed}
\hlblue{where $sim(\cdot)$ denotes the cosine similarity measure, $g_i$ represents the gradient of classifier $C_i$ with respect to original images, and $g'_i$ corresponds to the gradient of classifier $C_i$ for augmented images.} 

\hlblue{To mitigate the domain gap across decentralized source domains, inter-domain gradient matching leverages classifier heads from other domains as intermediaries. This method aims to minimize gradient discrepancies between the classifier head in use and those from disparate domains, thus reducing domain shift and bolstering model generalization across federated domains.}
% \begin{mdframed}[backgroundcolor=lightblue] 
    \begin{equation}
    L_{gm}^{inter} = \sum_{j=1}^n (1 - sim(g'_i, g_j^{t-1})) 
    \end{equation}    
% \end{mdframed}
\hlblue{where $g'_i$ denotes the gradient of classifier $C_i$ with respect to augmented images, while $g_j^{t-1}$ represents the gradient of classifier $C_j^{t-1}$ from the $j$-th source domain concerning original images. In addition, this approach can be adapted for FDA tasks through the refinement of the target model using pseudo-labeled data from the target domain.}

\hlblue{These studies demonstrate innovation and diversity in addressing domain variability in distributed data domains by adopting gradient consistency as a core mechanism. Although each method has its unique implementation and focus, they collectively highlight the importance of achieving domain-invariant learning and generalization capabilities in FDG.}

\subsubsection{Federated Transfer Learning}
Federated transfer learning \cite{yang2019federated} (FTL) is a valuable approach that can be utilized in scenarios where two datasets exhibit differences not only in terms of samples but also in their feature spaces.  It allows models to leverage knowledge learned from one dataset to improve the performance on a different dataset with distinct features while preserving data privacy and security. FTL can also be applied to FDG, where the goal is to improve the generalization performance of models across multiple federated datasets that exhibit domain shifts and variations. The objective of FTL for FDG is as follows:
\begin{equation}\label{}
\Theta^* = \mathop{\arg\min}\limits_{\Theta} \sum_{i=1}^n \delta_i * \mathcal{L}\left(\Theta, D_i\right) + \beta* \digamma\left(\Theta\right) + \sigma *\Omega\left(\Theta\right)
\end{equation}
where $\Theta$ denotes the model parameters to be optimized, $\Theta^*$ represents the optimal model parameters, $n$ signifies the number of clients within the FL system, $\mathcal{L}\left(\Theta, D_i\right)$ is the loss function that measures the discrepancy between the predicted outputs of the model and the ground truth labels on client $i$'s data $D_i$, $\delta_i$ is the weight assigned to the loss function $\mathcal{L}\left(\Theta, D_i\right)$ for client $i$, $\beta$ is a regularization hyperparameter, $\digamma\left(\Theta\right)$ is the regularization term that encourages model parameter smoothness or complexity control, $\sigma$ is a hyperparameter that controls the trade-off between the regularization term and the DG term, $\Omega\left(\Theta\right)$ is the DG term that aims to minimize the distribution discrepancy between different clients' data distributions, promoting generalization across domains.

\hlyellow{Chen \textit{et al.}\mbox{\cite{chen2022federated}} developed the D-WFA framework, which employs discrepancy-based weighted federated averaging to collaboratively train a global diagnosis model. This approach leverages locally labeled source domain data and unlabeled target domain data, prioritizing data privacy. Zhang \textit{et al.}\mbox{\cite{zhang2021federated1}} introduced an FTL method for fault diagnosis that addresses data privacy and the domain shift phenomenon by utilizing unique models for different users. This method features a federated initialization stage and a federated communication stage powered by deep adversarial learning. Zhang \textit{et al.}\mbox{\cite{zhang2022data}} proposed another FTL method for machinery fault diagnostics. This method uses deep learning-based models trained locally at each client to ensure data privacy and introduces a novel DA approach to bridge domain gaps without sharing local data, showing efficacy in real industrial settings. Li \textit{et al.}\mbox{\cite{li2021federated}} offered the FedSWP framework, enhancing the smart work packaging system for construction occupational health and safety management. This framework focuses on protecting construction workers' personal image information, providing accurate and personalized safety alerts and healthcare solutions while addressing privacy concerns.} \hlyellow{Zhou \textit{et al.}\mbox{\cite{zhou2022bearing}} introduced a bearing faulty prediction method that combines fFTL with knowledge distillation. This method, which includes signal-to-image conversion, multi-source FTL, and multi-teacher-based knowledge distillation, significantly improves generalization capability and achieves higher accuracy in bearing fault prediction tasks with a lower parameter size. Fan \textit{et al.}\mbox{\cite{fan2020iotdefender}} presented IoTDefender, a personalized and distributed intrusion detection framework for 5G IoT. Leveraging FTL and 5G edge computing, this framework tackles the issues of heterogeneity, data isolation, and limited data availability. It achieves high detection accuracy and generalization capability with lower false positive rates, all while preserving user privacy. Chen \textit{et al.}\mbox{\cite{chen2020fedhealth}} showcased FedHealth, an FTL framework for wearable healthcare. This framework tackles data isolation and the demand for personalization by aggregating data via FL and developing personalized models. It showcases accurate and privacy-preserving healthcare applications, including wearable activity recognition and Parkinson's disease diagnosis.}

By leveraging transfer learning techniques within an FL, models can learn from multiple domains while preserving data privacy, enabling improved generalization capabilities and robustness in diverse federated environments.

\subsubsection{Channel Decoupling}
Channel decoupling is a technique used in ML, particularly in DG, to disentangle domain-specific information from shared information in feature representations. It can be used in FDG to disentangle domain-specific knowledge from shared knowledge by decoupling channel-wise statistics, enabling better generalization across unseen domains by reducing domain-specific variations.

Yang \textit{et al.} \cite{yangclient} proposed a method for client-agnostic learning with mixed instance-global statistics. In this approach, local models aim to learn client-invariant representations by optimizing client-agnostic objectives using augmented data. \hlyellow{Feng \textit{et al.}\mbox{\cite{feng2022specificity}} put forward the FedMRI algorithm, which partitions the model into a shared encoder and client-specific decoders. This architecture preserves domain-specific characteristics while fostering a generalized representation, effectively tackling domain shift issues. Shen \textit{et al.}\mbox{\cite{shen2022cd2}} introduced the CD2-pFed framework, employing cyclic distillation-guided channel decoupling to tailor the global model for individual clients within FL. This approach effectively navigates the complexities of heterogeneous data distributions, enhancing generalization across clients. Tan \textit{et al.}\mbox{\cite{tan2022federated}} developed FedStar, a federated graph learning framework that tackles the Non-IID problem in graph data. By leveraging structure embeddings and independent structure encoders, FedStar extracts and shares essential structure information, securing structure-based domain-invariant knowledge and mitigating feature misalignment. Weng \textit{et al.}\mbox{\cite{weng2022robust}} proposed FedUCC, an innovative federated unsupervised cluster-contrastive learning method for Person ReID. FedUCC employs a three-stage strategy to discover generic, specialized, and patch knowledge, ensuring cross-domain consistency and maintaining local domain-specific insights.}

\hlyellow{By allocating distinct channels to each domain, the global model is capable of learning domain-specific representations, thereby efficiently overcoming the issues associated with heterogeneous data distributions among the participating domains.} Channel decoupling allows for fine-grained adaptation and enhances the model's ability to generalize across diverse domains in FL. 

\subsubsection{Style Transfer}
Style transfer in FDG refers to the process of transferring the style or visual characteristics of a source domain to a target domain in the FL setting. The goal is to generalize the learned style representations across domains while preserving the content of the target domain.

Nergiz \textit{et al.} \cite{nergiz2022collaborative} introduced a federated adaptation of the neural style transfer algorithm as a data augmentation technique to solve the issues of data collaboration in computational pathology, specifically focusing on the highly class-imbalanced Chaoyang colorectal cancer imaging dataset, while ensuring privacy preservation without any data leakage. Chen \textit{et al.} \cite{chen2023federated} introduced a DG method called \textit{cross-client style transfer} (CCST) for image recognition in FL, which surpasses state-of-the-art methods on DG while being compatible with existing DG techniques. Shenaj \textit{et al.} \cite{shenaj2023learning} introduced FFREEDA, a task for Semantic Segmentation in autonomous driving, proposing the LADD algorithm that utilizes pre-trained model knowledge through self-supervision and a federated clustered aggregation scheme based on clients' style, demonstrating superior performance compared to existing approaches. Fantauzzo \textit{et al.} \cite{ fantauzzo2022feddrive} presented FedDrive, a benchmark for FL in Semantic Segmentation for autonomous driving, addressing difficulties of statistical heterogeneity and DG, evaluating state-of-the-art algorithms with style transfer methods to enhance generalization capabilities.

It's important to note that style transfer for FDG is an evolving research area, and the specific techniques and methodologies may vary depending on the latest advancements.

\subsubsection{Federated Meta-Learning}
\textit{Model-agnostic meta-learning} (MAML) \cite{finn2017model} is a meta-learning approach that aims to learn an initial model that can quickly adapt to new tasks with limited labeled data. \hlyellow{In FDG, MAML is utilized to overcome the obstacle of generalization across diverse federated domains with varying data distributions. MAML targets domain shifts by fostering a meta-level optimization process that facilitates rapid adaptation of models to new tasks using limited data.} Instead of training a model from scratch for each new task, MAML algorithms aim to learn a good initialization or set of initial parameters that can be fine-tuned or adapted to new tasks with minimal data. The specific form of the objective function may vary depending on the MAML algorithm used, but in general, it involves a two-level optimization process: inner-loop optimization and meta-optimization. In the inner-loop optimization, for each task or episode, the model is trained on a support set of labeled examples specific to that task. The goal is to update the model parameters to minimize task-specific loss or error on the support set. This is typically done through gradient descent or other optimization techniques. In the meta-optimization, the meta-parameters of the model are updated based on the performance of the model on the query sets of multiple tasks. The objective is to find meta-parameters that can generalize well across tasks and enable fast adaptation. The meta-parameters are updated using the gradients computed from the meta-objective, which is often defined as the average loss or error across the query sets of multiple tasks. 

\paragraph{Inner-loop optimization:}
The base learner is fine-tuned using support samples $D_{S_\mathfrak{i}}$ from the learnable initialization $\Phi$ for a fixed number of weight updates via gradient descent. The task adaptation objective is to minimize the loss function $L(D_{S_\mathfrak{i}}, \Phi_{\mathfrak{i}, \mathfrak{j}})$ using gradient descent. At the $\mathfrak{j}$-th step of inner-loop optimization, the base learner parameters are updated as:  
\begin{equation}
    \Phi_{\mathfrak{i}, \mathfrak{j}+1} = \Phi_{\mathfrak{i}, \mathfrak{j}} -  \alpha \nabla_{\Phi_{\mathfrak{i},\mathfrak{j}}} L(D_{S_\mathfrak{i}}, \Phi_{\mathfrak{i}, \mathfrak{j}})  
\end{equation}
where $\alpha$ is the learning rate.
\paragraph{Inner-loop optimization iterations:}
After performing $J$ number of inner-loop update steps, the task-specific base learner parameters become $\Phi_{\mathfrak{i}, J}$.
\paragraph{Meta optimization:}
The meta-learned initialization $\Phi$ is evaluated based on the generalization performance of the task-specific base learner with parameters $\Phi_\mathfrak{i}$ (or $\ \Phi_{\mathfrak{i}, J})$ on unseen query examples $D_Q$. The objective of MAML is to minimize the meta-learning algorithm's objective function, denoted as $L(D_{Q_\mathfrak{i}}, \Phi_\mathfrak{i})$, as follows: 
\begin{equation}
    \Phi = \Phi -  \zeta \nabla_{\Phi_T} L(D_{Q_\mathfrak{i}}, \Phi_\mathfrak{i})  
\end{equation}
where $\zeta$ is the meta-learning rate.

\hlyellow{Chen \textit{et al.}\mbox{\cite{chen2018federated}} presented FedMeta, a federated meta-learning framework that tackles both statistical and systematic complications in FL. By reducing communication costs, enhancing convergence speed, increasing accuracy, and ensuring privacy, FedMeta surpasses existing optimization algorithms. Wang \textit{et al.}\mbox{\cite{wang2022graphfl}} inspired by (MAML, proposed GraphFL methods to navigate the Non-IID issue in graph data and new label domains, employing a structure that divides each task into a labeled training set and a separate query set. Jiang \textit{et al.}\mbox{\cite{jiang2019improving}} introduced a personalized FL approach utilizing MAML algorithms. This work elucidates the meta-learning interpretation of the widely adopted FedAvg algorithm and proposes a refined version tailored to enhance FL personalization, ultimately aiming to improve the global model's generalization capability. Lin \textit{et al.}\mbox{\cite{lin2020collaborative}} developed a collaborative learning framework aimed at real-time edge intelligence, leveraging federated meta-learning to train models across multiple source edge nodes and swiftly adapt them to new tasks at target nodes with limited data. Fallah  \textit{et al.}\mbox{\cite{fallah2021generalization}} delved into the generalization properties of MAML algorithms within supervised learning. They presented a novel stability definition that accounts for the number of tasks and samples per task, facilitating an in-depth analysis of MAML's generalization error.}

\hlyellow{Li \textit{et al.}\mbox{\cite{li2022generalizable}} introduced a cutting-edge pancreas segmentation model that leverages a meta-learning strategy and latent-space feature flow generation. This approach effectively reduces interference from background clutter and appearance-style discrepancies through a coarse-to-fine workflow, achieving superior performance across three pancreas datasets and outperforming current state-of-the-art generalization methods. Chen \textit{et al.}\mbox{\cite{chen2023industrial}} introduced FedMeta-FFD, a pioneering method that combines FL and meta-learning to address the few-shot fault diagnosis task in IIoT. This method allows clients to utilize external datasets during the training of a global meta-learner, leading to enhanced convergence speed and accuracy compared to existing techniques. Khodak \textit{et al.}\mbox{\cite{khodak2019adaptive}} proposed a theoretical framework that merges task-similarity, online convex optimization, and sequential prediction algorithms. This framework enhances few-shot learning and FL performance through adaptive task-similarity learning, improved transfer-risk bounds, and average-case regret bounds in dynamic or varied tasks.}

By utilizing MAML, the model parameters can be updated on meta-training data, which consists of raw input data from different federated domains, and then evaluated on held-out meta-test data generated from frequency space with different distributions. In addition to the above MAML methods, other approaches are being explored in the field of federated meta-learning for FDG. One is episodic learning, which can be applied in FDG to improve the transferability of models across different domains. Liu \textit{et al.} \cite{liu2021feddg, liu2023domain} introduced episodic meta-learning for training models in FL across varying data distributions by explicitly simulating domain shifts and employing a meta-learning approach to update model parameters. Another one is the meta-critic network, where the meta-critic network provides feedback on the performance of primary models across different domains, allowing for guided adaptation and improvement of their generalization capabilities. \hlyellow{Chen \textit{et al.}\mbox{\cite{chen2022learning}} proposed a personalized \textit{federated optimization framework with Meta Critic} (FedMC) to tackle heterogeneity in federated networks. This framework effectively identifies robust and generalizable domain-invariant knowledge across clients, \hlyellow{tackling the issues arising from} diverse client data distributions.}

\subsubsection{Data-free Methods}
Data-free \cite{frikha2023towards} methods aim to perform learning tasks without using any labeled data from the target domains by leveraging auxiliary information, prior knowledge, or transfer learning techniques to generalize to unseen domains.

\indent\textbf{Zero-shot and few-shot learning:}
Zero-shot learning and few-shot learning aim to recognize or classify objects from unseen domains for which no labeled samples are available during training. 

Sun \textit{et al.} \cite{sun2022federated} proposed Federated \textit{Zero-Shot Learning} (FZSL) to transfer mid-level semantic knowledge and optimize a generalizable FL model. Sariyildiz \textit{et al.} \cite{sariyildiz2019gradient} proposed a generative model combined with Gradient Matching loss, \hlyellow{transforming zero-shot learning issues into a supervised classification task. This approach significantly boosts generalized zero-shot classification accuracy, surpassing the performance of prior methodologies.} Micaelli \textit{et al.} \cite{micaelli2019zero} demonstrated the possibility of achieving zero-shot knowledge transfer through an adversarial approach, where the adversarial generator is trained to produce rigorous images for a student network to match the predictions of a teacher network. Zhang \textit{et al.} \cite{zhang2022fedzkt} introduced FedZKT, a data-free FL framework that enables devices to autonomously select on-device models based on their local resources, facilitating knowledge transfer among these heterogeneous on-device models through a zero-shot distillation technique. Li \textit{et al.} \cite{li2023federated} developed a federated zero-shot fault diagnosis framework that utilizes a semantic knowledge base, a bidirectional alignment network, and cloud-edge collaboration to effectively diagnose both local and global unseen fault categories. \hlblue{Yuan \textit{et al.}\mbox{\cite{yuan2023collaborative}} proposed the Collaborative Semantic Aggregation and Calibration framework, a pioneering approach that facilitates data-free domain generalization by merging model-based semantic information across multiple domains, ensuring adherence to privacy constraints. Additionally, the framework tackles the issue of semantic dislocation by implementing cross-layer semantic calibration, employing an attention mechanism to enhance the precision of semantic alignment.}

By incorporating zero-shot learning and few-shot learning techniques into FDG, models can improve their ability to generalize across multiple domains with limited labeled data.

\indent\textbf{\textit{Knowledge distillation:}} 
Data-free knowledge distillation (DFKD) \cite{micaelli2019zero, nayak2019zero, chen2019data} is a technique used to transfer knowledge from a large, well-trained model to a smaller, more compact model without the need for labeled training data. 

Frikha \textit{et al.} \cite{frikha2023towards} presented DEKAN, an approach named Domain Entanglement via Knowledge Amalgamation from Domain-Specific Networks, which effectively extracts and combines domain-specific knowledge from existing teacher models to create a domain-shift-robust student model. Niu \textit{et al.} \cite{niu2023mckd} introduced a Mutually Collaborative Knowledge Distillation, enhancing both federated UDA and DG by learning domain-invariant features in a data-free manner through collaborative distillation across disparate local models. Zhu \textit{et al.} \cite{zhu2021data} proposed a data-free knowledge distillation method for heterogeneous FL, which utilizes a lightweight generator to aggregate user information and improve the generalization performance of the global model with fewer communication rounds. Jeong \textit{et al.} \cite{jeong2018communication} developed data-free federated distillation, a distributed model training algorithm for on-device ML that minimizes inter-device communication overhead\hlyellow{. This method tackles the Non-IID data problem by implementing federated augmentation, which synthesizes an IID dataset to enhance performance. Zhang \textit{et al.}\mbox{\cite{zhang2022fine}} introduced FedFTG, a data-free knowledge distillation technique for fine-tuning the global model on the server, effectively mitigating data heterogeneity in FL and serving as a complementary enhancement to existing local optimizers.}

DFKD can be leveraged successfully in FDG, wherein the student model is trained using multiple teachers who have been trained on diverse source domains, facilitating the generalization capability to previously unseen target domains. 

\indent\textbf{\textit{Semantic Aggregation and Calibration:}} In FDG, semantic aggregation and calibration refer to techniques used to consolidate and refine semantic representations across multiple federated nodes in a privacy-preserving manner.

Yuan \textit{et al.} \cite{yuan2023collaborative} introduced CSAC, a novel privacy-preserving method that enables separated DG by unifying multi-source semantic learning and alignment through an iterative process of data-free semantic aggregation and cross-layer semantic calibration. Niu \textit{et al.} \cite{niu2023mckd} proposed a data-free semantic collaborative distillation approach aimed at acquiring domain-invariant representations, applicable to both federated UDA and DG.

By employing semantic aggregation and calibration techniques, FDG methods can effectively leverage shared knowledge across domains while adapting to the specific characteristics of individual domains, leading to improved generalization performance and robustness in FL scenarios.

\subsubsection{Federated Domain Translation}
Domain translation \cite{chu2018survey, saunders2022domain} refers to the task of converting or adapting content from one domain to another domain while preserving its meaning and quality. Federated domain translation is a method that aims to perform translation tasks across multiple domains in an FL framework while preserving data privacy and security.

Iterative naive barycenter \cite{zhao2018federated} for FDG aims to find a common representation or domain-invariant subspace by iteratively averaging the distributions or representations from multiple source domains. Zhou \textit{et al.} \cite{zhou2023efficient} proposed a federated domain translation method that generates pseudo-data for each client which could be useful for multiple downstream learning tasks. Wang \textit{et al.} \cite{xie2022fedmed} \hlyellow{introduced FedMed-GAN, a benchmark for federated domain translation. This benchmark effectively combats mode collapse, preserves generator performance, and showcases adaptability across varying data proportions and distributions.}

By translating data or knowledge across domains in the FL setup, models can benefit from a more diverse and comprehensive training dataset, leading to improved performance and generalization in real-world applications.

\subsubsection{Regularization Strategies}
Regularization strategies in FDG aim to improve the generalization performance of models across multiple domains by mitigating the negative effects of domain shift and improving domain-invariant representations. 

To combat overconfidence and dropping learning curves in pseudo-labeling for FDG, Shenaj \textit{et al.} \cite{shenaj2023learning} proposed using knowledge distillation loss based on soft predictions and stochastic weight averaging applied to clients' teachers to stabilize the learning curve and prevent the model from forgetting pre-training knowledge. Zhuang \textit{et al.} \cite{zhuang2022federated} presented a novel domain constraint loss (DCL) in FL to regularize source domain training where DCL serves as a regularization term during the training of the source domain model, promoting alignment towards the target domain within the aggregated global model. Zhu \textit{et al.} \cite{zhu2022fedilc}  proposed a new regularization methodology called federated invariant learning consistency to promote both inter-silo and intra-silo consistencies in federated networks by leveraging the domain-level gradient covariance and the geometric averaging of Hessians. In addition, the authors also introduced a novel weighted geometric mean mechanism to calculate geometric mean with inconsistent signs on gradients. Atwany \textit{et al.} \cite{atwany2022drgen} employed flatness in the training of separated DG, incorporating iteration-wise weight averaging and domain-level gradient variance regularization as effective techniques. Feng \textit{et al.} \cite{feng2022specificity} developed a weighted contrastive regularization method for the globally shared part of the model, which helps alleviate domain shifts among clients during training and improves convergence. Liu \textit{et al.} \cite{liu2021feddg} employed the InfoNCE objective to impose regularization, using positive and negative feature pairs, to enhance the domain-invariance and discriminability of the features in their FL model, ultimately improving its performance. \hlyellow{Huang  \textit{et al.}\mbox{\cite{huang2023rethinking}} implemented consistency regularization to align local instances with their respective unbiased prototypes, markedly enhancing both the effectiveness and efficiency of mitigating domain shifts in FL.}

Overall, regularization strategies play a crucial role in FDG by mitigating overfitting and improving the generalization performance of models across different unseen domains.

\subsection{Aggregation Optimization Algorithms}
\paragraph{Weight Optimization}
The FedAvg algorithm in FL aggregates global model parameters based on the number of local model parameters contributed by each client, disregarding the quality and importance of the local model parameters, which can adversely impact the performance of the global model. In FDG, optimizing aggregation weights is crucial to overcoming the dilemmas posed by heterogeneous data distributions and non-uniform network architectures across participating clients. 

Chen \textit{et al.} \cite{chen2022federated} proposed D-WFA, a dynamic weighting strategy based on MMD, adjusting the weights of clients and updating multiple local models with generalization capability. Zhang \textit{et al.} \cite{zhang2023federated2} presented the Generalization Adjustment (GA) method in FL dynamically calibrates aggregation weights to optimize the objective, achieving a tighter generalization bound by explicitly re-weighting aggregation instead of relying on implicit multi-domain data sharing as in conventional DG settings. Yuan \textit{et al.} \cite{yuan2023collaborative} introduced a semantic similarity-based attention mechanism that dynamically assigns weights to cross-layer pairs, prioritizing pairs with higher semantic similarity for precise alignment while reducing the emphasis on less similar pairs to prevent semantic dislocation. Yang \textit{et al.} \cite{yangclient} proposed to dynamically generate instance-wise interpolation parameters for mixing instance and global statistics with a learning-based network to build the global model deployed to unseen clients. \hlyellow{Alekseenko \textit{et al.}\mbox{\cite{alekseenko2024distance}} proposed a methodology that employs distance measurements to address Non-IID data in FL, enhancing model generalization by downgrading the influence (weight) of the most divergent client.}

Optimizing aggregation weights in FDG enables effective integration of client contributions, accounting for their importance and relevance, resulting in a more adaptive and robust global model capable of performing well across diverse data distributions and network architectures.

\paragraph{Gradient Optimization} 
Gradient optimization in FL refers to the process of optimizing the model parameters by leveraging the gradients computed on local participant data. Gradient optimization plays a crucial role in FDG by enabling efficient parameter updates across decentralized domains, ensuring convergence to an optimal global model while considering the variations in data distributions and network architectures, thereby enhancing the model's generalization capability and performance across different domains and scenarios.

Federated voting aimed to fine-tune the global model in FL without using the ground labels of the target domain samples, thus improving the effectiveness of feature distribution matching \cite{sun2023feature}. In the global mutual optimization \cite{li2021fedh2l}, each domain \hlyellow{underwent} model distillation to learn from its peers, effectively mitigating any potential negative impacts of domain-specific gradient updates on the overall global model. To enhance the robustness of the global feature encoder across clients, a domain critic \cite{chen2022learning} was introduced to regularize the feature representations learned by both the local and global feature encoders, utilizing the Wasserstein critic as the chosen distance metric due to its gradient superiority. Tian \textit{et al.} \cite{tian2021privacy} introduced an innovative gradient alignment loss to enhance the gradient aggregation process on centralized servers, aiming to bolster the model's generalization to novel yet related domains. Zeng \textit{et al.} \cite{zeng2022gradient} developed the Gradient Matching Federated fine-tuning method, which updates pre-trained local federated models using the One-Common-Source Adversarial Domain Adaptation strategy, minimizing the gradient matching loss between sites to drive the optimization direction towards specific local minima. Wu \textit{et al.} \cite{wu2021collaborative} introduced a prediction agreement mechanism to overcome local disparities towards central model aggregation for achieving better decentralized DG. \hlyellow{Zhang \textit{et al.}\mbox{\cite{zhang2023grace}} 
introduced a novel gradient correction technique, termed GRACE, leveraging feature alignment regularization within a meta-learning framework on the client side to mitigate the overfitting of personalized gradients.}

In summary, optimizing aggregated gradients in FDG is essential to \hlyellow{mitigate domain shift, accommodate data heterogeneity, and enhance model performance across various domains}
\begin{figure}
  \centering
  \begin{tikzpicture}

    \begin{axis}[
    xbar stacked,   % Stacked horizontal bars
    xmin=0,         % Start x axis at 0
    ytick=data,     % Use as many tick labels as y coordinates
    legend columns=8,
    legend style={
        at={(0.5,1.03)}, % Adjusted to place the legend above the chart
        anchor=south, % Anchors the legend at its bottom
        legend columns=-1, % Ensures the legend entries are in a single row
        font=\footnotesize % Makes the font size smaller
    },
    yticklabels from table={\testdata}{Label}  % Get the labels from the Label column of the \datatable
    ]
    % \addplot [fill=green!80] table [x=2017, meta=Label,y expr=\coordindex] {\testdata};   % "First" column against the data index
    % \addplot [fill=blue!60] table [x=2018, meta=Label,y expr=\coordindex] {\testdata};
    % \addplot [fill=red!60] table [x=2019, meta=Label,y expr=\coordindex] {\testdata};
    % \addplot [fill=orange!60] table [x=2020, meta=Label,y expr=\coordindex] {\testdata};
    % \addplot [fill=yellow!60] table [x=2021, meta=Label,y expr=\coordindex] {\testdata};
    % \addplot [fill=lime!60] table [x=2022, meta=Label,y expr=\coordindex] {\testdata};
    % \addplot [fill=green] table [x=2023, meta=Label,y expr=\coordindex] {\testdata};
    \addplot [fill=cyan!60!black!30] table [x=2017, meta=Label,y expr=\coordindex]{\testdata};
    \addplot [fill=yellow!60!black!30] table [x=2018, meta=Label,y expr=\coordindex]{\testdata};
    \addplot [fill=gray!50] table [x=2019, meta=Label,y expr=\coordindex]{\testdata};
    \addplot [fill=green!60!black!30] table [x=2020, meta=Label,y expr=\coordindex]{\testdata};
    \addplot [fill=violet!60!black!30] table [x=2021, meta=Label,y expr=\coordindex]{\testdata};
    \addplot [fill=orange!60!black!30] table [x=2022, meta=Label,y expr=\coordindex]{\testdata};
    \addplot [fill=teal!60!black!20] table [x=2023, meta=Label,y expr=\coordindex]{\testdata};
    \addplot [fill=red!60!black!30,
    point meta=x,
    nodes near coords,
    nodes near coords align={anchor=west},
    every node near coord/.append style={
        black,
        fill=white,
        fill opacity=0.75,
        text opacity=1,
        outer sep=\pgflinewidth % so the label fill doesn't overlap the plot
    }
    ] table [x=2024, meta=Label,y expr=\coordindex] {\testdata};
    \legend{2017,2018,2019, 2020, 2021, 2022, 2023, 2024}

    \end{axis}
    \end{tikzpicture}
  \caption{Publication trends across methods (As of February 2024.)}
  \label{fig:stacked_bar}
\end{figure}
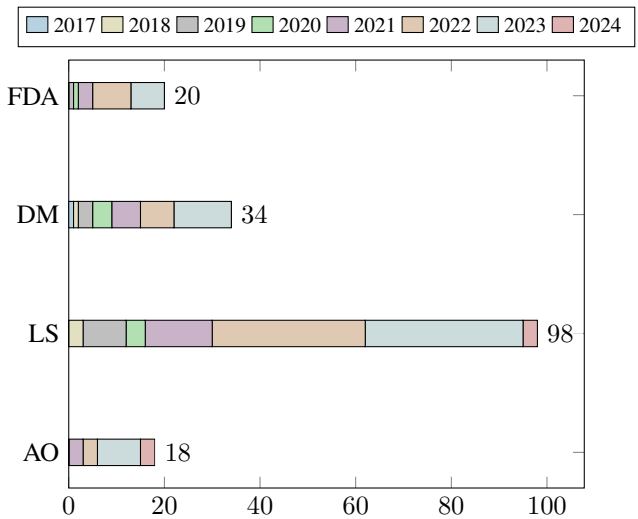

\paragraph{Aggregation Policy:}
The aggregation policy in FL refers to the strategy or algorithm used to combine the model updates from multiple participating clients and generate an updated global model. In FDG, designing an effective aggregation policy is essential for achieving accurate and robust models.

Shenaj \textit{et al.} \cite{shenaj2023learning} introduced a clustered and layer-aware aggregation policy where global parameters were aggregated over selected clients and cluster-specific parameters were averaged within clusters, where clients were clustered based on their transferred styles using H-Means clustering. \hlyellow{Cong \textit{et al.}\mbox{\cite{cong2023federated}} proposed the FDG-GRMA framework, which leverages a robust model aggregation strategy to construct a global model that both preserves data privacy and efficiently diagnoses faults through strategic parameter and feature transmissions between the central server and source domain clients. Zhang \textit{et al.}\mbox{\cite{zhang2023grace}} employed a consistency-enhanced reweighting aggregation on the server side to recalibrate the aggregated gradients, thereby improving the global model's generalization capabilities across heterogeneous datasets. Chung \textit{et al.}\mbox{\cite{chung2023federated}} advanced FDG by introducing a reparameterization technique that redefines local entropy, enabling the aggregation of perturbed local gradients to approximate global gradients, thus facilitating robust and generalized model learning across clients without data sharing. Song \textit{et al.}\mbox{\cite{song2024federated}} proposed aggregation strategy on the central server involves evaluating each local model's performance across other source clients, assigning higher weights to models with poorer generalization to ensure that the final aggregated model exhibits enhanced performance across all domains. Liu \textit{et al.}\mbox{\cite{liu2024domain}} introduced a dynamic aggregation strategy that adaptively weights clients based on domain-invariance principles, which helps to mitigate domain variations and enhance the domain generalization capability of the server model.}

Optimizing the aggregation policy effectively helps to combine and utilize knowledge from diverse domains, ensuring the generation of a robust and adaptable global model capable of performing well across different domains and scenarios.
\subsection{Dissecting Research Dynamics}
\hlyellow{Fig.~\ref{fig:stacked_bar} depicts the publication trends across various methodologies, identified as Aggregation Optimization (AO), Learning Strategies (LS), Data Manipulation (DM), and Federated Domain Alignment (FDA), from 2017 to 2024. Fig.~\ref{fig:stacked_bar} employs a stacked bar chart to elucidate the cumulative publication output per methodology alongside the yearly distribution. A comparative analysis of the bars reveals a marked increase in publications across all subareas, indicative of a growing interest in FDG research. The color-coded stratification within the bars enables an expedient evaluation of the relative volume of research amongst the subareas, highlighting shifts in the focus of the research community. It is particularly noteworthy that the learning strategies methodology has commanded a significantly higher interest in research, as demonstrated by its dominant proportion of the total publications.}

\section{Datasets and Applications} \label{dataset}
In this section, we first provide a summary of the existing commonly used datasets in FDG. Following this, we discuss the popular tasks/applications for this research area.
\subsection{Datasets}
Table ~\ref{tab:dataset} provides an overview of several commonly used datasets in the field of FDG. These datasets have been extensively utilized in various FDG applications, including handwritten digit recognition, object recognition, transportation imaging, medical imaging, person re-identification, industrial edge intelligence, and human activity recognition. They serve as valuable resources for evaluating and benchmarking the performance of FDG algorithms and methodologies. Here, provide a detailed description of several widely used datasets commonly employed in research studies.

\begin{table*}[!t]
\caption{Datasets for FDG.}\label{tab:dataset}
\centering
% \fbox{
\begin{tabular}{lcccp{5.5cm}>{\raggedright\arraybackslash}p{3cm}}
    \toprule
    \textbf{Area \& Dataset} & \textbf{\#Domain} & \textbf{\#Class} & \textbf{\#Sample} & \centering\textbf{Description} & \textbf{Ref.} \\
    \midrule
    \textbf{Handwritten digit recognition} & &  &  &  &  \\
    \textit{Digit} &  4 & 10 & 203,695 &  MNIST, MNIST-M, SVHN, USPS & \cite{frikha2023towards}  \\
    \textit{Digits-DG}  &4 & 10 & 219,289 & MNIST, MNIST-M, SVHN, SYN  & \cite{wu2021collaborative}\\
    \textit{Rotated MNIST}  & 6 & 10 & 70,000 & M0, M15, M30, M45, M60, M75 &  \cite{yuan2023collaborative, li2021fedh2l, nguyen2022fedsr, de2022mitigating}\\
    \textit{Digit-Five} & 5 & 10 & 215,695 &  MNIST, MNIST-M, Synthetic Digits, SVHN, USPS & \cite{sun2023feature, niu2023mckd}  \\
    \midrule
    \textbf{Object recognition} & &  &  &  &  \\
    \textit{Office-Home} &4 & 65 & 15,588 & Art, Clipart, Product, Real-world & \cite{wu2021collaborative, zhang2023federated, li2021fedh2l, yangclient, zhang2021federated2, nguyen2022fedsr, niu2023mckd, zhou2022source, chen2023federated, de2022mitigating} \\
    \textit{PACS} &4 & 7 & 9,991 & Art, Cartoon, Photos, Sketches & \cite{wu2021collaborative, zhang2023federated,yuan2023collaborative, li2021fedh2l, yangclient, zhang2021federated2, nguyen2022fedsr, frikha2023towards, niu2023mckd, chen2023federated, de2022mitigating} \\
    \textit{VLCS} & 4 & 5 & 39,305 & Ascal, LabelMe, Caltech, Sun & \cite{zhang2023federated, yuan2023collaborative, yangclient, zhang2021federated2} \\
    \textit{DomainNet} & 6 & 345 & 586,575 &  clipart, infograph, painting, quickdraw, real, sketch & \cite{nguyen2022fedsr}  \\
    \textit{Office-31} & 3 & 31 & 4,652 & Amazon, Webcam, DSLR & \cite{zhou2022source} \\
    \textit{Office-Caltech-Home}  & 7 & 65 & 17,964 & Amazon, Webcam, Caltech, Art, Clipart, Product, Real-World & \cite{yuan2023collaborative} \\
    \textit{Office-Caltech10}  & 4 & 10 & 2,533 & Caltech, Amazon, Webcam, DSLR & \cite{sun2023feature, niu2023mckd}  \\
    \textit{ImageCLEF-DA} & 3 & 12 & 1,800 & ImageNet ILSVRC 2012, Caltech-256, Pascal VOC 2012 & \cite{zhou2022source} \\
    \textit{Amazon Review}  & 4 & 2 & 7,671 &  Books, DVDs, Electronics, Kitchen \& housewares & \cite{sun2023feature}  \\
    \midrule
    \textbf{Transportation imaging} & &  &  &  &  \\
    \textit{Cityscapes} & 50 & 19 & 3,475 & Real photos taken in the streets of 50 different cities & \cite{fantauzzo2022feddrive, shenaj2023learning} \\
    \textit{IDDA}  & 105 & 16 &  &  A synthetic dataset in the field of self-driving cars & \cite{fantauzzo2022feddrive} \\
    \textit{Mapillary Vistas}  &  & 66 & 25,000 &  The dataset covers diverse scenes from different domains and cities around the world & \cite{shenaj2023learning} \\
    \textit{IDDA}  & 105 & 16 &  &  A synthetic dataset in the field of self-driving cars & \cite{fantauzzo2022feddrive} \\
    \midrule
    \textbf{Medical imaging} & & & & &  \\
    \textit{Skin Lesion} & 7 & 7 & & HAM10000, Dermofit, Derm7pt, MSK, PH2, SONIC, UDA & \cite{tian2021privacy}  \\
     \textit{ISIC 2018} & 7 & 7 & 10,015 & Melanocytic Nevus, Melanoma, Benign Keratosis, Basal Cell Carcinoma, Actinic Keratosis, Vascular Lesion, Dermatofibroma & \cite{li2023rare}  \\
     \textit{Fed-ISIC2019} & 6 & 8 & 25,331 & Melanoma, Melanocytic nevus, Basal cell carcinoma, Actinic keratosis, Benign keratosis, Dermatofibroma, Vascular lesion, Squamous cell carcinoma, None of the others & \cite{zhang2023grace}  \\
    \textit{Dermofit Image Library} & 10 & 10 & 1,300 & A comprehensive collection of dermatology images for study, analysis, and diagnosis of skin conditions & \cite{li2023rare}  \\
    \textit{YawDD} & 2 & 3 &  & A dataset of videos, recorded by an in-car camera & \cite{li2021federated}  \\
    \midrule
    \textbf{Person Re-identification} & &  &  &  &  \\
    \textit{Person Re-ID dataset} & 8 & 7,270 & 93,796 & DukeMTMCReID, Market1501, CUHK03-NP, PRID,
CUHK01, VIPeR, 3DPeS, iLIDt. & \cite{weng2022robust}  \\

    \midrule
    \textbf{Industrial Edge Intelligence} & &  &  &  &  \\
    \textit{CWRU} &  4 & 10 & 900 & A widely used dataset for fault diagnosis and condition monitoring of rotating machinery & \cite{zhang2021federated1, chen2023industrial, zhang2022data, wang2022federated}  \\
    \textit{PU} & / & 10 & 52,497 & A 6203 bearing dataset, obtained from Paderborn University, includes artificially induced and real damages. & \cite{chen2023industrial} \\
    \midrule
   \textbf{Human Activity Recognition} & & & & &  \\
    \textit{UCI Smartphone} & 30 & 6 & 10,299 & Walking, Walking\_upstairs, walking\_downstairs, Sitting, Standing, Laying & \cite{chen2020fedhealth}  \\
    \bottomrule
\end{tabular}
% }
\end{table*}

\textit{Rotated MNIST}, a variant of the original MNIST, incorporates rotation transformations to the original handwritten digit images (M0, M15, M30, M45, M60, M75), serving as a benchmark dataset for evaluating the robustness and generalization ability of ML algorithms and models in the context of image recognition and computer vision tasks. \textit{PACS} is a demanding dataset for DG, comprising seven distinct object categories from four diverse domains (Art Painting, Cartoon, Photo, and Sketch), intricate models to effectively generalize across significant domain discrepancies and variations in artistic styles, renders techniques, and visual representations. \textit{Office-Home}, encompassing four distinct domains (Artistic, Clipart, Product, and Real-World), has gained widespread usage in the fields of domain adaptation and transfer learning, serving as a valuable resource for evaluating models' ability to generalize across diverse visual domains that encompass stylized and real-world images. 

\textit{Digits-DG} is a notable dataset specifically designed for DG in digit recognition tasks incorporating four distinct digit datasets (MNIST, MNIST-M, SVHN, and SYN), introducing variations in font styles and backgrounds to simulate real-world scenarios and complex models to generalize across diverse writing styles, backgrounds, and digit representations. \textit{VLCS} is a comprehensive benchmark dataset designed to bridge the gap between vision and language understanding tasks by providing a rich collection of visual scenes accompanied by linguistic descriptions (Ascal, LabelMe, Caltech, Sun), enabling to explore and develop models that integrate visual perception and linguistic reasoning. \textit{CWRU} is a widely recognized benchmark in ML and fault diagnosis, comprising vibration signals from diverse rotating machinery under different operating conditions and fault scenarios, providing researchers with a valuable resource for developing and evaluating fault diagnosis and condition monitoring algorithms.

\subsection{Applications}
FDG has broad applications in scenarios where data is distributed across multiple domains (organizations), and data sharing is limited due to privacy concerns or regulatory reasons. Here, we will discuss some of the popular tasks/applications for FDG.

\subsubsection{Healthcare} In the field of healthcare, each hospital collects medical data from its patients, and each hospital has its patient population. The data collected may differ in terms of demographics, medical conditions, and treatment protocols. However, the goal is to develop a model that can accurately predict disease outcomes for patients from all hospitals, regardless of the differences in data distribution. The FDG approach \hlyellow{facilitates overcoming this issue by enabling models to be trained locally at each hospital, thereby ensuring data privacy} \cite{xu2023federated, ding2023generalizable, abdel2022collaborative, luo2023influence}. The locally trained models can then be combined to obtain a more generalized model that predicts well on any unseen domain. 

\subsubsection{Finance}
The finance industry deals with sensitive and confidential data and the data is often distributed among multiple institutions, such as banks, insurance companies, and payment processors. FL \hlyellow{holds significant value in financial applications as it tackles data privacy concerns, facilitates cross-silo collaboration, enhances model accuracy,} and provides better financial services. While FL in finance offers significant benefits in terms of data privacy and cross-institutional collaboration, the federated model may indeed face challenges when applied to unseen finance domains. FDG plays a crucial role in finance to tackle the above issue by providing financial distress predictions \cite{imteaj2022leveraging}, risk analysis \cite{li2023research, cheng2021secureboost}, and fault detection \cite{zheng2021federated}. The adoption of FDG in the finance industry contributes to the advancement of more precise, flexible, and comprehensive models tailored for a range of finance applications, thereby yielding mutual benefits for financial institutions and their clients.

\subsubsection{Education}
Educational institutions are increasingly engaging in international collaborations, allowing students to connect and learn from peers in different countries, promoting cultural understanding and global perspectives. However, educational institutions must adhere to privacy and data protection regulations when collecting, storing, and analyzing student data. FDG could be employed to achieve international collaborations in educational institutions while preserving data privacy, such as grades classification \cite{xu2022federated}, digital educational environment \cite{soboleva2020characteristics}. Introducing FDG in educational institutions helps to understand data privacy protection, handle data distribution differences, strengthen transfer learning, foster innovation and research, integrate theory with practice, and develop the ability to tackle the complexities of training and generalizing models in distributed data environments.

\subsubsection{Transportation}
The transportation sector plays a vital role in facilitating the movement of people and goods within and between cities, regions, and countries. Transportation data is typically collected and managed by transportation authorities, government agencies, transportation service providers, and technology companies, which is crucial to ensure the security and privacy of information. FDG in the transportation domain can facilitate data collaboration between cities and regions, support the design and optimization of intelligent transportation systems \cite{wang2022ai, abdel2021federated, zhang2021toward}, autonomous drive \cite{shenaj2023learning}, optimal transport \cite{farnia2022optimal}, and enhance capabilities for traffic safety \cite{wang2022ai}. By applying FDG methods and techniques, the transportation domain can leverage distributed data for modeling and decision-making, leading to more intelligent, efficient, and sustainable transportation systems.

\subsubsection{Natural Language Processing}
In NLP, data plays a crucial role as it provides the foundation for building and training various NLP models and systems. It's worth noting that NLP models often require large and diverse datasets to achieve high performance and generalization. However, there exists data quality, data bias, insufficient data volume, arduous data annotation, and data privacy and ethical concerns in NLP, addressing them effectively is essential for developing high-quality and reliable NLP models. \hlyellow{The adoption of FDG in NLP tackles issues related to domain shift, data distribution, privacy, and adaptability. It achieves this by improving the generalization ability, promoting collaboration while safeguarding data privacy, and supporting the development of sturdy models for previously unseen domains\mbox{\cite{bai2023benchmarking, lin2021fednlp, linworkshop}}}.

\subsubsection{Robotics} 
In general, robotic systems often involve sensitive data (i.e., such as images or sensor readings) provided via robots deployed in various locations or from their surrounding environments, which may pose privacy and security risks for centralized data transfer. The convergence of FDG and robotics has substantial potential to foster knowledge transfer, and facilitate cross-robot collaboration, resulting in improved performance,  adaptability, and privacy protection of robotic systems. FDG has paid attention to the robotic areas, such as learning autonomously \cite{xianjia2021federated} to smart robotic wheelchairs \cite{casado2022federated}, learning basic table tennis movements from human inputs \cite{mulling2013learning}, cloud robotic systems \cite{liu2020federated, liu2019lifelong}, object recognition and grasp planning in surface decluttering \cite{tanwani2019fog}. The significance of FDG for robotics lies in its ability to enhance the generalization capability and robustness of robotic systems across different environments, tasks, and robots, thereby improving their adaptability, performance, and privacy protection in real-world applications.

\subsubsection{Industrial Edge Intelligence}  
Intelligent instrumentation and measurement technology has advanced rapidly which has resulted in a significant increase in industrial data. This has led to the inevitable arrival of the "era of industrial big data" \cite{he2021modified,cui2021spectrum, murturi2023communityai}. 
However, the utilization of industrial big data poses challenges in terms of data management, analysis, and privacy. FDG is crucial in the era of industrial big data as it enables scalable processing and analysis, ensures data privacy and security, facilitates domain-specific knowledge sharing, and supports real-time analytics. For instance, \cite{chen2022federated, zhang2021federated1, hazra2022cooperative} utilize FDG techniques to solve the faulty diagnosis issue in industrial edge intelligence. Plus, \cite{chen2023industrial} allows clients to leverage indirect datasets from collaborators, training a global meta-learner capable of addressing few-shot problems and adapting to new clients or fault categories with minimal labeled examples and iterations. By leveraging FDG, industrial organizations can effectively utilize and derive value from the wealth of data generated in the industrial environment.

\subsubsection{Distributed Computing Continuum Systems}
The advent of edge computing, complementing cloud computing, has catalyzed the creation of Distributed Computing Continuum Systems (DCCS) \cite{casamayor2023fundamental, casamayor2023distributed, donta2022promising, donta2023governance, dustdar2022distributed}, exploiting the cloud's expansive resources along with the edge's diversity and immediacy. However, current research on edge computing and the distributed computing continuum primarily targets niche problems, yielding solutions of restricted scope. Traditional computer system architectures fall short of encapsulating the DCCS's complexity, characterized by its heterogeneous devices and networks. Since the functional requirements of these systems may evolve, services may be dynamically added or changed, or unexpected events may occur, resulting in alterations to the underlying infrastructure configuration. This necessitates a new and innovative representation of DCCS that is not limited by outdated architectural models. Furthermore, several research works emphasize the need to distribute functionalities and processes among computing continuum entities in a decentralized manner \cite{dustdar2021towards,murturi2022utilizing, dst2020, murturi2021decentralized,murturi2019edge, tsigkanos2019dependable}. Therefore, this has opened the possibility of developing FDG in DCCS for maintaining data privacy and security \cite{murturi2023learning}, enabling cross-domain model generalization, and enhancing system performance and robustness in distributed computing environments.

\section{Evaluations and Benchmarks} \label{evaluations}
\hlred{In this section, we provide commonly used evaluation criteria and benchmarks for FDG.}
\subsection{Evaluation}
\hlred{In general, FDG can be evaluated under various strategies and metrics, they are further discussed below:}
\subsubsection{Evaluation Strategies}
When evaluating FDG methods, diverse strategies can be employed to comprehensively assess the model's capacity to generalize effectively to previously unseen domains. Here is a summary of evaluation strategies commonly used for FDG:
\begin{itemize}
    \item \textbf{Leave-one-domain-out cross-validation (LODO)\cite{liu2021feddg, zhang2023federated}:} Leave-one-domain-out cross-validation is an evaluation strategy where one domain is held out during training, allowing the model's performance to be assessed on unseen domains and providing insights into its generalization capabilities across federated settings.
    \item \textbf{Test-domain validation set (TEVS)\cite{gulrajani2021search}:} Test-domain validation set refers to a subset of data from a specific domain that is used for model evaluation during the validation phase.
    \item \textbf{Training-domain validation set (TRVS)\cite{gulrajani2021search}:} Training-domain validation set refers to a subset of data from the same domain as the training data that is used for model validation during the training phase. 
\end{itemize}
\subsubsection{Evaluation Metrics}
Evaluation metrics in FDG aim to assess the performance and generalization ability of models across different domains within the FL setting. 
\begin{itemize}
    \item \textbf{Accuracy:} Accuracy is a prevalent evaluation metric employed to quantitatively assess the performance of machine learning models. \hlblue{Except for average accuracy, previous work\mbox{\cite{qin2022uncertainty}} followed\mbox{\cite{saito2018open}} using the average accuracy among all classes and the average accuracy of the unknown classes in federated open-set DA:}
    
    \textit{The average accuracy among all classes (OS):}
    \begin{equation}
        Acc(OS) = \frac{1}{\mathbb{K}+1} \sum_{\mathrm{k}=1}^{\mathbb{K}+1} \cfrac{|x^t \in \mathcal{D}^\mathcal{T}_\mathrm{k} \land \hat{y}_\mathrm{k}=y_\mathrm{k}|}{\mathcal{D}^\mathcal{T}_\mathrm{k}}
    \end{equation}
    
    % \begin{mdframed}[backgroundcolor=lightred] 
    \textit{The accuracy of the unknown classes ($OS^*$):}
    \begin{equation}
        Acc(OS^*) = \frac{1}{\mathbb{K}} \sum_{\mathrm{k}=1}^{\mathbb{K}} \cfrac{|x^t \in \mathcal{D}^\mathcal{T}_\mathrm{k} \land \hat{y}_\mathrm{k}=y_\mathrm{k}|}{|\mathcal{D}^\mathcal{T}_\mathrm{k}|}
    \end{equation}
    % \end{mdframed}

    \textit{The average accuracy among known classes (UNK):}
    \begin{equation}
        Acc(UNK) =  \cfrac{|x^t \in \mathcal{D}^\mathcal{T}_{\mathbb{K}+1} \land \hat{y}_{\mathbb{K}+1}=y_{\mathbb{K}+1}|}{\mathcal{D}^\mathcal{T}_{\mathbb{K}+1}}
    \end{equation}
    where $\hat{y}$ denotes the predicted value, $\mathcal{D}^\mathcal{T}_k$ represents the set of target samples with the label $y_k$, $\mathcal{K} := \{1, \cdot\cdot\cdot, \mathrm{k}, \cdot\cdot\cdot, \mathbb{K}\}$ and the unknown class is defined as class $\mathbb{K}$. However, accuracy may not be suitable for imbalanced datasets (Non-IID in FL) where the class distribution is skewed, as it can be influenced by the majority class. Therefore, it is important to consider other evaluation metrics.

    \item \hlyellow{\textbf{Mean Absolute Error (MAE)\mbox{\cite{balint2023using}}} In FDG, the MAE serves as a critical metric to evaluate the accuracy of models across distributed domains.}

    \begin{equation}
        MAE = \frac{1}{N} \sum_{\mathrm{k}=1}^N |y_\mathrm{k} - \hat{y}_\mathrm{k}|
    \end{equation}
    where $\hat{y}$ denotes the predicted value, $y$ denotes the actual observation, $\mathbb{N}$ is the numbers of samples. In FDG, where models are trained on data from multiple domains without sharing raw data, MAE provides a valuable assessment of how well a model generalizes across different sites or domains.
    
    \item \textbf{Standard deviation (SD)\cite{yangclient}:} In FDG, the standard deviation can be used to assess the distributional differences or variations across different domains.
    \item \textbf{Dice coefficient(Dice)\cite{liu2021feddg}:} The Dice coefficient, also known as the Sørensen-Dice coefficient, is a similarity metric commonly used in image segmentation tasks and medical image analysis. It measures the agreement or overlap between two domains. The Dice coefficient is calculated as the ratio of twice the intersection of the two domains to the sum of the sizes of the individual domains:
    \begin{equation}
    Dice(\mathcal{P}, \mathcal{Q}) = \frac{2*Comm(\mathcal{P}, \mathcal{Q}))}{|\mathcal{P}| + |\mathcal{Q}|}
\end{equation}
where $Comm(\mathcal{P}, \mathcal{Q})$ represents the number of common samples between the two domains, while $|\cdot|$ represents the total number of samples in each domain. \hlred{This metric is widely applied in tasks like image segmentation to assess the accuracy and similarity of segmented domains against a ground truth or reference. It offers a valuable metric for evaluating model performance in precisely delineating domains of interest, proving especially beneficial in scenarios characterized by class imbalance or an uneven distribution of domains.}

\item \textbf{Hausdorff distance (HD)\cite{liu2021feddg}:} Hausdorff distance is a metric used to measure the dissimilarity or distance between two domains of points, contours, or shapes. It quantifies the maximum distance between any point in one domain and its nearest point in the other domain. The Hausdorff distance is defined as follows:
\begin{equation}
    \begin{aligned}
        HD(\mathcal{P}, \mathcal{Q}) & = \max(h(\mathcal{P}, \mathcal{Q}), h(\mathcal{Q}, \mathcal{P}))       \\
        h(\mathcal{P}, \mathcal{Q}) & = \max \limits_{p \in \mathcal{P}}\{\min \limits_{q \in \mathcal{Q}}\|p - q\|\}    \\
        h(\mathcal{Q}, \mathcal{P}) & = \max \limits_{q \in \mathcal{Q}}\{\min \limits_{p \in \mathcal{P}}\|q - p\|\}    \\
    \end{aligned}
\end{equation}
where $\mathcal{P}$ and $\mathcal{Q}$ are the two domains being compared, and $h(\mathcal{P}, \mathcal{Q})$ represents the directed Hausdorff distance from domain $\mathcal{P}$ to domain $\mathcal{Q}$. It is calculated by finding the maximum distance from each point in $\mathcal{P}$ to its nearest point in $\mathcal{Q}$. Similarly, $h(\mathcal{Q}, \mathcal{P})$ represents the directed Hausdorff distance from domain $\mathcal{Q}$ to domain $\mathcal{P}$.
\item \textbf{Wasserstein Distance (WD)\cite{zhou2023efficient}:} Wasserstein Distance is a mathematical metric that quantifies the dissimilarity between two probability distributions. In FDG, the WD measures the dissimilarity or discrepancy between the probability distributions of different domains in the FL setting. The WD is calculated as follows:
\begin{equation}
    WD = \frac{1}{M^2}\sum_m\sum_{m'} \widehat{WD}(x_m, f_{m'\to m}{(x_m)})
\end{equation}
where $M$ denotes the number of domains, each $\widehat{WD}$ \cite{cuturi2013sinkhorn} is computed with the Sinkhorn algorithm.
\item \textbf{Fréchet Inception Distance (FID) score\cite{zhou2023efficient}:} The FID score is a metric used to evaluate the quality of generative models, particularly in GANs. In FDG, it can be used as an evaluation metric to provide a quantitative measure of the dissimilarity between the generated samples from different domains. The FID is defined as follows:
\begin{equation}
    FID = \frac{1}{M^2}\sum_m\sum_{m'} \widehat{FID}(x_m, f_{m'\to m}{(x_m)})
\end{equation}

The FID \cite{heusel2017gans} could be utilized to assess GAN performance and effectively quantify disturbance levels.
\item \textbf{$\mathcal{A}$-distance\cite{qin2022uncertainty, zhang2023grace}:} $\mathcal{A}$-distance (also known as the adversarial distance) is a metric used in DG and DA tasks, which measures the discrepancy between probability distributions of different domains or datasets. 
\begin{equation}
    \hat{d}_{\mathcal{A}} = 2(1-2\varepsilon)
\end{equation}
where $\varepsilon$ represents the generalization error of a two-sample classifier trained on the binary task of distinguishing input samples between the source and target domains.
\item \textbf{The standard deviation of the mean Intersection over Union (mIoU)\cite{fantauzzo2022feddrive}:} By computing the mean and standard deviation of the mean IoU, the average performance and the variability in performance across multiple evaluations can be accessed. The mean provides an overall measure of accuracy, while the standard deviation indicates the consistency or instability of the results.

\begin{enumerate}
\item Mean (mIoU): $mIoU = \frac{1}{N}*\sum_{\mathfrak{i}=1}^N(IoU_\mathfrak{i})$. 

where $N$ is the total number of evaluations and $IoU_\mathfrak{i}$ represents the Intersection over Union value for the $\mathfrak{i}$-$th$ evaluation.

\item Standard Deviation: $SD = \sqrt{\frac{\sum_{\mathfrak{i}=1}^N((IoU_\mathfrak{i}-mIoU)^2)}{N}}$.

\item The standard deviation of mIoU: $mIoU \pm SD$.
\end{enumerate}

\item \textbf{Group Effect (GE) \cite{sun2023feature}:} Group effect evaluates negative transfer caused by inefficient model aggregation in FL, aiming to quantify the impact of client data differences on diverse local model updates that ultimately lead to negative transfer in the parameter space.
\begin{equation}
    \begin{aligned}
        &TTA_f(\mathcal{G}_t)  = \frac{\sum_{(x, y) \in D_T} \mathbbm{1} {\{\mathop{\arg\max}_j \mathcal{F}(x; \mathcal{G}_t)_j = y\}}}{|D_T|}  \\
        &GE_t = \frac{1}{M} \sum_{i \in {1, 2, \cdot\cdot\cdot, M}} TTA_f(\mathcal{G}_t + \Delta_t^{(i)}) + TTA_f(\mathcal{G}_{t+1})
    \end{aligned}
\end{equation}
where $\mathcal{F}(x)$ is the neural network classifier and $\mathcal{F}(x)_j$ denotes the $j$-th element of $\mathcal{F}(x)$, $D_T$ represents the target domain dataset, $|D_T|$ denotes the size of the target domain dataset, $\mathcal{G}_t$ is the global model at time step $t$ in the target task, $\Delta_t^{(i)}$ is the update of domain $i$.
\end{itemize}

\begin{table}[!t]
\caption{Summary of evaluations and benchmarks.}\label{tab:evaluation}
\centering
% \fbox{
\begin{tabular}{cccc}
    \toprule
    \textbf{Paradigm} & \textbf{Method} & \textbf{Strategy} & \textbf{Metrics} \\
    \midrule
    \multirow{9}*{DG} & DANN \cite{ganin2016domain}& TEVS & Accuracy\\
    ~  & JiGen \cite{carlucci2019domain} & LODO,TRVS & Accuracy\\
    ~ & Epi-FCR \cite{li2019episodic} & TEVS & Accuracy\\
    ~ & MTSSL \cite{albuquerque2020improving} & LODO & Accuracy\\
    ~ & EISNet \cite{wang2020learning} & LODO,TRVS & Accuracy\\
    ~ & L2A-OT \cite{zhou2020learning} & LODO & Accuracy\\
    ~ & DSON \cite{seo2020learning} & LODO & Accuracy\\
    ~ & Mixstyle \cite{zhou2021domain} & LODO & Accuracy\\
    ~ & RSC \cite{huang2020self} & LODO & Accuracy\\
    \midrule
     
    \multirow{4}*{FL} & FedAvg\cite{mcmahan2017communication} & TRVS & Accuracy \\
    ~ & FedProx\cite{li2020federated2} & TRVS & Accuracy\\
    ~ & Scaffold \cite{karimireddy2020scaffold} & TRVS & Accuracy\\
    ~ & Moon \cite{li2021model}& TRVS & Accuracy\\
    \midrule
     \multirow{10}*{FDG} & FedDG\cite{liu2021feddg} & LODO & Dice, HD\\
    ~ & CSAC\cite{yuan2023collaborative} & LODO & Accuracy\\
     ~ & FedADG\cite{zhang2023federated} & LODO & Accuracy\\
     ~ & COPA\cite{wu2021collaborative} & LODO & Accuracy \\
    ~ & FedHealth\cite{chen2020fedhealth}& TEVS & Accuracy\\
    ~ & FedIG(-A)\cite{yangclient}& LODO & Accuracy, SD\\
    ~ & FADH \cite{xu2023federated} & LODO & Accuracy\\
    ~ & CCST \cite{chen2023federated} & LODO & Accuracy\\
    ~ & FedINB \cite{zhou2023efficient}& TEVS & WD, FID\\
    ~ & FOSDA \cite{qin2022uncertainty}& TEVS & OS, OS*, UNK\\
    ~ & FedDrive \cite{fantauzzo2022feddrive}& TEVS & mIoU $\pm$ SD\\
    ~ & FedKA \cite{sun2023feature} & LODO & GE\\
     ~ & FOSDA \cite{qin2022uncertainty} & LODO & $\mathcal{A}$-distance\\
    \bottomrule
\end{tabular}
% }
\end{table}
\subsection{Benchmarks}
We present a comprehensive summary of existing benchmarks for evaluating the performance of FDG algorithms. Our analysis covers benchmarks in three categories: centralized DG, FL, and benchmarks specifically designed for FDG. Also, the detailed summary of these evaluations and benchmarks can be found below and in Table~\ref{tab:evaluation}. 
\begin{itemize}
    \item \textbf{DANN\cite{ganin2016domain}} is a neural network architecture designed to accomplish precise classification of source data while simultaneously learning feature representations that exhibit invariance across multiple source domains.
    \item \textbf{JiGen\cite{carlucci2019domain}}  is a supervised framework that leverages jigsaw puzzles as a training task to learn effective generalization across diverse visual domains.
    \item \textbf{Epi-FCR\cite{li2019episodic}} is a scheme that learns domain shift using episodic training.
    \item \textbf{MTSSL\cite{albuquerque2020improving}} is a method that facilitates the learning of transferable features by employing a self-supervised task focused on predicting Gabor filter bank responses.
    \item \textbf{EISNet\cite{wang2020learning}} is an innovative network that synergistically integrates self-supervised learning and metric learning approaches, effectively enhancing classifier performance specifically in target domains.
    \item \textbf{L2A-OT\cite{zhou2020learning}} is a method that leverages synthetic data augmentation to learn domain-invariant features, facilitating effective generalization across different domains.
    \item \textbf{DSON\cite{seo2020learning}} is a novel scheme that effectively enhances the generalization performance on target domains by integrating batch normalization and instance normalization methods.
    \item \textbf{Mixstyle\cite{zhou2021domain}} is an innovative method that combines features from different source domains to generate synthetic source domains, comprehensively enabling optimization of model generalization.
    \item \textbf{RSC\cite{huang2020self}} is an approach that selectively discards dominant features present in the training data to optimize the generalization capability of a model.
    \item \textbf{FedAvg\cite{mcmahan2017communication}} (Federated Averaging) is a distributed learning algorithm commonly used in FL settings without any generalization technique.
    \item \textbf{FedProx\cite{li2020federated2}} is a novel framework that tackles heterogeneity in federated networks by serving as a generalized and re-parametrized version of FedAvg.
    \item \textbf{Scaffold\cite{karimireddy2020scaffold}} is a novel algorithm that utilizes control variates to effectively address 'client-drift' and minimize communication rounds, demonstrating resilience to data heterogeneity and client sampling while leveraging data similarity for faster convergence.
    \item \textbf{Moon\cite{li2021model}} is an innovative FL framework, employing model-level contrastive learning to tackle the issue of heterogeneity in local data distribution.
    \item \textbf{FedDG\cite{liu2021feddg}} (FDG) is a solution that aims to learn a federated model such that it can directly generalize to completely unseen domains.
    \item \textbf{CSAC\cite{yuan2023collaborative}} is a novel privacy-preserving method for the separated DG task.
    \item \textbf{FedADG\cite{zhang2023federated}} (Federated Adversarial Domain Generalization) is a scheme that utilizes federated adversarial learning to solve the DG problem in FL for IoT devices.
    \item \textbf{COPA\cite{wu2021collaborative}} (Collaborative Optimization and Aggregation) is a decentralized approach for DG and multisource unsupervised DA, utilizing a collaborative optimization and aggregation process to construct a generalized target model without sharing data across domains.
    \item \textbf{FedHealth\cite{chen2020fedhealth}} is a pioneering FTL framework for personalized wearable healthcare, leveraging FL and TL to overcome cloud-based personalization limitations.
    \item \textbf{FedIG(-A)\cite{yangclient}} presents a novel approach that integrates client-agnostic learning with a combination of local training using mixed instance-global statistics and zero-shot adaptation through estimated statistics for inference.
    \item \textbf{FADH\cite{xu2023federated}} (Federated Adversarial Domain Hallucina) is an innovative FDG approach that prioritizes domain hallucination to generate samples, optimizing the global model's entropy while minimizing the cross-entropy of the local model.
    \item \textbf{CCST\cite{chen2023federated}} is a novel DG method for in FL, enabling style transfer across clients without data exchange to promote uniform source client distributions and mitigate model biases by aligning local models with the image styles of all clients.
    \item \textbf{FedINB\cite{zhou2023efficient}} is an innovative federated domain translation method that generates pseudodata specific to each client, offering potential benefits for multiple downstream learning tasks.
    \item \textbf{FOSDA\cite{qin2022uncertainty}} (Federated OSDA) is an advanced federated algorithm that incorporates an uncertainty-aware mechanism to generate a global model by prioritizing source clients with high uncertainty while preserving high consistency.
    \item \textbf{FedDrive\cite{fantauzzo2022feddrive}} is an innovative benchmark framework that includes real-world challenges of statistical heterogeneity and DG.
    \item \textbf{FedKA\cite{sun2023feature}} (Federated Knowledge Alignment) is an innovative FDG method that utilizes feature distribution matching and a federated voting mechanism to enable the global model to learn domain-invariant client features and refine its performance with target domain pseudo-labels.
\end{itemize}
\section{Future research directions} \label{future}
% \subsection{Challenges}
This paper offers an extensive and meticulous survey of the existing literature on FDG, highlighting its inherent advantages as a fusion of FL and DG, resulting in privacy-preserving, scalable, and diverse models capable of effective generalization across multiple domains.  \hlyellow{To be precise, FDG is a promising technological breakthrough as a subfiled within ML that aims to develop models capable of generalizing well across multiple unseen domains while achieving privacy-preserving. Most technological developments come with benefits, challenges and limitations, and FDG is not exceptional. In the near future, several challenges need to beovercome in order to retain more benefits out of FDG. One of the primary difficulties faced in FDG is domain shift, which occurs when a model trained on data from one domain performs poorly on data from a different domain due to significant differences in statistical properties. Additionally, data heterogeneity, limited data availability, high communication overhead and privacy concerns pose significant dilemmas for FL.} %In addition to these, some challenges still need to be addressed, which are discussed below. %The distribution of data across multiple devices or organizations in FDG can result in significant heterogeneity, limited data for training, and privacy concerns due to the distribution of data. Moreover, high communication overhead can hinder efficient communication between domains during FL, which can further exacerbate these obstacles. 
% Collectively, by \hlyellow{considering} these challenges, FDG can advance the development of privacy-preserving, robust, and generalized ML models. This will not only benefit academic research but also have practical implications in real-world scenarios where the ability to generalize across multiple unseen domains is crucial.
% However, it is important to emphasize that despite these benefits, there remain numerous unresolved issues that hinder the realization of FDG. 
\hlyellow{To effectively tackle these issues}, a comprehensive identification and analysis of these challenges is crucial, accompanied by the pursuit of novel theoretical and technical solutions. Within this context, this study examines key problems in FDG, including privacy-preserving FDG, communication-efficient FDG, computation-efficient FDG, heterogeneity, label shift, scalable FDG, continuous FDG, and FDG to novel categories.
\subsection{Privacy-preserving FDG}
\hlyellow{While FL inherently safeguards privacy by retaining data within local domains, there is a need for further development of privacy-preserving mechanisms to guarantee that sensitive information remains secure throughout the model aggregation phase.} The existing FDG methods usually neglect the research on privacy-preserving mechanisms \cite{li2021fedh2l, yangclient, chen2023federated, liu2021feddg, yuan2023collaborative, zhou2022source, zhou2023efficient}. Therefore, how to design robust and privacy-preserving models that can generalize well across diverse domains while protecting the sensitive information of individual clients or domains is worth investigating. Here are some approaches related to privacy-preserving FDG: Secure aggregation \cite{bonawitz2017practical}, differential privacy \cite{wei2020federated}, privacy-aware model selection\cite{chang2023pagroup}, privacy-preserving data preprocessing \cite{choudhury2020anonymizing}, and privacy regulations \cite{cheng2020federated}.
\subsection{Communication-efficient FDG}
Communication is a major bottleneck for many real-world applications in FL, as transmitting large model updates from multiple clients to a central server can be time-consuming and resource-intensive. Some work \cite{wu2021collaborative, chen2023federated} has demonstrated that certain proposed schemes in FDG can lead to increased communication costs. On the other hand, \cite{zhou2023efficient} has highlighted that neglecting communication limitations can result in poor model performance in FDG. In this case, it is of utmost importance to achieve communication efficiency in FDG, enabling faster and more scalable learning across diverse domains while minimizing the communication overhead and associated costs. Here are some strategies and techniques for achieving communication efficiency: Model compression and quantization \cite{wang2019eigendamage, li2020few}, differential updates, selective model aggregation \cite{ye2020federated}, local adaptation \cite{deng2020adaptive}, and communication-efficient aggregation algorithms \cite{guo2020v}.
\subsection{Computation-efficient FDG}
Similar to communication efficiency, Computation efficiency is also a crucial aspect of FDG, as it affects the speed and scalability of the learning process. While there are some studies in the field of FDG have indeed highlighted the need for increased computational resources \cite{wu2021collaborative,yangclient} and certain proposed methods may not be suitable where the clients have limited computational resources \cite{chen2023federated, liu2021feddg}, there are only very few investigations on reducing the requirement for computational resources \cite{khodak2019adaptive}. Here are some strategies and techniques to achieve computation efficiency: model architecture optimization \cite{zhu2021federated}, adaptive learning algorithms \cite{wang2019adaptive}, communication optimization \cite{wang2019eigendamage, li2020few}, collaborative learning \cite{yu2021toward}, efficient data sampling and preprocessing \cite{bettini2021personalized}.
\subsection{Heterogeneity}
Heterogeneity in FL leads to performance disparities, communication inefficiency, privacy and security risks, bias and fairness issues, scalability limitations, and barriers to generalization across domains. In FDG, the participating clients or domains may have significant variations in terms of data distributions, feature representations, or label spaces, leading to a crucial dilemma. Only a limited number of recent studies paid attention to this concern, such as data and model heterogeneity tasks \cite{chen2022learning, huang2022learn, zhou2023efficient} and Non-IID \cite{wang2022graphfl}. Here are some approaches to solving heterogeneity in FDG: data preprocessing (feature scaling\cite{yu2020heterogeneous}, data augmentation \cite{hao2021towards, de2022mitigating}, or DA \cite{shenaj2023learning, andreux2020siloed}), model adaptation \cite{ma2022state, luo2021cost}, model personalization \cite{tan2022towards}, adaptive aggregation \cite{ji2021emerging}. 
\subsection{Label shift}
Label shift in FDG refers to the situation where there is a discrepancy in label distributions between different domains in the FL setting, which could lead to performance disparities and hinder the overall generalization capability of the FL system. Therefore, mitigating label shifts is crucial to ensure the robustness and accuracy of ML models in real-world scenarios. Very few recent works \cite{xu2023federated, jiang2023federated} paid attention to Label shift. Here are some potential approaches to solve label shifts in FDG: domain adaptation \cite{ji2021emerging} and transfer learning \cite{gao2019}.
\subsection{Scalable FDG}
Scalability is a crucial aspect of FDG, as it ensures that the approach can effectively handle large-scale and diverse datasets from multiple domains. The work \cite{yuan2023collaborative} pointed out that exploring scalable methods to implement FSG for large-scale datasets is very important. Up to now, only FADH \cite{xu2023federated} has been proven to be scalable to different numbers of source domains. Here are some key considerations for achieving scalability in FDG: distributed computation \cite{zhang2022federated, guberovic2021dew}, communication efficiency \cite{reisizadeh2020fedpaq, kang2022communication}, model compression \cite{lang2023cpa}, model aggregation\cite{lee2021layer}, and parallel processing \cite{gupta2022fl}.
\subsection{Continuous FDG}
Continuous FDG refers to the ability of the FL system to continuously adapt and generalize across multiple domains over time. However, the predominant research on FDG assumes fixed source domains and a one-time model learning process, which may not fully capture the dynamic nature of real-world scenarios. \hlyellow{Only a limited number of studies have explored continuous FDG, with a notable recent investigation} \cite{li2023d} \hlyellow{employing incremental learning strategies to tackle this issue. This approach effectively mitigates catastrophic forgetting and boosts generalization performance.}
Here are some key aspects and approaches related to continuous FDG: incremental learning \cite{you2022incremental}, adaptive learning \cite{deng2020adaptive}, DA \cite{wang2022framework}, and lifelong learning\cite{kopparapu2020fedfmc, shoham2019overcoming}.
\subsection{FDG to novel categories}
FDG to novel categories highlights the proficiency of FL systems in extending their generalization capabilities to previously unseen or novel categories not included in the training datasets of participating domains. Typically, existing FDG algorithms operate under the assumption that the label spaces—defined as the set of potential categories—across different domains are uniform. However, a more nuanced and applicable scenario emerges when the source and target domains possess both overlapping and distinct label spaces, introducing an augmented category gap between them\cite{You_un}. Here are some key techniques that could be used for FDG to novel categories: transfer learning \cite{zellinger2021beyond}, meta-learning \cite{nguyen2022fedsr}, or DA \cite{qin2022uncertainty}.
\section{Conclusion} \label{conclusion}
FDG is a significant research area within machine learning, focusing on generalization learning and privacy-preserving in distributed scenarios. This paper offers a comprehensive analysis of the field of FDG, encompassing theoretical foundations, existing methodologies, available datasets, practical applications, evaluation strategies, evaluation metrics, and benchmarks. Additionally, through a meticulous analysis of these methods, several prospective research challenges are identified, which can pave the way for future investigations in the field. This survey aims to offer valuable insights to researchers and serve as a source of inspiration for future advancements in the field.

\section*{Acknowledgment}
This work is supported by the National Natural Science Foundation of China under Grant No.92267206 and No. 62032013. Research also has partially received funding from the European Commission Horizon 2020 with grant agreement No. 101135576 (INTEND) and No. 101070186 (TEADAL).

\bibliographystyle{IEEEtran}
\bibliography{IEEE}

\begin{IEEEbiography}
[{\includegraphics[width=1in,height=1.25in,clip,keepaspectratio]{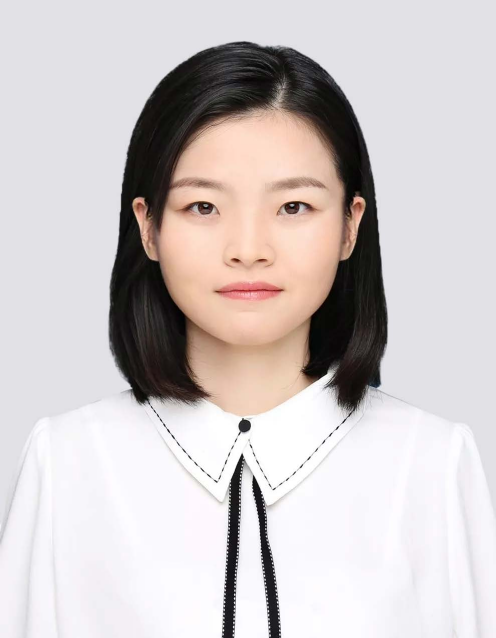}}]
{Ying Li} received the B.S. degree in the Internet of Things from Anyang Institute Of Technology, Anyang, China, in 2017, and the M.S. degree in computer technology from Northeastern University, Shenyang, China, in 2020, where she is currently pursuing the Ph.D. degree in computer science and technology. She is a visiting PhD at Distributed Systems Group, TU Wien, Austria from 2022 to 2024. 

Her research interests include distributed machine learning, blockchain, and knowledge-defined networking. 
\end{IEEEbiography}\vskip -2\baselineskip plus -1fil
\begin{IEEEbiography}
[{\includegraphics[width=1in,height=1.25in,clip,keepaspectratio]{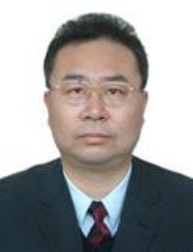}}]
{Xingwei Wang} received the B.S., M.S., and Ph.D.
degrees in computer science from Northeastern University, Shenyang, China, in 1989, 1992, and 1998, respectively.
He is currently a Professor with the College of Computer Science and Engineering, Northeastern University. He has published more than 100 journal articles, books and book chapters, and refereed conference papers.

His research interests include cloud computing and future Internet.
Prof. Wang has received several best paper awards.
\end{IEEEbiography}\vskip -2\baselineskip plus -1fil
\begin{IEEEbiography}
[{\includegraphics[width=1in,height=1.25in,clip,keepaspectratio]{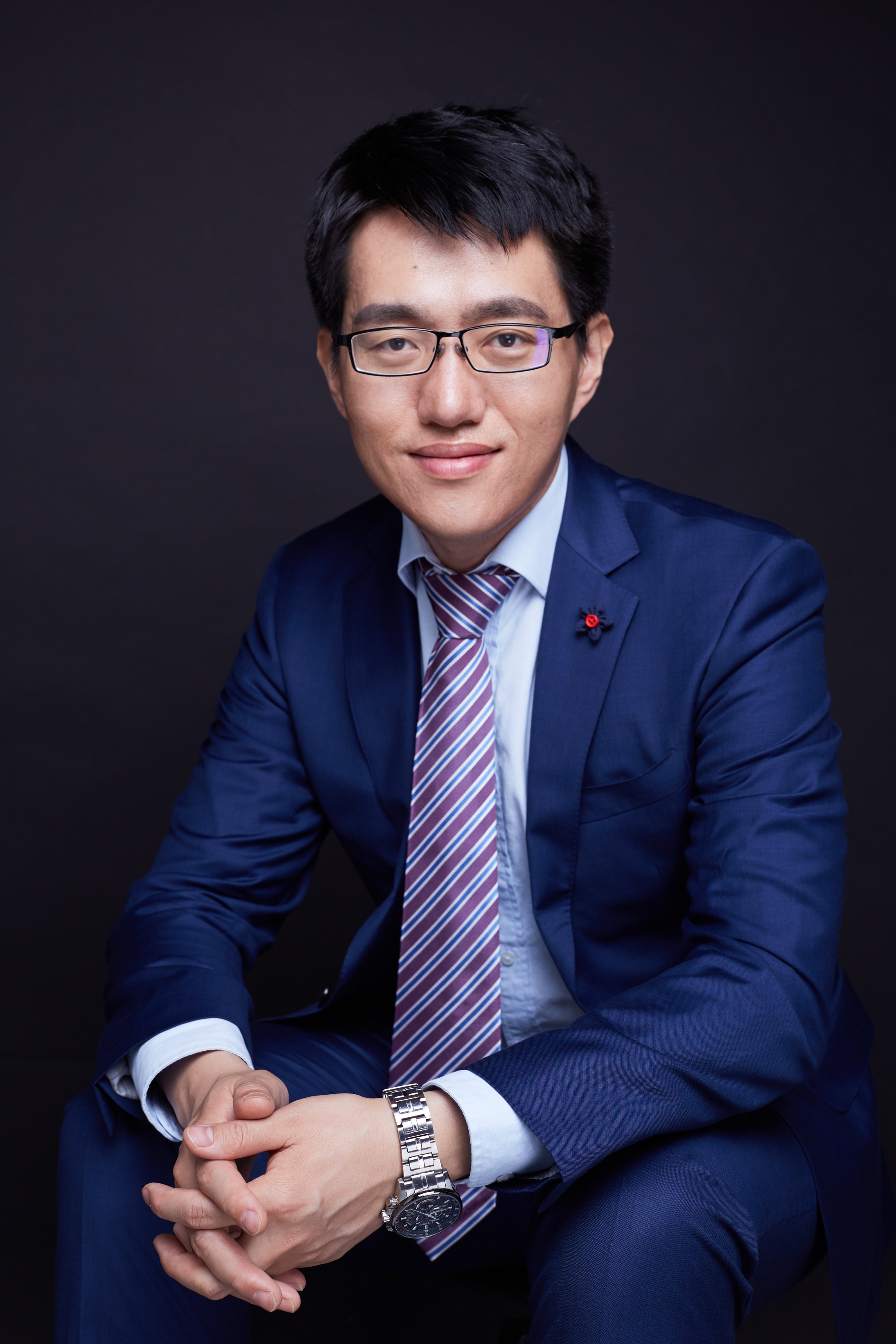}}]
{Rongfei Zeng}  is the Associate Professor of Software College at Northeastern University. He received his Ph.D. degree in Computer Science and Technology from Tsinghua University with honor in 2012. He has published several papers in top journals and conferences such as IEEE Transactions on Parallel and Distributed Systems (TPDS), Elsevier Computer Networks, IEEE ICDCS.

His research interests include network security and privacy, machine learning and its security, and industrial networks and IoT.
\end{IEEEbiography}\vskip -2\baselineskip plus -1fil

\begin{IEEEbiography}[{\includegraphics[width=1in,height=1.25in, clip,keepaspectratio]{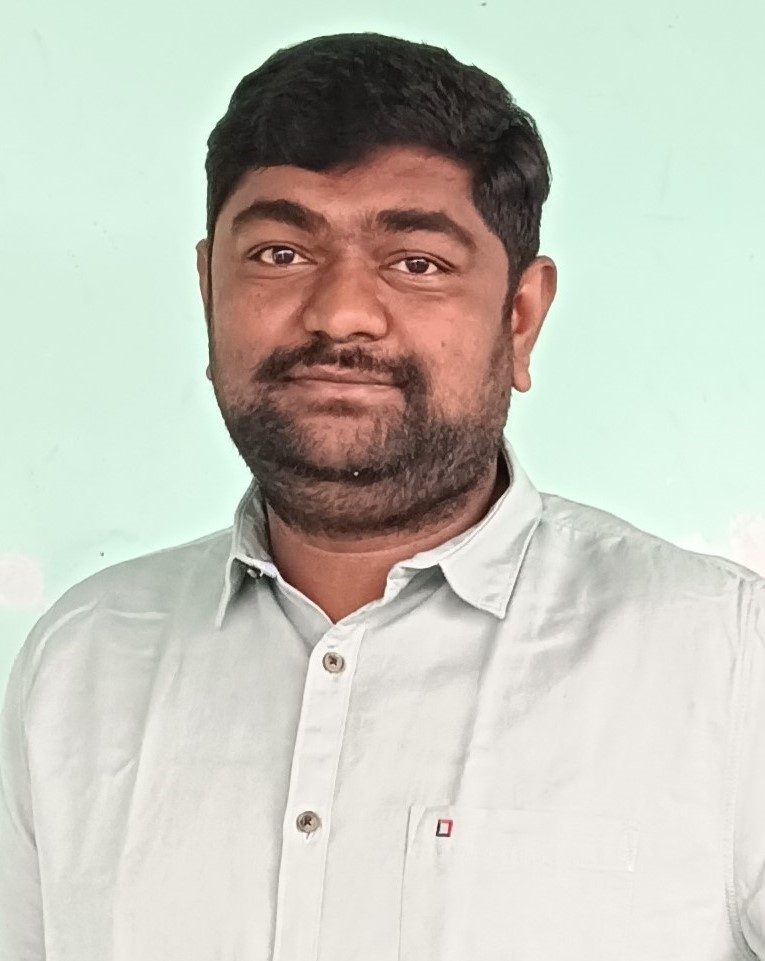}}] {Praveen Kumar Donta (SM'22)}~is a Postdoctoral researcher in the Distributed Systems Group, TU Wien, Austria. He received his Ph.D. from the Department of Computer Science and Engineering in Indian Institute of Technology (Indian School of Mines), Dhanbad, India. He was a visiting Ph.D. student at the University of Tartu, Estonia. He received his Masters and Bachelor of Technology from JNTU Anantapur, India in 2014, and 2012. He is serving as Editorial board member for several journals including Measurement, Measurement: Sensors, Computer Communications, Elsevier, Computing Springer, and PLOS ONE.  His current research on Learning-driven distributed computing continuum systems, Edge Intelligence, and Intelligent data protocols. 
\end{IEEEbiography}\vskip -2\baselineskip plus -1fil

\begin{IEEEbiography}[{\includegraphics[width=1in,height=1.25in, clip,keepaspectratio]{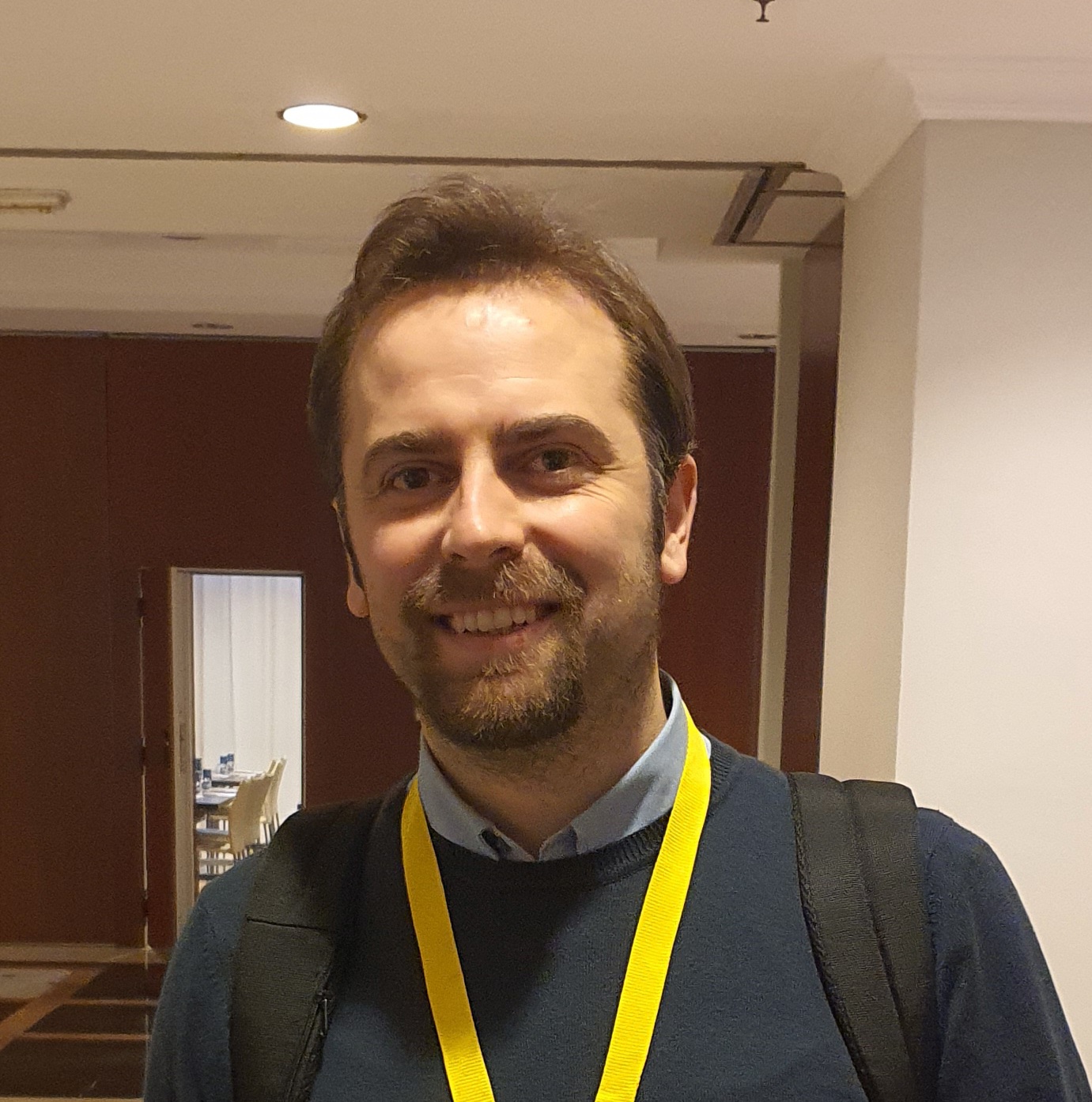}}] {Ilir Murturi} (Member, IEEE) is a Postdoctoral Researcher in the Distributed Systems Group, TU Wien, Austria. He received a Ph.D. in the Distributed Systems Group, Technische Universität Wien (TU Wien), Vienna, Austria, and an MSc in Computer Engineering from the University of Prishtina, Prishtina, Kosova. He is a member of IEEE and a reviewer for several journals and conferences. His current research interests include the Internet of Things, Distributed Computing Continuum Systems,  EdgeAI, and privacy in distributed, self-adaptive and cyber-physical systems. 
\end{IEEEbiography}\vskip -2\baselineskip plus -1fil

\begin{IEEEbiography}
[{\includegraphics[width=1in,height=1.25in,clip,keepaspectratio]{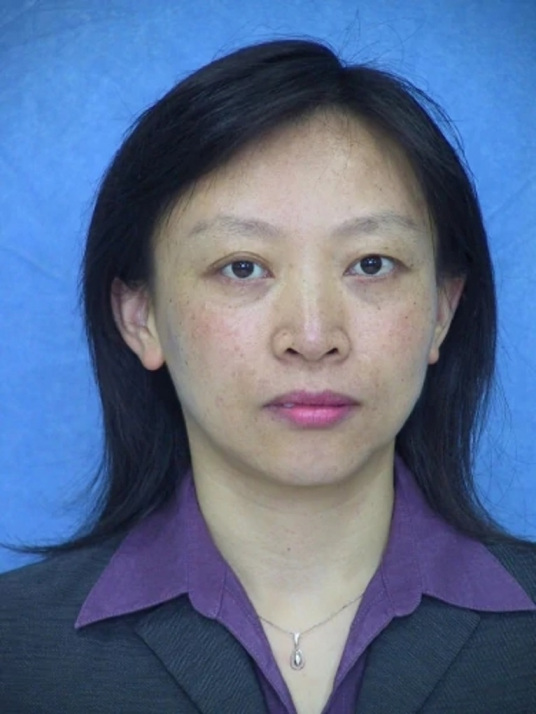}}]
{Min Huang} (Member, IEEE) received the B.S. degree in automatic instrument, the M.S. degree in systems engineering, and the Ph.D. degree in control theory from Northeastern University, Shenyang, China, in 1990, 1993, and 1999, respectively. She is currently a Professor with the College of Information Science and Engineering, Northeastern University. She has published more than 100 journal articles, books, and refereed conference papers. 
Her research interests include modeling and optimization for logistics and supply chain system.
\end{IEEEbiography}\vskip -2\baselineskip plus -1fil

\begin{IEEEbiography}[{\includegraphics[width=1in,height=1.25in, clip,keepaspectratio]{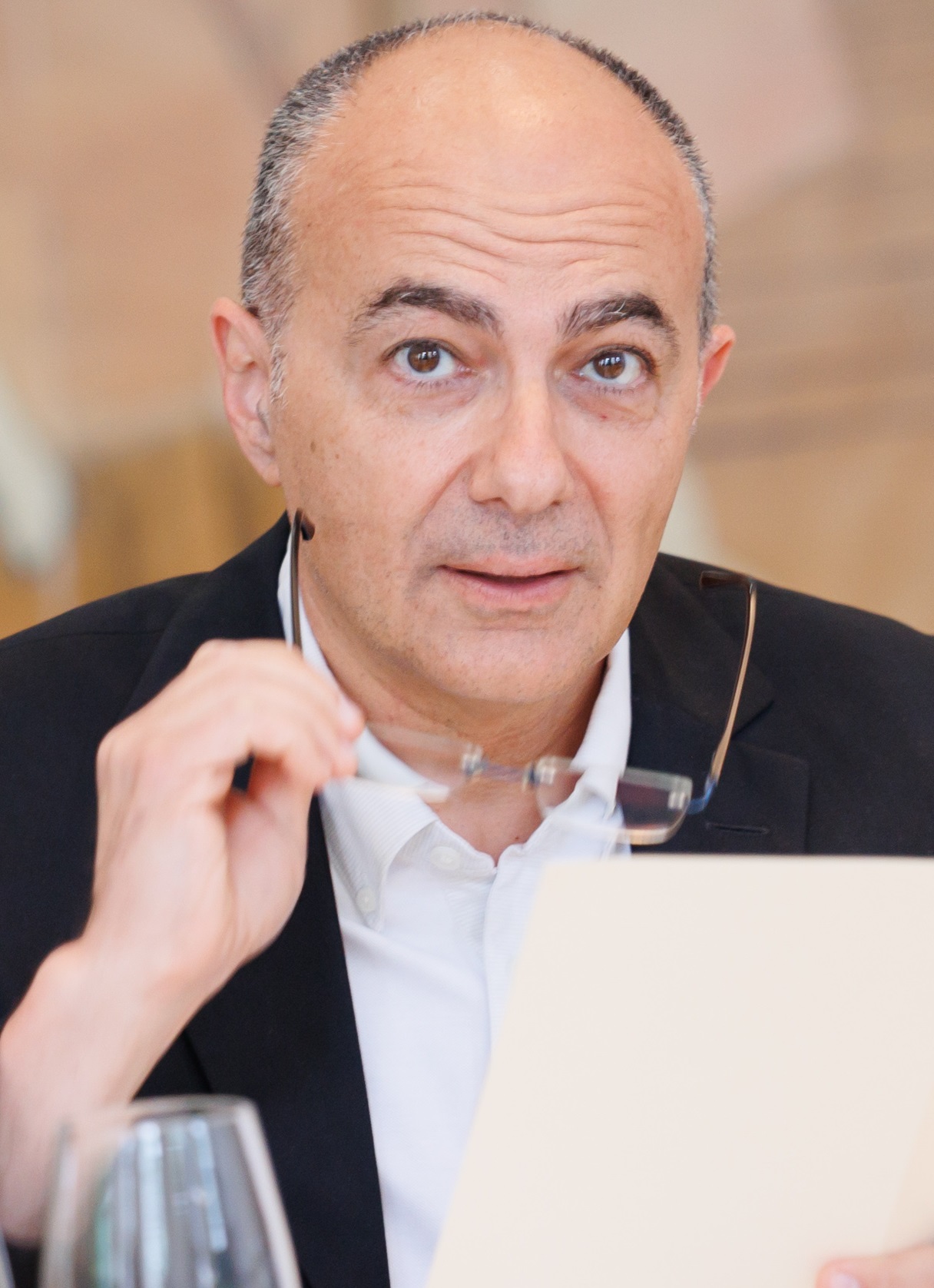}}] {Schahram Dustdar} (Fellow, IEEE) is a full professor of computer science (informatics) with a focus on Internet Technologies heading the Distributed Systems Group at the TU Wien. He is member of the Academia Europaea. He is the recipient of the ACM Distinguished Scientist Award and Distinguished Speaker award, and the IBM Faculty Award. He is an associate editor of IEEE Transactions on Services Computing, ACM Transactions on the Web, and ACM Transactions on Internet Technology, and on the editorial board of IEEE Internet Computing and IEEE Computer. He is the editor-in-chief of Computing (an SCI-ranked journal of Springer).
\end{IEEEbiography}
\vfill
\end{document}